\documentclass[10pt,twocolumn,letterpaper]{article}

\usepackage{iccv}              
\usepackage{graphicx}

\usepackage{amsmath,amsfonts,bm}

\def\eqref#1{equation~\ref{#1}}

\def\1{\bm{1}}

\DeclareMathAlphabet{\mathsfit}{\encodingdefault}{\sfdefault}{m}{sl}
\SetMathAlphabet{\mathsfit}{bold}{\encodingdefault}{\sfdefault}{bx}{n}

\usepackage{url}
\usepackage{booktabs}

\usepackage{nicefrac}
\usepackage{arydshln}
\usepackage{array}
\usepackage{adjustbox}
\usepackage{array, makecell} 

\usepackage{color, colortbl}
\usepackage{pifont} 
\definecolor{Gray}{gray}{0.9}
\definecolor{brightturquoise}{rgb}{0.85, 1, 1}
\definecolor{newpurple}{HTML}{BC61F5}

\newcommand{\myccnew}{\cellcolor[rgb]{0.61, 0.87, 1.0}}

\definecolor{cvprblue}{rgb}{0.21,0.49,0.74}
\usepackage[pagebackref,breaklinks,colorlinks,citecolor=cvprblue]{hyperref}

\def\paperID{7794} 
\def\confName{ICCV}
\def\confYear{2025}
\usepackage{multirow}

\newcommand{\reffig}[1]{\text{Figure~\ref{#1}}}
\newcommand{\reftab}[1]{\text{Table~\ref{#1}}}

\newcommand\blfootnote[1]{%
  \begingroup
  \renewcommand\thefootnote{}\footnote{#1}%
  \addtocounter{footnote}{-1}%
  \endgroup
}
\title{Vision-Language Models Can't See the Obvious}

\author{
Yasser Dahou\thanks{Joint first authors} \hspace{2em}
Ngoc Dung Huynh\footnotemark[1] \hspace{2em}
Phuc H. Le-Khac \hspace{2em} \\
Wamiq Reyaz Para \hspace{2em}
Ankit Singh \hspace{2em}
Sanath Narayan \\
\\
Technology Innovation Institute, Abu Dhabi, UAE \\
\texttt{\url{https://salbench.github.io}}
}

\begin{document}

\maketitle

\begin{abstract}

We present Saliency Benchmark (SalBench), a novel benchmark designed to assess the capability of Large Vision-Language Models (LVLM) in detecting visually salient features that are readily apparent to humans, such as a large circle amidst a grid of smaller ones. This benchmark focuses on low-level features including color, intensity, and orientation, which are fundamental to human visual processing. Our SalBench consists of images that highlight rare, unusual, or unexpected elements within scenes, and naturally draw human attention. It comprises three novel tasks for evaluating the perceptual capabilities of LVLM: Odd-One-Out Detection, Referring Odd-One-Out, and Visual Referring Odd-One-Out. We perform a comprehensive evaluation of state-of-the-art LVLM using SalBench and our findings reveal a surprising limitation: LVLM struggle to identify seemingly obvious visual anomalies, with even the advanced GPT-4o achieving only 47.6\% accuracy on such a simple task. SalBench will be an important step in measuring the capabilities of LVLM that align with the subtle definition of human attention.

\blfootnote{Correspondence: yasser.djilali@tii.ae}

\end{abstract}

\section{Introduction}

Large Vision-Language Models (LVLM) have emerged as a central focus in recent computer vision research~\cite{alayrac2022flamingo, liu2024visual, zhu2023miniGPT, chen2024internvl, liu2023mmbench, schwenk2022okvqa}. The primary advantage of these models lies in their ability to reason about images using the Large Language Model (LLM) knowledge about the world, and solve tasks that surpass the capabilities of traditional vision models. For example, while classical object detectors excel at identifying objects within a scene, they struggle with more complex tasks such as understanding object relationships and spatial correspondences. LVLM address this limitation by enabling better reasoning about the visual content using the prior world knowledge.

\begin{figure}[t]
    \centering
    \includegraphics[clip=true, trim=0.5em 1.5em 0.4em 1.5em, width=\columnwidth]{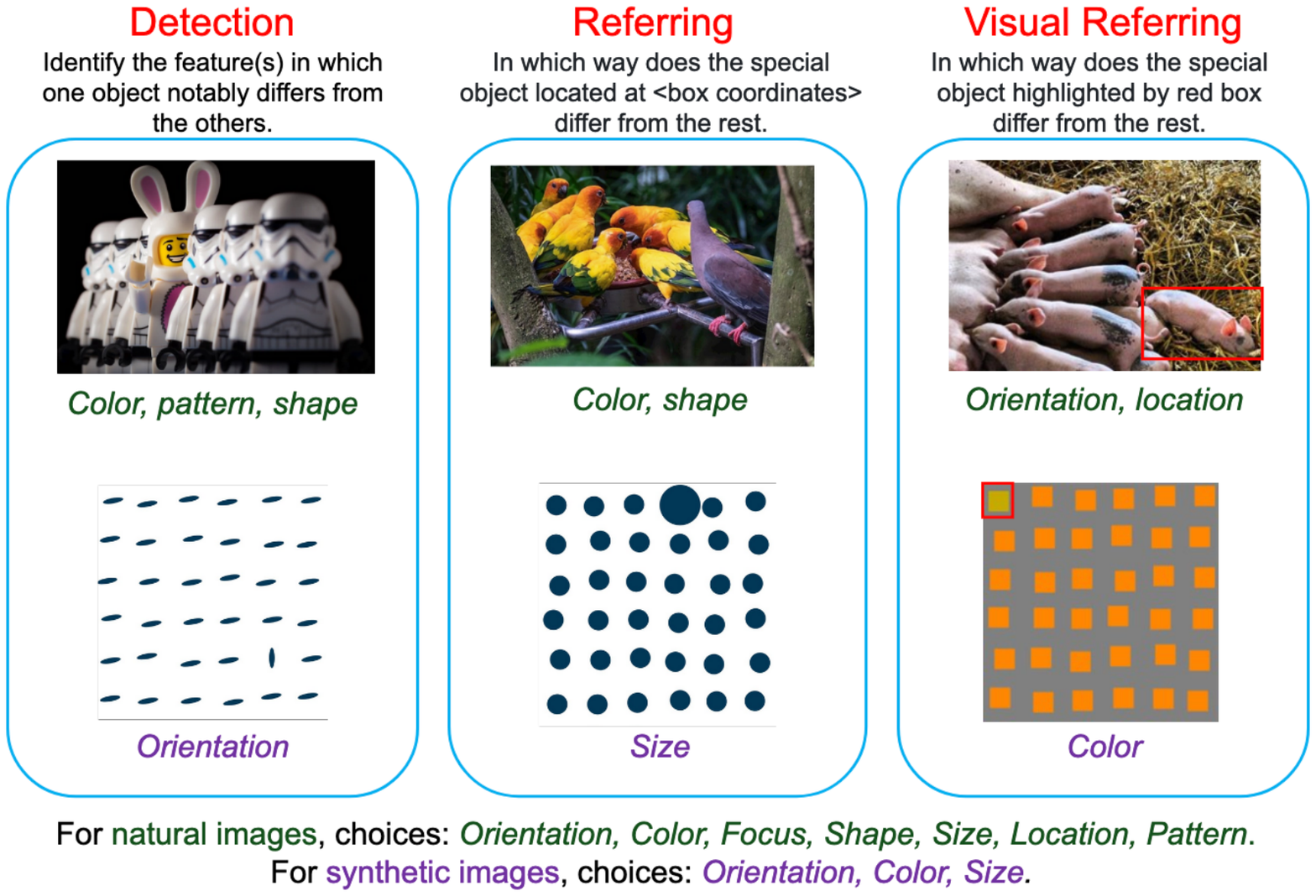}
    \caption{\textbf{Our SalBench evaluates the perceptual capabilities of vision-language models}. It comprises natural and synthetic images, which consist of a single salient target among multiple distractors. In natural images, the target can vary from the rest in orientation, color, focus, shape, size, location and pattern, while it differs only in orientation, color or size in synthetic images. SalBench evaluates the odd-one-out understanding under three tasks: Detection, Referring and Visual Referring. In detection task, the odd features must be directly predicted. Moreover, the box coordinates of the target are provided as context in text (referring task) or as a highlighted box in the image (visual referring task) for predicting the odd features of the target. Best viewed zoomed-in.\vspace{-0.3cm}}
    \label{fig:SalBench}
\end{figure}
\vspace{-0.1cm}
Current LVLM exhibit impressive performance on existing benchmarks~\cite{lu2023mathvista, goyal2017making, seedbench, chen2024we, yu2023mm, yue2024mmmu}, which evaluate on a range of capabilities, from general visual question answering~\cite{goyal2017making} to tasks requiring college-level subject knowledge and critical reasoning, such as MMMU~\cite{yue2024mmmu}. With the main focus of these benchmarks being high-level complex tasks, it raises a fundamental question: can LVLM perform equally well on simple perceptual tasks, such as identifying a black dot on a white background? This aligns with Moravec's paradox~\cite{moravec1988mind}, which suggests that high-level reasoning tasks are computationally simpler for artificial intelligence systems than low-level perceptual and sensorimotor skills. Consequently, we may find that LVLM excel at complex tasks present in existing benchmarks while struggling with seemingly simple perceptual tasks that humans perform effortlessly. To this end, we propose a benchmark for quantifying the alignment of vision-language models on low-level perceptual tasks of the human attention.

While LVLM effectively capture high-level features such as cars and humans, they are likely to struggle to represent crucial aspects of human visual attention that have been extensively studied in neuroscience. Visual search, which is a fundamental process shaping human attention ~\cite{treisman1980feature, kotseruba2020saliency} involves the brain's parallel processing of regions that differ significantly in one feature dimension, such as color, intensity, or orientation. These low-level features serve as basic mechanisms of the human visual system. By examining the LVLM performance on simple saliency-driven images, we aim to gain insights into the current state of these models relative to human perceptual abilities and identify areas for necessary improvements. Our key contributions are:

\begin{itemize}

    \item We propose SalBench as an open-source benchmark, for evaluating and aiding the improvement of the perceptual capabilities of LVLM. To this end, we augment the P3/O3 datasets~\cite{kotseruba2020saliency}, which comprise of images with a single distinctive target among many similar distractors, with language instructions and create three novel tasks: Odd-One-Out Detection, Referring Odd-One-Out, and Visual Referring Odd-One-Out, as shown in \reffig{fig:SalBench}. Our benchmark aims to serve as a tool for assessing the progress in aligning LVLM with human visual attention. 

    \item We conduct a comprehensive analysis of LVLM performance on SalBench, uncovering striking discrepancies between these advanced models and human visual capabilities. Our findings show that even state-of-the-art LVLM, including GPT-4o~\cite{GPT4o}, struggle with basic saliency detection tasks that are trivial for humans.

    \item We provide insights into the limitations of LVLM  in processing low-level visual features, highlighting the need for improved alignment between these models and human visual attention mechanisms. Our work demonstrates the importance of incorporating neuroscience principles into the development of future vision-language models.

\end{itemize}

\section{Related Works}

Many vision-language benchmarks have been introduced in the literature for evaluating and comparing LVLM on various tasks. The early benchmarks were mostly single-task oriented and can be broadly classified into captioning, general visual-question answering (VQA), and text-centric VQA. The captioning task is used to measure the LVLM's caption generation quality and is mostly evaluated on COCO~\cite{coco} and NoCaps~\cite{agrawal2019nocaps} benchmarks using BLEU, CIDER, ROUGE metric scores. Similarly, general VQA benchmarks like VQAv2~\cite{goyal2017making}, GQA~\cite{hudson2019gqa}, ScienceQA~\cite{scienceqa}, OK-VQA~\cite{schwenk2022okvqa}, VizWiz~\cite{gurari2018vizwiz}, Pope~\cite{pope} typically ask general questions about the objects/scene in the image. Furthermore, text-centric VQA benchmarks such as OCRVQA~\cite{ocrvqa}, TextVQA~\cite{textvqa}, ST-VQA~\cite{stvqa}, DocVQA~\cite{mathew2021docvqa}, ChartQA~\cite{masry2022chartqa}, InfoVQA~\cite{mathew2022infographicvqa}, and AI2D~\cite{kembhavi2016diagram} additionally focus on the vision-language model's ability to detect text in the image (Optical Character Recognition) and then answer the questions. 

With the increasing capabilities of vision-language models in handling different types of tasks, the benchmarks have also evolved to evaluate the LVLM in a fine-grained manner. To this end, more recently, MMBench~\cite{liu2023mmbench}, MME~\cite{fu2023mme}, MM-Star~\cite{chen2024we}, and MMMU~\cite{yue2023mmmu} are curated to test the LVLM on a mix of different aspects, such as reasoning, perception, knowledge, chart interpretation, \etc. Similarly, MathVista~\cite{lu2023mathvista}, MathVision~\cite{wang2024measuring} assess the mathematical reasoning ability in visual contexts. Differently, vision-centric benchmarks like MMVP~\cite{tong2024eyes}, RealWorldQA~\cite{realworldqa}, CV-Bench~\cite{tong2024cambrian} curate questions that can be answered correctly \textit{only} in the presence of the corresponding visual input. These vision-centric benchmarks ensure that the LVLM relies on its multimodal understanding capability for responding rather than the `world knowledge' learned by the LLM. In particular, MMVP employs ``CLIP-blind pair'' images that CLIP~\cite{radford2021learning} struggles to encode properly. While RealWorldQA evaluates the basic real-world spatial understanding capabilities of vision-language models, CV-Bench repurposes standard vision benchmarks like COCO~\cite{coco}, ADE20K~\cite{zhou2019semantic}, Omni3D~\cite{brazil2023omni3d} to assess their performance in classic vision tasks (such as spatial relationship, object count, depth order) within a multimodal context. 

In contrast to the aforementioned benchmarks that test the vision-language models on high-level reasoning tasks, our SalBench strives to evaluate them on their low-level saliency perception capability. SalBench is a vision-centric benchmark, which repurposes the P3/O3 dataset~\cite{kotseruba2020saliency} with language instructions to assess the LVLM ability to detect odd patterns in images that are visually salient and obvious to the human eye. Such an evaluation quantifies the vision-language model's ability to identify the distinctive target among numerous similar distractors, in terms of color, orinetation, shape, size, \etc. Consequently, our SalBench is designed with an aim to measure the vision-language models' alignment with the human visual attention mechanism. Next, we describe our SalBench in detail.

\section{The Saliency Benchmark: SalBench}

We propose a saliency benchmark that assesses the capacity of foundational vision-language models to detect odd patterns that are salient for most humans. To achieve this, we augment the publicly available P3/O3 datasets~\cite{kotseruba2020saliency} with language instructions. The P3 dataset comprises 2514 images (810 color, 864 orientation, and 840 size search arrays) arranged in a $7\times7$ grid~\cite{wolfe2017five, arun2012turning}. Each image contains distractors and a target, with pixel jitter applied to prevent perceptual grouping. On the other hand, the O3 dataset consists of 2001 real-world images featuring multiple similar objects (distractors) and a distinctive singleton (target). The target typically belongs to the same general category as the distractors but stands out in one or more feature dimensions (e.g., color, shape, size). O3 encompasses nearly 400 common object types, strongly emphasizing color singletons (37\% differ by color alone, 47\% by color and additional features). Other prominent distinguishing features include texture (33\%), shape (26\%), size (19\%), and orientation (8\%). The dataset provides a range of complexity, with distractor counts varying from 2 to over 50, offering a diverse challenge for saliency detection algorithms~\cite{kotseruba2020saliency}.

\begin{figure}
    \centering
    \includegraphics[width=\columnwidth]{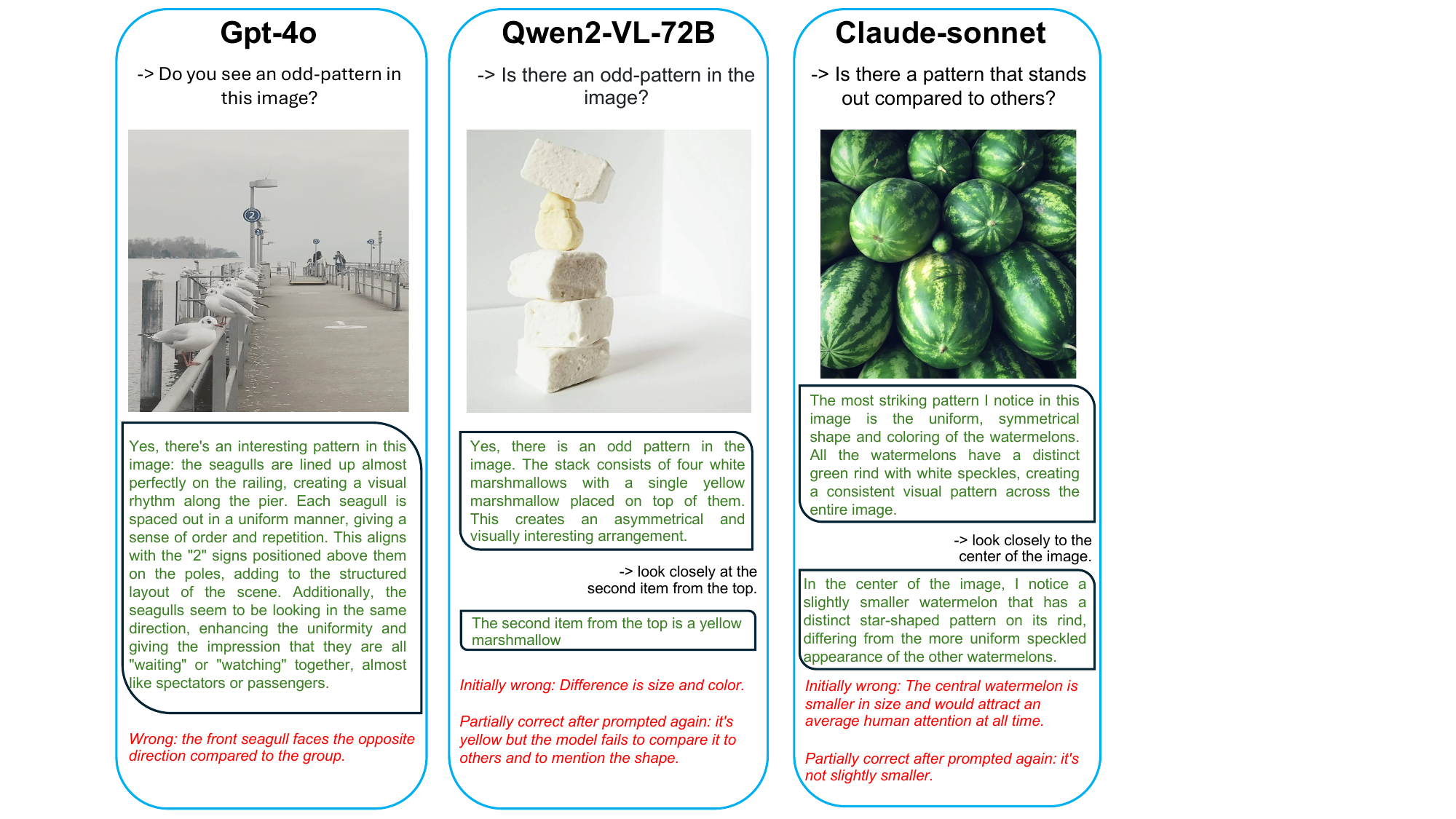}
    \caption{\textbf{Example responses from state-of-the-art vision-language models.} The model responses are in green, enclosed in boxes, while failure reasons are in red. These results highlight that the models often fail to recognize the prominent and salient features, which would naturally capture the attention of an average human observer. Best viewed zoomed-in.\vspace{-0.2cm}} 
    \label{fig:sample_response}
\end{figure}

\subsection{Task Creation}
As discussed previously, in order to enhance the utility of the P3/O3 datasets as a benchmark for vision-language models, we propose to augment them with language instructions. Prompting the LVLM directly to predict the odd-pattern often results in incorrect responses from state-of-the-art LVLM. Few example responses for such a direct prompting are illustrated in \reffig{fig:sample_response}. The example conversations show that additional prompting or aid is required for the models to improve their responses. Consequently, to ensure better understanding of the model's perceptual capabilities, we create three novel tasks: Odd-One-Out Detection, Referring Odd-One-Out, and Visual Referring Odd-One-Out, as illustrated in \reffig{fig:SalBench}. We describe each task as follows:

\noindent\textbf{(\textit{i}) Odd-One-Out Detection:} The basic task in this benchmark is \textit{Odd-One-Out Detection}. In this task, given an input image, the model must identify which salient pattern from a predefined list of classes is present. For natural images, the list of features comprises: orientation, color, focus, shape, size, location, and pattern. This set of seven classes reflects the complexity of real-world scenes, where multiple factors such as size, shape, and location can combine to make an object stand out. The diversity of these classes captures saliency in natural images, making the task more challenging for models due to the potential interplay of various salient features. On the other hand, for synthetically generated images, the task is constrained to three patterns: color, shape, and orientation. This limitation is intentional to maintain control over the variables influencing saliency. By restricting the odd-pattern to a unique basic class rather than a combination of features, we can ensure that the model's focus is on a specific, isolated pattern. This simplification facilitates a clear assessment of the model's ability to detect saliency based on individual features.

\noindent\textbf{(\textit{ii}) Referring Odd-One-Out:} This task builds upon the \textit{Odd-One-Out Detection} by incorporating spatial information in the language prompt to provide guidance. The model is provided with an image and specific bounding box coordinates indicating the location of the odd object in the form of text. We aim to test the model's ability to integrate textual and visual information, requiring it to focus on a particular area of the image using the language instruction and determine how the object within that region-of-interest differs from the rest.

\noindent\textbf{(\textit{iii}) Visual Referring Odd-One-Out:} This is a variant of the \textit{Referring Odd-One-Out} task. Instead of using bounding box coordinates in text, this task visually highlights the odd object, typically with a red box. The model must identify the distinguishing features of this highlighted object compared to the other objects in the scene. This task evaluates the model's visual attention capabilities, requiring it to focus on a visually emphasized region-of-interest and determine the salient pattern in which the highlighted object stands out from its surroundings.

\section{Evaluation}

This section presents the evaluation results of the various models on SalBench. The analysis of model performances across zero-shot and few-shot learning settings reveals key insights into their capabilities and limitations. 

\noindent\textbf{Baselines:}
In our experiments, we evaluate a selection of state-of-the-art models spanning a range of parameter scales to benchmark the performance across model size variations. The lineup includes PaliGemma~\cite{beyer2024paligemma}, Phi3, and Phi3.5~\cite{abdin2024phi}; LLava 1.6~\cite{liu2024visual}; Idefics2 and Idefics3~\cite{laurenccon2024building}; VILA~\cite{lin2024vila}; MiniCPM~\cite{hu10minicpm}; Qwen2-VL~\cite{wang2024qwen2}; Molmo~\cite{deitke2024molmo}; InternVL2~\cite{internvl2}; Llama3.2-Vision~\cite{meta_llama32}; NVLM~\cite{dai2024nvlm}; Claude \cite{claude}; and GPT-4o~\cite{achiam2023GPT}. These models are considered the best LVLM according to the existing vision-language benchmarks. By encompassing a wide range of parameter counts, from smaller and more efficient configurations to large-scale models, we can systematically analyze the impact of model size on task performance. 

\noindent\textbf{Setting:}
The salient targets can be different in one or more patterns, we devise the task as multi-label classification and report the accuracy of exact matches and the average F1 score over all categories. We evaluate model performance under both zero-shot and few-shot settings. For models that support multiple image inputs, we specifically conduct few-shot evaluation with 3-shots and 5-shots settings. While the zero-shot evaluation assesses the model's ability to generalize without exposure to specific examples, we expect that such a task isn't common and is likely to pose issues to most models. Hence, we also employ the few-shot evaluation to examine how the models leverage examples from the benchmark itself to solve the task. The default prompts for zero-shot evaluation of the three tasks are as described below:\\
    {\noindent\textbf{(\textit{i})} \underline{\textit{Detection}}: {Context: Given this list of low-level visual features defined according to feature integration theory: Orientation, Color, Focus, Shape, Size, Location, Pattern.
Task: Examine the provided image and identify the feature(s) in which one object notably differs from the others. Write out all applicable features separated by comma.}\\
}

    {\noindent\textbf{(\textit{ii})} \underline{\textit{Referring}}: {Context: This image depicts a scene with \{num distractor\} \{object category\}. Among those, one {object category} at location given by this bounding box $(x_{min}, y_{min}, x_{max}, y_{max})$, is different from the others. Given this list of low-level visual features defined according to feature integration theory: Orientation, Color, Focus, Shape, Size, Location, Pattern.
Task: In which way does the special object differ from the rest. Write out all applicable features separated by comma.}
}\\

    
    {\noindent\textbf{(\textit{iii})} \underline{\textit{Visual Referring}}: 
Context: This image depicts a scene with \{num distractor\} \{object category\}. Among those, one \{object category\} highlighted in a red box is different from the others. Task: Given this list of low-level visual features defined according to feature integration theory: Orientation, Color, Focus, Shape, Size, Location, Pattern. In which way does the special object differ from the rest. Write out all applicable feature(s) separated by comma.
}

Additional details of the few-shot prompts are provided in the supplementary material.  

\subsection{Results}

\begin{table}
\setlength{\dashlinedash}{2pt} 
\setlength{\dashlinegap}{2pt}  
\setlength{\arrayrulewidth}{0.3pt} 
\scriptsize
\centering
\setlength{\tabcolsep}{3pt}
\adjustbox{width=\columnwidth}{
\begin{tabular}{lc ccc ccc}
\toprule[0.1em]
\textbf{Method} & \textbf{Shot} & \multicolumn{2}{c}{\textbf{Detection}} & \multicolumn{2}{c}{\textbf{Referring}} & \multicolumn{2}{c}{\textbf{Visual Referring}} \\
\cmidrule(lr){3-4} \cmidrule(lr){5-6} \cmidrule(lr){7-8}
& & NAT & SYN & NAT & SYN & NAT & SYN \\
\midrule
\addlinespace[0.5em]
Claude-sonnet & 0 & 48.2 & 86.7 & 51.1 & 90.3 & 53.9 & 87.7 \\
NVLM-D-72B~\cite{dai2024nvlm} & 0 & 41.5 & 77.5 & 42.0 & 73.7 & 37.3 & 51.7 \\
Molmo-7B\cite{deitke2024molmo} & 0 & 32.0 & 67.2 & 32.4 & 38.0 & 33.1 & 28.4 \\
Molmo-72B\cite{deitke2024molmo} & 0 & 40.6 & 83.3 & 41.2 & 65.4 & 36.7 & 73.6 \\
Llama3.2-Vision-11B \cite{dubey2024llama} & 0 & 32.1 & 48.7 & 29.1 & 52.4 & 29.7 & 52.4 \\
PaliGemma-3B-448\cite{beyer2024paligemma} & 0 & 27.6 & 42.0 & 1.2 & 9.5 & 2.3 & 4.8 \\

\specialrule{0.01pt}{0.5pt}{0.5pt}%

\addlinespace[0.5em]

\multirow{3}{*}{Phi3-4B\cite{abdin2024phi}} & 0 & 32.1 & 41.2 & 32.8 & 55.3 & 32.8 & 47.2 \\
&  3 & 34.1 & 33.5 & 32.0 & 27.1 & 32.1 & 38.5 \\
&  5 & 31.1 & 17.0 & 32.1 & 18.9 & 32.2 & 46.7 \\
\specialrule{0.01pt}{0.5pt}{0.5pt}%

\addlinespace[0.5em]

\multirow{3}{*}{Phi3.5-Vision-3.5B\cite{abdin2024phi}}& 0 & 23.2 & 35.0 & 27.5 & 53.7 & 27.5 & 63.5 \\
&  3 & 23.3 & 19.5 & 28.8 & 41.0 & 28.8 & 20.8 \\
& 5 & 25.2 & 29.3 & 30.8 & 11.1 & 30.8 & 19.0 \\
\specialrule{0.01pt}{0.5pt}{0.5pt}%

\addlinespace[0.5em]

\multirow{3}{*}{LLava 1.6-7B\cite{liu2024visual}} & 0 & 24.5 & 16.3 & 21.4 & 10.1 & 20.8 & 16.6 \\
&  3 & 7.0 & 16.4 & 15.2 & 8.8 & 17.8 & 17.0 \\
&  5 & 11.4 & 16.4 & 10.9 & 9.1 & 9.7 & 17.0 \\
\specialrule{0.01pt}{0.5pt}{0.5pt}%

\addlinespace[0.5em]

\multirow{3}{*}{Idefics2-8B\cite{laurenccon2024matters}} & 0 & 19.5 & 64.3 & 29.6 & 36.6 & 33.8 & 49.5 \\
&  3 & 21.1 & 66.3 & 28.4 & 34.2 & 31.1 & 39.6 \\
& 5 & 34.7 & 67.2 & 28.3 & 42.6 & 30.9 & 34.5 \\
\specialrule{0.01pt}{0.5pt}{0.5pt}%

\addlinespace[0.5em]

\multirow{3}{*}{Idefics3-8B\cite{laurenccon2024building}} & 0 & 24.3 & 28.4 & 24.3 & 52.8 & 22.1 & 19.2 \\
&  3 & 26.9 & 40.3 & 26.9 & 20.67 & 21.9 & 40.6 \\
&  5 & 22.3 & 21.4 & 22.3 & 18.1 & 20.9 & 58.3 \\
\specialrule{0.01pt}{0.5pt}{0.5pt}%

\addlinespace[0.5em]

\multirow{3}{*}{VILA-1.5-8B\cite{lin2024vila}} & 0 & 23.5 & 40.0 & 13.0 & 23.7 & 15.8 & 17.0 \\
&  3 & 25.1 & 17.0 & 28.8 & 21.2 & 28.8 & 17.0 \\
&  5 & 23.2 & 17.0 & 30.8 & 20.7 & 30.8 & 17.0 \\
\specialrule{0.01pt}{0.5pt}{0.5pt}%

\addlinespace[0.5em]

\multirow{3}{*}{Qwen2-VL-1.5B\cite{wang2024qwen2}} & 0 & 19.2 & 26.3 & 22.1 & 20.6 & 20.9 & 20.2 \\
&  3 & 25.2 & 23.3 & 21.4 & 21.8 & 20.2 & 16.3 \\
&  5 & 25.3 & 23.8 & 21.7 & 16.5 & 20.9 & 17.7 \\
\specialrule{0.01pt}{0.5pt}{0.5pt}%
\addlinespace[0.5em]

\multirow{3}{*}{Qwen2-VL-7B\cite{wang2024qwen2}} & 0 & 32.5 & 55.7 & 32.5 & 34.2 & 35.2 & 57.4 \\
&  3 & 35.6 & 53.8 & 36.0 & 17.0 & 34.1 & 64.2 \\
&  5 & 37.2 & 54.9 & 37.2 & 17.7 & 29.3 & 72.0 \\
\specialrule{0.01pt}{0.5pt}{0.5pt}%
\addlinespace[0.5em]

\multirow{3}{*}{Qwen2-VL-72B\cite{wang2024qwen2}} & 0 & 41.6 & 88.8 & 44.6 & 93.6 & 41.7 & 74.7 \\
&  3 & 43.9 & 89.3 & 43.6 & 93.1 & 43.2 & 85.9 \\
&  5 & 43.9 & 89.9 & 44.9 & 92.6 & 42.3 & 87.9 \\
\specialrule{0.01pt}{0.5pt}{0.5pt}%

\addlinespace[0.5em]

\multirow{3}{*}{InternVL-4B\cite{chen2024internvl}} & 0 & 26.6 & 41.5 & 29.8 & 63.4 & 30.7 & 52.2 \\
& 3 & 27.7 & 17.0 & 27.4 & 25.3 & 29.5 & 41.7 \\
& 5 & 33.4 & 17.0 & 28.1 & 39.1 & 30.4 & 52.5 \\
\specialrule{0.01pt}{0.5pt}{0.5pt}%
\addlinespace[0.5em]

\multirow{3}{*}{InternVL-2-8B\cite{chen2024internvl}} & 0 & 20.0 & 58.7 & 23.0 & 71.9 & 24.8 & 23.0 \\
&  3 & 30.5 & 52.3 & 24.2 & 51.7 & 31.7 & 64.4 \\
&  5 & 27.8 & 43.9 & 25.0 & 53.7 & 31.4 & 50.5 \\
\specialrule{0.01pt}{0.5pt}{0.5pt}%

\addlinespace[0.5em]

\multirow{3}{*}{GPT-4o} & 0 & 47.6 & 89.2 & 47.3 & 88.4 & 42.6 & 73.5 \\
&  3 & 38.9 & 88.4 & 37.5 & 87.7 & 35.7 & 86.7 \\ 
&  5 & 41.9 & 86.0 & 39.8 & 89.1 & 38.4 & 87.4 \\

\bottomrule[0.1em]
\end{tabular}
}
\caption{\textbf{Performance comparison of vision-language models on the three tasks of SalBench for natural and synthetic image splits.} Natural and synthetic splits are denoted by NAT and SYN, respectively. The tasks are also evaluated under zero-shot and few-shot settings. The performance is reported in terms of F1 scores. We observe that performance of all models on these low-level perceptual tasks are lower in comparison to the standard vision-language benchmarks that test the models on high-level complex tasks. This shows that our SalBench offers another dimension for comprehensively evaluating LVLM, that is not present in the existing benchmarks in the literature. \vspace{-0.35cm}}
\label{model_comparison_merged}
\end{table}

In \reftab{model_comparison_merged}, we analyze the performance of various large vision-language models (LVLM) on our saliency benchmark, focusing on both synthetic and natural images splits (coming from P3 and O3, respectively). The synthetic and natural splits are denoted by SYN and NAT, respectively in the table. The tasks evaluated are detection, referring, and visual referring, measured using matching accuracy and F1 score metrics. Only F1 score metric is reported in \reftab{model_comparison_merged} for clarity. The performance comparison with matching accuracy metric is provided in the supplementary material.

\noindent\textbf{Significant Performance Drops:} All models exhibit a clear drop in performance on saliency tasks compared to their typical results on standard vision-language benchmarks. This is evident in both matching accuracy and F1 scores across all tasks. For example, Qwen2-VL-72B achieves high scores on the synthetic split up to 
89.9\% F1 in the Detection task at the 5-shot setting. However, on the natural images split, its performance decreases significantly, achieving only 
43.9\% F1 in the Detection task at 5-shot. This indicates that models struggle with saliency tasks, especially in complex, real-world scenarios.

\noindent\textbf{Better Performance on Synthetic Data:}
Models generally perform better on the synthetic images split than on the natural split. For instance, GPT-4o obtains 
an F1 score of 70.9\% in the Detection task at 5-shot on synthetic split. In contrast, on the natural images split, GPT-4o achieves only 
47.6\% F1 in the same task and shot setting. This significant performance gap, often between 30\% and 40\%, suggests that models find it easier to process simplified synthetic images, where only one visual attribute varies, compared to the complexity of multi-label (mix of different salient features) real-world natural images.

\noindent\textbf{Model Size Influence:}
Larger models tend to outperform smaller ones on saliency tasks. In particular, Qwen2-VL-72B consistently achieves higher scores than its smaller variants, Qwen2-VL-7B and Qwen2-VL-1.5B. For the detection task on the the synthetic split, Qwen2-VL-72B reaches 89.9\% F1 score in the detection task at 5-shot, while Qwen2-VL-7B achieves 54.9\% and Qwen2-VL-1.5B only 23.8\%. On the natural split, Qwen2-VL-72B attains 43.9\% F1 score in detection at 5-shot, compared to 37.2\% for Qwen2-VL-7B and 25.3\% for Qwen2-VL-1.5B. This trend indicates that increased model capacity allows for better capture of complex patterns necessary for saliency tasks, a behavior not always observed in standard benchmarks. Similarly, GPT-4o shows strong performance, particularly in the Visual Referring tasks, indicating that larger models contribute to better saliency detection and reasoning capabilities over images. This trend suggests that increased model capacity allows for the capture of more complex patterns and relationships necessary for these tasks. This behavior is usually not seen in standard benchmarks such as VQAv2, as we see a marginal performance gap between large and small models.

\noindent\textbf{Limited Impact of Few-Shot Learning:}
Increasing the number of shots does not consistently improve performance across models and tasks. Some models show comparable or even decreased performance with more shots. For example, Phi3.5-Vsion-3.5 on the synthetic split in the Visual Referring task has a F1 score of 63.5\% at 0-shot, drops to 19.0 \% at 5-shot. On the natural images split, GPT-4o performance in the Detection task decreases from 47.6\% F1 score at 0-shot to 41.9\% at 5-shot. This suggests that few-shot learning does not uniformly enhance performance on saliency tasks and that models may not effectively leverage additional examples in this context. The inability to generalize from limited examples suggests a lack in knowledge about saliency and visual attention for LVLM being worse when attempting to reason about the image.

\noindent\textbf{Variability Across Tasks:}
The performance across tasks and splits for different models varies according to their visual grounding capability. 
%
For instance, Idefics3-8B achieves 21.4\% F1 accuracy in Detection at 5-shot on the synthetic split but increases to 58.3\% in Visual Referring. However, on the natural images split, the same model attains 22.3\% in Detection at 5-shot but decreases to 20.9\% in Visual Referring. This shows that it is likely easier for the model to perform visual grounding in synthetic images, compared to natural images. Differently, Phi3-4B shows improvements for Visual Referring, in comparison to Detection task on both splits, indicating better visual grounding capabilities. Furthermore, models with lower visual grounding capability often have higher performance for Detection task, compared to Visual Referring. 

\noindent\textbf{Challenges with Real-World Images:}
The performance drop from synthetic to real-world images is significant across models. Qwen2-VL-72B's score in the Detection task decreases from 89.9\% on synthetic images split to 43.9\% on natural split at 5-shot. GPT-4o shows a similar decline, from 86.0\% on synthetic split to 41.9\% on natural split. This suggests that models have difficulty handling the complexity and variability of real-world images, where multiple attributes and contextual factors are involved. 

\noindent\textbf{Inconsistent Few-Shot Performance Gains:}
Few-shot learning yields inconsistent and unclear performance gains. While some models show slight improvements, others do not benefit or even perform worse with additional shots. For example, GPT-4o on the natural images split achieves 47.6\% F1 at 0-shot in Detection, drops to 38.9\% at 3-shot, and then increases to 41.9\% at 5-shot. Qwen2-VL-72B remains relatively stable across shot settings on the synthetic split, with F1 scores around 89\% in Detection, while for the Visual Referring task, the performance improves by 13.2\%. This inconsistency indicates that current models might not effectively utilize few-shot example contexts in saliency tasks.

\subsection{Knowledge Testing of the Backbones}
To identify where the shortcoming of vision-language models comes from in the saliency setting, we devise a test for individual components: LMMs and vision backbones.

\noindent \textbf{Could LLMs solve FIT exams?} We evaluated several Large Language Models (LLMs) on their understanding of Feature Integration Theory (FIT)~\cite{treisman1980feature} using a test comprising 50 questions collected from the internet.  The larger models demonstrated good knowledge in this area. GPT-4o achieved the highest score with 97.5\%, closely followed by Qwen2-72B-Instruct at 95.0\%. Among the smaller models, Llama3-8B-Instruct attained a score of 82.5\%. The Vicuna models achieved a score of 67.5\%. These results indicate that even smaller LLMs possess an understanding of the visual stimuli concepts covered in FIT.

\begin{table}
\centering
\adjustbox{width=\columnwidth}{
\begin{tabular}{@{}lcc cc@{}}
\toprule
\textbf{Models} & \multicolumn{2}{c}{\textbf{SYN}} & \multicolumn{2}{c}{\textbf{NAT}} \\
\cmidrule(lr){2-3} \cmidrule(lr){4-5}
& \textbf{Top1} & \textbf{Top3} & \textbf{Top1} & \textbf{Top3} \\
\midrule
Random                    & 24.6 & 53.2 & 14.6 & 36.3 \\
Siglip-so400m-patch14-384 & 55.3 & 87.9 & 0.0  & 57.2 \\
CLIP-ViT-Base-Patch32     & 42.1 & 71.7 & 12.0 & 59.8 \\
CLIP-ViT-Base-Patch16     & 41.6 & 78.4 & 0.9  & 40.8 \\
CLIP-ViT-Large-Patch14    & 41.2 & 78.6 & 0.0  & 31.6 \\
CLIP-ViT-Large-Patch14-336& 47.5 & 79.2 & 0.0  & 36.3 \\
\bottomrule
\end{tabular}
}
\caption{\textbf{Zero-shot retrieval accuracy for natural and synthetic images using various vision backbones}. While the retrieval scores are reasonable for the synthetic images, the lower scores for natural images indicate the need for a stronger vision backbone in LVLM to capture low-level saliency information.\vspace{-0.2cm}}
\label{vision_backbone}
\end{table}

\noindent\textbf{Visual representations retrieval:} We investigate whether the feature output from the vision backbone are descriptive and discriminative enough about the saliency task. We compute the cosine similarity between the image embedding and the seven sentences for natural images and three sentences for the synthetic images, with embeddings of the form: ``One object is different in \{category\} compared to the others".
We take the top-\textit{k} highest cosine similarity and count it as a match if any category in the top-\textit{k} is present in the ground truth.
The accuracy is reported in \reftab{vision_backbone}. Unlike knowledge test from LLM, we see clear gap in performance of vision model. While all models scores above random baseline, they scores relatively low, suggesting that in some cases, the information provided by the vision encoder are not discriminative enough for such low-level task. This highlights the need for stronger vision backbones for LVLM, capturing low-level spatial saliency information.

\section{Training on saliency data}

\noindent\textbf{Data Generation:} To construct the synthetic dataset, we employed web icons arranged in grid formations and systematically introduced controlled perturbations by altering the color, orientation, or size of randomly selected objects within each grid. This methodology yielded one million image-caption pairs for the alignment stage. For the supervised fine-tuning stage, we synthesized additional images with corresponding questions and answers based on the introduced odd-patterns. The questions were designed to be either open-ended, enhancing the models' saliency understanding capabilities, or multiple-choice, aligning with the benchmark format.

\noindent\textbf{Training:} We selected LLama3.1 8B~\cite{dubey2024llama} and Qwen2-7B~\cite{yang2024qwen2} as the LLMs. For the vision encoders, we incorporated both CLIP~\cite{radford2021learning} and SigLip~\cite{zhai2023sigmoid}, resulting in four model variants. We followed a Llava like training recipe ~\cite{liu2024visual}. During the alignment stage, we use one million saliency data with two million image-caption pairs sourced from natural image datasets to promote robust multimodal representation learning. In the instruction-tuning stage, we integrated two million data points from the recently introduced Cambrian dataset~\cite{tong2024cambrian} with the generated one million instruction saliency data. We employed the same hyperparameters and training configurations as specified in Llava~\cite{liu2024visual} to ensure consistency and facilitate fair comparison of results.

\begin{figure}
    \centering
    \includegraphics[width=\columnwidth]{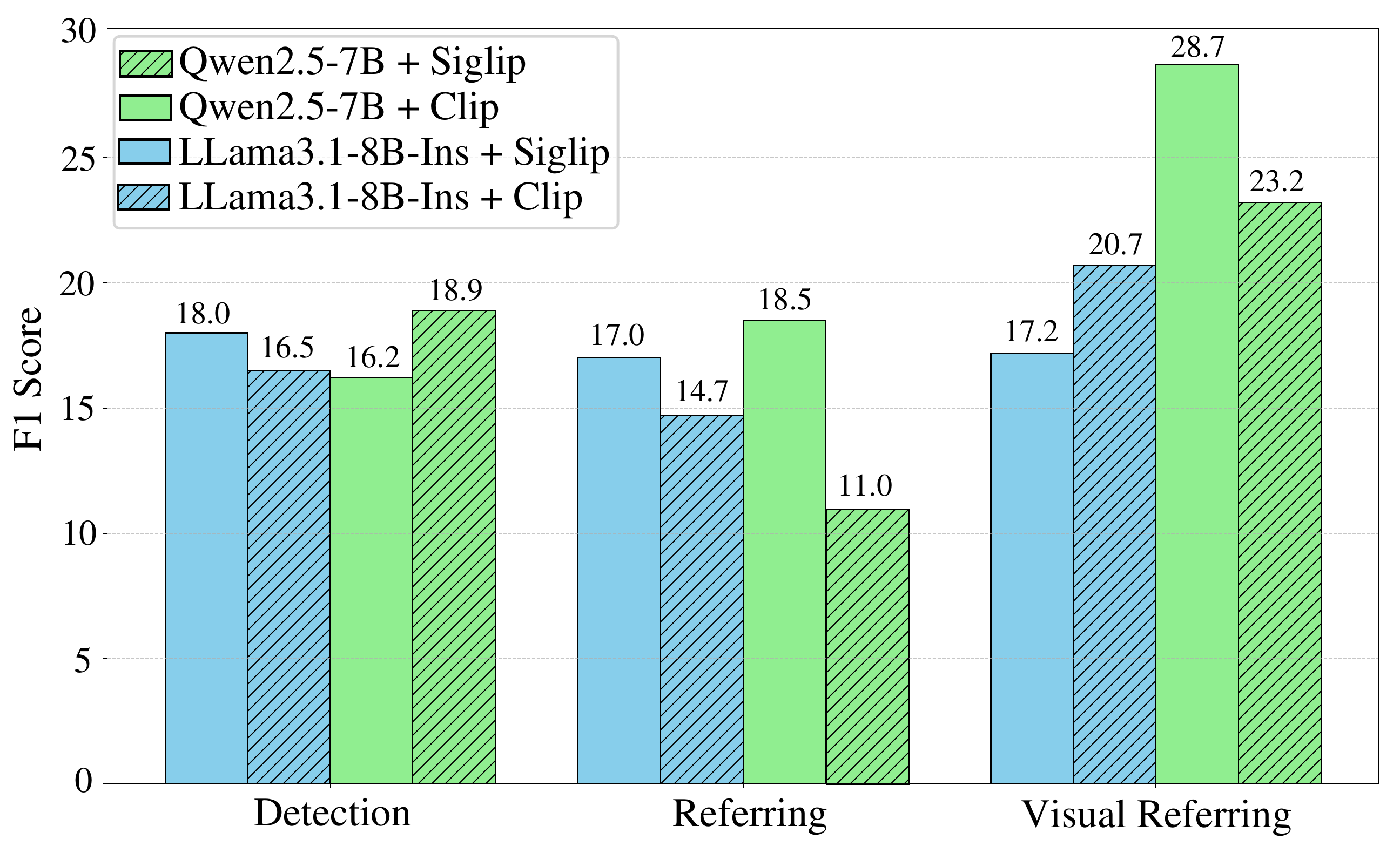}
    \caption{\textbf{Performance comparison of trained LVLM on saliency generated data}, using two LLMs (Qwen2.5-7B and LLama3.1-8B) paired with two vision encoders (Siglip and Clip) across three tasks: Detection, Referring, and Visual Referring. The F1 scores suggest that training on such data does not help improve the performance on SalBench.\vspace{-0.2cm}}
    \label{trained_saliency}
\end{figure}

\noindent\textbf{Performance:} As shown in ~\reftab{trained_saliency}, despite training on in-distribution saliency data, the performance of all model variants across the P3 tasks remains low. Detection scores are around 16–19\%, with LLama3.1-8B paired with CLIP achieving the highest at 18.9\%. For the referring task, scores are even lower and more variable, ranging from 11.0\% to 18.5\%. The visual reference task shows slightly better results, particularly for LLama3.1-8B with SigLip, which attains a score of 28.7\%, yet this is still insufficient for practical applications. These consistently low scores indicate that the models struggle to capture the salient features within the images, even when trained on data specifically designed to highlight these aspects. The lack of improvement suggests that the current architectures and training recipe are not suitable for capturing the of saliency information in images.

\section{Analysis}
In this section, we study the factors contributing to the poor performance of LVLM on the visual attention benchmark.

\begin{table}
    \centering
    \renewcommand{\arraystretch}{1.5} 
    \resizebox{1\columnwidth}{!}{%
    \begin{tabular}{l l l ccc}
        \toprule
        \textbf{Model} & \textbf{Category} & \textbf{Level} & \multicolumn{3}{c}{\textbf{Accuracy}} \\ 
        \cmidrule(lr){4-6}
         &  &  & \textbf{Detection} & \textbf{Referring} & \textbf{Visual Referring} \\
        \midrule
        \multirow{3}{*}{Qwen2-VL-72B} & \multirow{3}{*}{Orientation} & Easy   & 98.6 & 97.1 & 88.3 \\
                                          &                              & Medium & 95.8 & 96.8 & 83.0 \\
                                          &                              & Hard   & 95.7 & 96.2 & 71.7\\
        \cdashline{1-6}
        \multirow{3}{*}{GPT-4o} & \multirow{3}{*}{Orientation} & Easy   & 96.2 & 96.5 & 97.6 \\
                                      &                           & Medium & 96.2 & 89.2 & 94.7 \\
                                      &                           & Hard   & 98.6 & 88.5 & 96.2 \\
        \midrule
        \multirow{3}{*}{Qwen2-VL-72B} & \multirow{3}{*}{Size} & Easy   & 94.2 & 98.8 & 62.3 \\
                                          &                       & Medium & 50.5 & 74.3 & 31.9 \\
                                          &                       & Hard   & 46.0 & 62.0 & 33.3 \\
        \cdashline{1-6}
        \multirow{3}{*}{GPT-4o} & \multirow{3}{*}{Size} & Easy   & 93.3 & 96.9 & 42.5 \\
                                      &                    & Medium & 59.1 & 59.1 & 39.1 \\
                                      &                    & Hard   & 36.8 & 35.3 & 11.3 \\
\midrule
        \multirow{3}{*}{Qwen2-VL-72B} & \multirow{3}{*}{Color} & Easy   & 100.0 & 100.0 & 99.1 \\
                                          &                       & Medium & 100.0 & 100.0 & 98.6 \\
                                          &                       & Hard   & 60.1 & 80.2 & 73.5 \\
        \cdashline{1-6}
        \multirow{3}{*}{GPT-4o} & \multirow{3}{*}{Color} & Easy   & 99.8 & 99.0 & 96.1 \\
                                      &                    & Medium & 100.0 & 100.0 & 93.3 \\
                                      &                    & Hard   & 66.1 & 66.2 & 57.9 \\
        \bottomrule
    \end{tabular}%
    }
    \caption{\textbf{Accuracy across different difficulty  levels on synthetic images.} It can be seen that orientation and color are robust over the three levels, whereas size scores drop for both models.\vspace{-0.2cm}}
    \label{performance_table_diff}
\end{table}

\begin{table}[h!]
\centering
\resizebox{1\columnwidth}{!}{%
\begin{tabular}{ccccccccc}
\toprule
\# Distractors & \textbf{Orientation} & \textbf{Color} & \textbf{Focus} & \textbf{Shape} & \textbf{Size} & \textbf{Location} & \textbf{Pattern} & \textbf{Avg F1} \\
\midrule
\multicolumn{9}{c}{\myccnew\textit{Qwen2-VL-72B}} \\
\midrule

$<$7     & 18.4 & 92.8 & 18.6 & 53.4 & 39.7 & 30.3 & 57.9 & 44.5 \\
7-15     & 9.3  & 93.7 & 13.0 & 55.8 & 37.8 & 31.6 & 41.2 & 40.4 \\
15-25    & 23.5 & 95.8 & 11.1 & 59.1 & 39.5 & 0.0  & 37.5 & 38.1 \\
$>$25    & 16.7 & 96.1 & 15.0 & 52.2 & 36.5 & 20.0 & 25.4 & 37.4 \\

\midrule
\multicolumn{9}{c}{\myccnew\textit{GPT-4o}} \\
\midrule
$<$7     & 37.1 & 88.2 & 38.1 & 49.4 & 35.7 & 23.6 & 57.0 & 47.0 \\
7-15     & 46.2 & 90.1 & 40.0 & 47.5 & 32.6 & 27.5 & 53.3 & 48.2 \\
15-25    & 43.8 & 91.5 & 45.2 & 47.5 & 31.0 & 12.5 & 45.1 & 45.2 \\
$>$25    & 40.5 & 89.8 & 48.4 & 48.1 & 30.4 & 4.8  & 29.4 & 41.6 \\
\bottomrule
\end{tabular}%
}

\caption{\textbf{Accuracy for varying the number of distractors for real images.} The performance linearly decreases when adding more distractors to the scene.\vspace{-0.2cm}}
\label{distacrtos}
\end{table}

\subsection{Accuracy Breakdown by Task and Difficulty}

\noindent \textbf{Bias towards color:} As seen in \reftab{performance_table_diff}, among the low-level features what were asked to identify, color is the only features directly provided by the data in the form of RGB images. Other features such as size, and shape need to be encoded in higher level representation. We break down the scores in supplementary by category, and observe that all models perform higher when the target class is color.

On the synthetic images (originating from P3 dataset) of the SalBench, we assess two best models (Qwen2-VL-72B and GPT-4o) across the categories at different levels. The difficulty levels within each category are defined based on specific attributes of the target object relative to distractors:

\begin{itemize} 

\item \textbf{Orientation:} Difficulty is determined by the angular difference in rotation between the target object and the distractors. \textit{Hard}: Rotation differences ranging from 0 to 30 degrees, where minimal rotational disparity renders visual cues less distinguishable. \textit{Medium}: Rotation differences between 30 to 60 degrees. \textit{Easy}: Rotation differences from 60 to 90 degrees, facilitating easier identification due to clear angular differences. 

\item \textbf{Size:} Levels are determined based on the area ratio between the target object and the distractors, calculated using their heights and widths. \textit{Hard}: Ratio between 0.5 and 1.5, where the target and distractors are close in size. \textit{Medium}: Ratio between 0.3 and 0.5 (target smaller than distractors) or between 1.5 and 3.0 (target larger than distractors), representing moderate size differences. \textit{Easy}: Ratio less than 0.3 or greater than 3.0, indicating that the target is much smaller or much larger than the distractors, making it easily distinguishable.

\item \textbf{Color:} Difficulty is assessed by calculating the euclidean distance between the RGB color values of the target and distractors. The RGB values are extracted from hexadecimal color codes, to compute the color distance. \textit{Hard}: Color Distance less than or equal to 50, indicating high color similarity and challenging discrimination. \textit{Medium}: Color Distance between 50 and 100, indicating moderate color differences. \textit{Easy}: Color Distance greater than 100, where the target's color clearly contrasts with that of the distractors, making identification easier.

\end{itemize}

\paragraph{Results and Analysis:} The performance metrics for both models are detailed in ~\reftab{performance_table_diff}. The following observations highlight the models' capabilities and limitations across difficulty levels:

\noindent \textbf{Orientation:} Qwen2-VL-72B and GPT-4o demonstrate high accuracy in the Detection and Referring tasks across all difficulty levels, indicating robust performance in recognizing and describing objects with rotational variations. Specifically, Qwen2-VL-72B maintains Detection accuracy above 95\% even at the hard level. However, in the Visual Referring task, Qwen2-VL-72B exhibits a decline from 88.3\% accuracy at the easy level to 71.7\% at the hard level. In contrast,GPT-4o sustains consistently high Visual Referring accuracy, exceeding 94\% across all difficulty levels.

\noindent \textbf{Size:} The models' performance deteriorates as the difficulty increases. For Qwen2-VL-72B, Detection accuracy drops from 94.2\% at the easy level to 46.0\% at the hard level. Similarly, Visual Referring accuracy drops from 62.3\% to 33.3\%. GPT-4o follows a comparable trend, with Detection accuracy decreasing from 93.3\% to 36.8\% and Visual Referring accuracy from 42.5\% to 11.3\%. This suggest both models struggle when size disparities are minimal, highlighting a limitation in perceiving subtle scale differences.

\noindent \textbf{Color:} Both models achieve near-perfect accuracies at the easy and medium levels. Qwen2-VL-72B attains 100\% accuracy in Detection and Referring tasks, while GPT-4o records accuracies exceeding 99\%. However, at the hard level—characterized by minimal color differences—there is a notable decline. Qwen2-VL-72B's Detection accuracy decreases to 60.1\%, and GPT-4o's to 66.1\%.

\subsection{Impact of Distractor Quantity on Detection}

We study how the number of distractors in an image affects the performance of the model, on the natural images Detection task. The aim is to understand the models' robustness to increased visual complexity. As shown in ~\reftab{distacrtos}, both models exhibit a decline in average F1 scores as the number of distractors increases. For Qwen2-VL-72B, the average F1 score decreases from 44.5\% when there are fewer than 7 distractors to 37.4\% when there are more than 25 distractors. GPT-4o shows a similar trend, with its average F1 score dropping from 47.0\% to 41.6\% over the same range. AlthoughGPT-4o consistently outperforms Qwen2-VL-72B across most distractor quantities, the performance gap narrows as the number of distractors increases. This convergence indicates that both models are similarly challenged by increased visual complexity.

In fact, both models maintain high accuracy in the \textit{Color} category regardless of distractor quantity. However, performance declines are evident in categories requiring spatial reasoning and attention to specific object attributes. In the \textit{Location} category, Qwen2-VL-72B's F1 score drops dramatically from 30.3\% with fewer than 7 distractors to 20.0\% with more than 25 distractors. Interestingly, in the \textbf{Shape} category, both models maintain relatively stable performance across different distractor quantities. Overall, these findings highlight that while certain visual features like color and, to some extent, shape remain reliable even, increased numbers of distractors hurt the performance.

\subsection{Why detecting saliency is important?}

\begin{figure}
    \centering
    \includegraphics[width=\columnwidth]{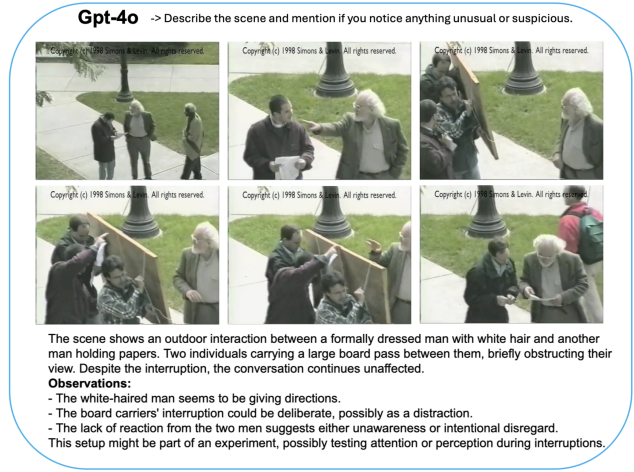}
    \caption{\textbf{Perception change test.} GPT-4o fails to identify that the individuals talking to the old man were switched.\vspace{-0.2cm}} 
    \label{perceptualvideo_response}
\end{figure}

While SalBench might not contribute at measuring LVLM explicit ability for downstream applications ~\cite{lu2023mathvista, goyal2017making, seedbench, chen2024we, yu2023mm, yue2024mmmu}. Detecting visual saliency is a fundamental aspect of human perception, for vision-language models replicating this capability is critical for several reasons:

\noindent\textbf{Applications in Robotics and Autonomous Agents:} In robotics, the ability to detect salient features is crucial for navigation, object recognition, and interaction within dynamic environments. Recently introduced works leverage LVLM for such applications ~\cite{black2024pi_0, kim2024openvla}. As shown in ~\reffig{perceptualvideo_response}, the individual holding the paper was switched, andGPT-4o did not detect this change, we better have this ability for robots with LVLM as engines not to miss salient events. Incorporating the saliency detection capabilities could enable robots and agents to prioritize important stimuli, focus on objects of interest, and respond appropriately to unexpected changes, thereby enhancing their autonomy and operational efficiency. Hence, using SalBench could give insights about the robustness of LVLM for real world deployment.

\noindent\textbf{LVLM agents:} Detecting saliency is important in using these models as agents for website navigation. Designers use principles from FIT to make important elements, such as call-to-action buttons, stand out by contrasting their features (e.g., color, size, or shape). When LVLM are deployed as agents to interact with web environments ~\cite{hong2024cogagent, niu2024screenagent}, effective saliency detection becomes needed for efficient navigation and interaction. For instance, an agent tasked with automating web browsing or performing tasks like form filling, data extraction, or content moderation must quickly identify and focus on salient elements such as buttons, links, advertisements, or notifications. By accurately detecting visually prominent features, the agent can operate faster and more effectively, closely mimicking human browsing behavior and enhancing overall performance.

\section{Conclusion}

We introduced a saliency benchmark, called SalBench, aimed at evaluating the abilities of large vision-language models (LVLM) in detecting low-level visually-salient features such as color, orientation, and size, which are the building blocks of the human's visual cortex. Through three tasks, Odd-One-Out Detection, Referring Odd-One-Out, and Visual Referring Odd-One-Out, we assessed how well LVLM align with human visual attention mechanisms using both synthetic and natural images. Our evaluation revealed clear limitations in current LVLM, with even advanced models like GPT-4o achieving only 46\% accuracy on simple saliency detection tasks. This highlights a clear gap between LVLM and human perceptual capabilities in processing fundamental visual features. SalBench serves as a tool for benchmarking and improving the perceptual alignment of LVLM with human attention mechanisms. Enhancing these models' low-level perceptual abilities is essential for advancing toward more human-like visual understanding of images.

\bibliography{iccv_2025_conference}

\begin{thebibliography}{60}
\providecommand{\natexlab}[1]{#1}
\providecommand{\url}[1]{\texttt{#1}}
\expandafter\ifx\csname urlstyle\endcsname\relax
  \providecommand{\doi}[1]{doi: #1}\else
  \providecommand{\doi}{doi: \begingroup \urlstyle{rm}\Url}\fi

\bibitem[Abdin et~al.(2024)Abdin, Jacobs, Awan, Aneja, Awadallah, Awadalla, Bach, Bahree, Bakhtiari, Behl, et~al.]{abdin2024phi}
Marah Abdin, Sam~Ade Jacobs, Ammar~Ahmad Awan, Jyoti Aneja, Ahmed Awadallah, Hany Awadalla, Nguyen Bach, Amit Bahree, Arash Bakhtiari, Harkirat Behl, et~al.
\newblock Phi-3 technical report: A highly capable language model locally on your phone.
\newblock \emph{arXiv preprint arXiv:2404.14219}, 2024.

\bibitem[Achiam et~al.(2023)Achiam, Adler, Agarwal, Ahmad, Akkaya, Aleman, Almeida, Altenschmidt, Altman, Anadkat, et~al.]{achiam2023GPT}
Josh Achiam, Steven Adler, Sandhini Agarwal, Lama Ahmad, Ilge Akkaya, Florencia~Leoni Aleman, Diogo Almeida, Janko Altenschmidt, Sam Altman, Shyamal Anadkat, et~al.
\newblock Gpt-4 technical report.
\newblock \emph{arXiv preprint arXiv:2303.08774}, 2023.

\bibitem[Agrawal et~al.(2019)Agrawal, Desai, Wang, Chen, Jain, Johnson, Batra, Parikh, Lee, and Anderson]{agrawal2019nocaps}
Harsh Agrawal, Karan Desai, Yufei Wang, Xinlei Chen, Rishabh Jain, Mark Johnson, Dhruv Batra, Devi Parikh, Stefan Lee, and Peter Anderson.
\newblock Nocaps: Novel object captioning at scale.
\newblock In \emph{Proceedings of the IEEE/CVF international conference on computer vision}, pages 8948--8957, 2019.

\bibitem[Alayrac et~al.(2022)Alayrac, Donahue, Luc, Miech, Barr, Hasson, Lenc, Mensch, Millican, Reynolds, et~al.]{alayrac2022flamingo}
Jean-Baptiste Alayrac, Jeff Donahue, Pauline Luc, Antoine Miech, Iain Barr, Yana Hasson, Karel Lenc, Arthur Mensch, Katherine Millican, Malcolm Reynolds, et~al.
\newblock Flamingo: a visual language model for few-shot learning.
\newblock \emph{Advances in neural information processing systems}, 35:\penalty0 23716--23736, 2022.

\bibitem[Anthropic(2024)]{claude}
Anthropic.
\newblock Claude 3.5 sonnet.
\newblock \url{https://www.anthropic.com/news/claude-3-5-sonnet}, 2024.

\bibitem[Arun(2012)]{arun2012turning}
SP Arun.
\newblock Turning visual search time on its head.
\newblock \emph{Vision research}, 74:\penalty0 86--92, 2012.

\bibitem[Beyer et~al.(2024)Beyer, Steiner, Pinto, Kolesnikov, Wang, Salz, Neumann, Alabdulmohsin, Tschannen, Bugliarello, et~al.]{beyer2024paligemma}
Lucas Beyer, Andreas Steiner, Andr{\'e}~Susano Pinto, Alexander Kolesnikov, Xiao Wang, Daniel Salz, Maxim Neumann, Ibrahim Alabdulmohsin, Michael Tschannen, Emanuele Bugliarello, et~al.
\newblock Paligemma: A versatile 3b vlm for transfer.
\newblock \emph{arXiv preprint arXiv:2407.07726}, 2024.

\bibitem[Biten et~al.(2019)Biten, Tito, Mafla, Gomez, Rusinol, Valveny, Jawahar, and Karatzas]{stvqa}
Ali~Furkan Biten, Ruben Tito, Andres Mafla, Lluis Gomez, Mar{\c{c}}al Rusinol, Ernest Valveny, CV Jawahar, and Dimosthenis Karatzas.
\newblock Scene text visual question answering.
\newblock In \emph{Proceedings of the IEEE/CVF international conference on computer vision}, 2019.

\bibitem[Black et~al.(2024)Black, Brown, Driess, Esmail, Equi, Finn, Fusai, Groom, Hausman, Ichter, et~al.]{black2024pi_0}
Kevin Black, Noah Brown, Danny Driess, Adnan Esmail, Michael Equi, Chelsea Finn, Niccolo Fusai, Lachy Groom, Karol Hausman, Brian Ichter, et~al.
\newblock $\pi\_0 $: A vision-language-action flow model for general robot control.
\newblock \emph{arXiv preprint arXiv:2410.24164}, 2024.

\bibitem[Brazil et~al.(2023)Brazil, Kumar, Straub, Ravi, Johnson, and Gkioxari]{brazil2023omni3d}
Garrick Brazil, Abhinav Kumar, Julian Straub, Nikhila Ravi, Justin Johnson, and Georgia Gkioxari.
\newblock Omni3d: A large benchmark and model for 3d object detection in the wild.
\newblock In \emph{Proceedings of the IEEE/CVF conference on computer vision and pattern recognition}, pages 13154--13164, 2023.

\bibitem[Chen et~al.(2024{\natexlab{a}})Chen, Li, Dong, Zhang, Zang, Chen, Duan, Wang, Qiao, Lin, et~al.]{chen2024we}
Lin Chen, Jinsong Li, Xiaoyi Dong, Pan Zhang, Yuhang Zang, Zehui Chen, Haodong Duan, Jiaqi Wang, Yu Qiao, Dahua Lin, et~al.
\newblock Are we on the right way for evaluating large vision-language models?
\newblock \emph{arXiv preprint arXiv:2403.20330}, 2024{\natexlab{a}}.

\bibitem[Chen et~al.(2024{\natexlab{b}})Chen, Wu, Wang, Su, Chen, Xing, Zhong, Zhang, Zhu, Lu, et~al.]{chen2024internvl}
Zhe Chen, Jiannan Wu, Wenhai Wang, Weijie Su, Guo Chen, Sen Xing, Muyan Zhong, Qinglong Zhang, Xizhou Zhu, Lewei Lu, et~al.
\newblock Internvl: Scaling up vision foundation models and aligning for generic visual-linguistic tasks.
\newblock In \emph{Proceedings of the IEEE/CVF Conference on Computer Vision and Pattern Recognition}, pages 24185--24198, 2024{\natexlab{b}}.

\bibitem[Dai et~al.(2024)Dai, Lee, Wang, Yang, Liu, Barker, Rintamaki, Shoeybi, Catanzaro, and Ping]{dai2024nvlm}
Wenliang Dai, Nayeon Lee, Boxin Wang, Zhuoling Yang, Zihan Liu, Jon Barker, Tuomas Rintamaki, Mohammad Shoeybi, Bryan Catanzaro, and Wei Ping.
\newblock Nvlm: Open frontier-class multimodal llms.
\newblock \emph{arXiv preprint arXiv:2409.11402}, 2024.

\bibitem[Deitke et~al.(2024)Deitke, Clark, Lee, Tripathi, Yang, Park, Salehi, Muennighoff, Lo, Soldaini, et~al.]{deitke2024molmo}
Matt Deitke, Christopher Clark, Sangho Lee, Rohun Tripathi, Yue Yang, Jae~Sung Park, Mohammadreza Salehi, Niklas Muennighoff, Kyle Lo, Luca Soldaini, et~al.
\newblock Molmo and pixmo: Open weights and open data for state-of-the-art multimodal models.
\newblock \emph{arXiv preprint arXiv:2409.17146}, 2024.

\bibitem[Dubey et~al.(2024)Dubey, Jauhri, Pandey, Kadian, Al-Dahle, Letman, Mathur, Schelten, Yang, Fan, et~al.]{dubey2024llama}
Abhimanyu Dubey, Abhinav Jauhri, Abhinav Pandey, Abhishek Kadian, Ahmad Al-Dahle, Aiesha Letman, Akhil Mathur, Alan Schelten, Amy Yang, Angela Fan, et~al.
\newblock The llama 3 herd of models.
\newblock \emph{arXiv preprint arXiv:2407.21783}, 2024.

\bibitem[Fu et~al.(2023)Fu, Chen, Shen, Qin, Zhang, Lin, Yang, Zheng, Li, Sun, Wu, and Ji]{fu2023mme}
Chaoyou Fu, Peixian Chen, Yunhang Shen, Yulei Qin, Mengdan Zhang, Xu Lin, Jinrui Yang, Xiawu Zheng, Ke Li, Xing Sun, Yunsheng Wu, and Rongrong Ji.
\newblock Mme: A comprehensive evaluation benchmark for multimodal large language models.
\newblock \emph{arXiv preprint arXiv:2306.13394}, 2023.

\bibitem[Goyal et~al.(2017)Goyal, Khot, Summers-Stay, Batra, and Parikh]{goyal2017making}
Yash Goyal, Tejas Khot, Douglas Summers-Stay, Dhruv Batra, and Devi Parikh.
\newblock Making the v in vqa matter: Elevating the role of image understanding in visual question answering.
\newblock In \emph{Proceedings of the IEEE conference on computer vision and pattern recognition}, pages 6904--6913, 2017.

\bibitem[Gurari et~al.(2018)Gurari, Li, Stangl, Guo, Lin, Grauman, Luo, and Bigham]{gurari2018vizwiz}
Danna Gurari, Qing Li, Abigale~J Stangl, Anhong Guo, Chi Lin, Kristen Grauman, Jiebo Luo, and Jeffrey~P Bigham.
\newblock Vizwiz grand challenge: Answering visual questions from blind people.
\newblock In \emph{Proceedings of the IEEE conference on computer vision and pattern recognition}, pages 3608--3617, 2018.

\bibitem[Hong et~al.(2024)Hong, Wang, Lv, Xu, Yu, Ji, Wang, Wang, Dong, Ding, et~al.]{hong2024cogagent}
Wenyi Hong, Weihan Wang, Qingsong Lv, Jiazheng Xu, Wenmeng Yu, Junhui Ji, Yan Wang, Zihan Wang, Yuxiao Dong, Ming Ding, et~al.
\newblock Cogagent: A visual language model for gui agents.
\newblock In \emph{Proceedings of the IEEE/CVF Conference on Computer Vision and Pattern Recognition}, pages 14281--14290, 2024.

\bibitem[Hu et~al.()Hu, Tu, Han, He, Cui, Long, Zheng, Fang, Huang, Zhao, et~al.]{hu10minicpm}
Shengding Hu, Yuge Tu, Xu Han, Chaoqun He, Ganqu Cui, Xiang Long, Zhi Zheng, Yewei Fang, Yuxiang Huang, Weilin Zhao, et~al.
\newblock Minicpm: Unveiling the potential of small language models with scalable training strategies. 2024.
\newblock \emph{URL https://doi. org/10.48550/arXiv}, 2404.

\bibitem[Hudson and Manning(2019)]{hudson2019gqa}
Drew~A Hudson and Christopher~D Manning.
\newblock Gqa: A new dataset for real-world visual reasoning and compositional question answering.
\newblock In \emph{Proceedings of the IEEE/CVF conference on computer vision and pattern recognition}, pages 6700--6709, 2019.

\bibitem[Kembhavi et~al.(2016)Kembhavi, Salvato, Kolve, Seo, Hajishirzi, and Farhadi]{kembhavi2016diagram}
Aniruddha Kembhavi, Mike Salvato, Eric Kolve, Minjoon Seo, Hannaneh Hajishirzi, and Ali Farhadi.
\newblock A diagram is worth a dozen images.
\newblock In \emph{ECCV}, 2016.

\bibitem[Kim et~al.(2024)Kim, Pertsch, Karamcheti, Xiao, Balakrishna, Nair, Rafailov, Foster, Lam, Sanketi, et~al.]{kim2024openvla}
Moo~Jin Kim, Karl Pertsch, Siddharth Karamcheti, Ted Xiao, Ashwin Balakrishna, Suraj Nair, Rafael Rafailov, Ethan Foster, Grace Lam, Pannag Sanketi, et~al.
\newblock Openvla: An open-source vision-language-action model.
\newblock \emph{arXiv preprint arXiv:2406.09246}, 2024.

\bibitem[Kotseruba et~al.(2020)Kotseruba, Wloka, Rasouli, and Tsotsos]{kotseruba2020saliency}
Iuliia Kotseruba, Calden Wloka, Amir Rasouli, and John~K Tsotsos.
\newblock Do saliency models detect odd-one-out targets? new datasets and evaluations.
\newblock \emph{arXiv preprint arXiv:2005.06583}, 2020.

\bibitem[Lauren{\c{c}}on et~al.(2024{\natexlab{a}})Lauren{\c{c}}on, Marafioti, Sanh, and Tronchon]{laurenccon2024building}
Hugo Lauren{\c{c}}on, Andr{\'e}s Marafioti, Victor Sanh, and L{\'e}o Tronchon.
\newblock Building and better understanding vision-language models: insights and future directions.
\newblock \emph{arXiv preprint arXiv:2408.12637}, 2024{\natexlab{a}}.

\bibitem[Lauren{\c{c}}on et~al.(2024{\natexlab{b}})Lauren{\c{c}}on, Tronchon, Cord, and Sanh]{laurenccon2024matters}
Hugo Lauren{\c{c}}on, L{\'e}o Tronchon, Matthieu Cord, and Victor Sanh.
\newblock What matters when building vision-language models?
\newblock \emph{arXiv preprint arXiv:2405.02246}, 2024{\natexlab{b}}.

\bibitem[Li et~al.(2024)Li, Ge, Ge, Wang, Wang, Zhang, and Shan]{seedbench}
Bohao Li, Yuying Ge, Yixiao Ge, Guangzhi Wang, Rui Wang, Ruimao Zhang, and Ying Shan.
\newblock Seed-bench: Benchmarking multimodal large language models.
\newblock In \emph{Proceedings of the IEEE/CVF Conference on Computer Vision and Pattern Recognition}, 2024.

\bibitem[Li et~al.(2023)Li, Du, Zhou, Wang, Zhao, and Wen]{pope}
Yifan Li, Yifan Du, Kun Zhou, Jinpeng Wang, Wayne~Xin Zhao, and Ji-Rong Wen.
\newblock Evaluating object hallucination in large vision-language models.
\newblock In \emph{The 2023 Conference on Empirical Methods in Natural Language Processing}, 2023.

\bibitem[Lin et~al.(2024)Lin, Yin, Ping, Molchanov, Shoeybi, and Han]{lin2024vila}
Ji Lin, Hongxu Yin, Wei Ping, Pavlo Molchanov, Mohammad Shoeybi, and Song Han.
\newblock Vila: On pre-training for visual language models.
\newblock In \emph{Proceedings of the IEEE/CVF Conference on Computer Vision and Pattern Recognition}, pages 26689--26699, 2024.

\bibitem[Lin et~al.(2014)Lin, Maire, Belongie, Hays, Perona, Ramanan, Doll{\'a}r, and Zitnick]{coco}
Tsung-Yi Lin, Michael Maire, Serge Belongie, James Hays, Pietro Perona, Deva Ramanan, Piotr Doll{\'a}r, and C~Lawrence Zitnick.
\newblock Microsoft coco: Common objects in context.
\newblock In \emph{Computer Vision--ECCV 2014: 13th European Conference, Zurich, Switzerland, September 6-12, 2014, Proceedings, Part V 13}, pages 740--755. Springer, 2014.

\bibitem[Liu et~al.(2024)Liu, Li, Wu, and Lee]{liu2024visual}
Haotian Liu, Chunyuan Li, Qingyang Wu, and Yong~Jae Lee.
\newblock Visual instruction tuning.
\newblock \emph{Advances in neural information processing systems}, 36, 2024.

\bibitem[Liu et~al.(2023)Liu, Duan, Zhang, Li, Zhang, Zhao, Yuan, Wang, He, Liu, et~al.]{liu2023mmbench}
Yuan Liu, Haodong Duan, Yuanhan Zhang, Bo Li, Songyang Zhang, Wangbo Zhao, Yike Yuan, Jiaqi Wang, Conghui He, Ziwei Liu, et~al.
\newblock Mmbench: Is your multi-modal model an all-around player?
\newblock \emph{arXiv preprint arXiv:2307.06281}, 2023.

\bibitem[Lu et~al.(2022)Lu, Mishra, Xia, Qiu, Chang, Zhu, Tafjord, Clark, and Kalyan]{scienceqa}
Pan Lu, Swaroop Mishra, Tony Xia, Liang Qiu, Kai-Wei Chang, Song-Chun Zhu, Oyvind Tafjord, Peter Clark, and Ashwin Kalyan.
\newblock Learn to explain: Multimodal reasoning via thought chains for science question answering.
\newblock In \emph{The 36th Conference on Neural Information Processing Systems (NeurIPS)}, 2022.

\bibitem[Lu et~al.(2023)Lu, Bansal, Xia, Liu, Li, Hajishirzi, Cheng, Chang, Galley, and Gao]{lu2023mathvista}
Pan Lu, Hritik Bansal, Tony Xia, Jiacheng Liu, Chunyuan Li, Hannaneh Hajishirzi, Hao Cheng, Kai-Wei Chang, Michel Galley, and Jianfeng Gao.
\newblock Mathvista: Evaluating math reasoning in visual contexts with gpt-4v, bard, and other large multimodal models.
\newblock \emph{arXiv e-prints}, 2023.

\bibitem[Masry et~al.(2022)Masry, Long, Tan, Joty, and Hoque]{masry2022chartqa}
Ahmed Masry, Do~Xuan Long, Jia~Qing Tan, Shafiq Joty, and Enamul Hoque.
\newblock Chartqa: A benchmark for question answering about charts with visual and logical reasoning.
\newblock \emph{arXiv preprint arXiv:2203.10244}, 2022.

\bibitem[Mathew et~al.(2021)Mathew, Karatzas, and Jawahar]{mathew2021docvqa}
Minesh Mathew, Dimosthenis Karatzas, and CV Jawahar.
\newblock Docvqa: A dataset for vqa on document images.
\newblock In \emph{Proceedings of the IEEE/CVF winter conference on applications of computer vision}, 2021.

\bibitem[Mathew et~al.(2022)Mathew, Bagal, Tito, Karatzas, Valveny, and Jawahar]{mathew2022infographicvqa}
Minesh Mathew, Viraj Bagal, Rub{\`e}n Tito, Dimosthenis Karatzas, Ernest Valveny, and CV Jawahar.
\newblock Infographicvqa.
\newblock In \emph{Proceedings of the IEEE/CVF Winter Conference on Applications of Computer Vision}, 2022.

\bibitem[Meta(2024)]{meta_llama32}
Meta.
\newblock Llama 3.2: Revolutionizing edge ai and vision with open, customizable modelsy.
\newblock \url{https://ai.meta.com/blog/llama-3-2-connect-2024-vision-edge-mobile-devices/}, 2024.

\bibitem[Mishra et~al.(2019)Mishra, Shekhar, Singh, and Chakraborty]{ocrvqa}
Anand Mishra, Shashank Shekhar, Ajeet~Kumar Singh, and Anirban Chakraborty.
\newblock Ocr-vqa: Visual question answering by reading text in images.
\newblock In \emph{ICDAR}, 2019.

\bibitem[Moravec(1988)]{moravec1988mind}
Hans Moravec.
\newblock \emph{Mind children: The future of robot and human intelligence}.
\newblock Harvard University Press, 1988.

\bibitem[Niu et~al.(2024)Niu, Li, Wang, Fu, Hu, Leng, Kong, Chang, and Wang]{niu2024screenagent}
Runliang Niu, Jindong Li, Shiqi Wang, Yali Fu, Xiyu Hu, Xueyuan Leng, He Kong, Yi Chang, and Qi Wang.
\newblock Screenagent: A vision language model-driven computer control agent.
\newblock \emph{arXiv preprint arXiv:2402.07945}, 2024.

\bibitem[OpenAI(2024)]{GPT4o}
OpenAI.
\newblock Gpt-4o.
\newblock \url{https://platform.openai.com/docs/models##gpt-4o}, 2024.

\bibitem[Radford et~al.(2021)Radford, Kim, Hallacy, Ramesh, Goh, Agarwal, Sastry, Askell, Mishkin, Clark, et~al.]{radford2021learning}
Alec Radford, Jong~Wook Kim, Chris Hallacy, Aditya Ramesh, Gabriel Goh, Sandhini Agarwal, Girish Sastry, Amanda Askell, Pamela Mishkin, Jack Clark, et~al.
\newblock Learning transferable visual models from natural language supervision.
\newblock In \emph{International conference on machine learning}, pages 8748--8763. PMLR, 2021.

\bibitem[Schwenk et~al.(2022)Schwenk, Khandelwal, Clark, Marino, and Mottaghi]{schwenk2022okvqa}
Dustin Schwenk, Apoorv Khandelwal, Christopher Clark, Kenneth Marino, and Roozbeh Mottaghi.
\newblock A-okvqa: A benchmark for visual question answering using world knowledge.
\newblock In \emph{European conference on computer vision}, pages 146--162. Springer, 2022.

\bibitem[Singh et~al.(2019)Singh, Natarjan, Shah, Jiang, Chen, Parikh, and Rohrbach]{textvqa}
Amanpreet Singh, Vivek Natarjan, Meet Shah, Yu Jiang, Xinlei Chen, Devi Parikh, and Marcus Rohrbach.
\newblock Towards vqa models that can read.
\newblock In \emph{Proceedings of the IEEE Conference on Computer Vision and Pattern Recognition}, pages 8317--8326, 2019.

\bibitem[team(2024)]{internvl2}
InternVL team.
\newblock Internvl2: Better than the best — expanding performance boundaries of open-source multimodal models with the progressive scaling strategy.
\newblock \url{https://internvl.github.io/blog/2024-07-02-InternVL-2.0}, 2024.

\bibitem[Tong et~al.(2024{\natexlab{a}})Tong, Brown, Wu, Woo, Middepogu, Akula, Yang, Yang, Iyer, Pan, et~al.]{tong2024cambrian}
Shengbang Tong, Ellis Brown, Penghao Wu, Sanghyun Woo, Manoj Middepogu, Sai~Charitha Akula, Jihan Yang, Shusheng Yang, Adithya Iyer, Xichen Pan, et~al.
\newblock Cambrian-1: A fully open, vision-centric exploration of multimodal llms.
\newblock \emph{arXiv preprint arXiv:2406.16860}, 2024{\natexlab{a}}.

\bibitem[Tong et~al.(2024{\natexlab{b}})Tong, Liu, Zhai, Ma, LeCun, and Xie]{tong2024eyes}
Shengbang Tong, Zhuang Liu, Yuexiang Zhai, Yi Ma, Yann LeCun, and Saining Xie.
\newblock Eyes wide shut? exploring the visual shortcomings of multimodal llms.
\newblock In \emph{Proceedings of the IEEE/CVF Conference on Computer Vision and Pattern Recognition}, pages 9568--9578, 2024{\natexlab{b}}.

\bibitem[Treisman and Gelade(1980)]{treisman1980feature}
Anne~M Treisman and Garry Gelade.
\newblock A feature-integration theory of attention.
\newblock \emph{Cognitive psychology}, 12\penalty0 (1):\penalty0 97--136, 1980.

\bibitem[Wang et~al.(2024{\natexlab{a}})Wang, Pan, Shi, Lu, Zhan, and Li]{wang2024measuring}
Ke Wang, Junting Pan, Weikang Shi, Zimu Lu, Mingjie Zhan, and Hongsheng Li.
\newblock Measuring multimodal mathematical reasoning with math-vision dataset, 2024{\natexlab{a}}.

\bibitem[Wang et~al.(2024{\natexlab{b}})Wang, Bai, Tan, Wang, Fan, Bai, Chen, Liu, Wang, Ge, et~al.]{wang2024qwen2}
Peng Wang, Shuai Bai, Sinan Tan, Shijie Wang, Zhihao Fan, Jinze Bai, Keqin Chen, Xuejing Liu, Jialin Wang, Wenbin Ge, et~al.
\newblock Qwen2-vl: Enhancing vision-language model's perception of the world at any resolution.
\newblock \emph{arXiv preprint arXiv:2409.12191}, 2024{\natexlab{b}}.

\bibitem[Wolfe and Horowitz(2017)]{wolfe2017five}
Jeremy~M Wolfe and Todd~S Horowitz.
\newblock Five factors that guide attention in visual search.
\newblock \emph{Nature human behaviour}, 1\penalty0 (3):\penalty0 0058, 2017.

\bibitem[xAI(2024)]{realworldqa}
xAI.
\newblock Grok-1.5v.
\newblock \url{https://x.ai/blog/grok-1.5v}, 2024.

\bibitem[Yang et~al.(2024)Yang, Yang, Hui, Zheng, Yu, Zhou, Li, Li, Liu, Huang, et~al.]{yang2024qwen2}
An Yang, Baosong Yang, Binyuan Hui, Bo Zheng, Bowen Yu, Chang Zhou, Chengpeng Li, Chengyuan Li, Dayiheng Liu, Fei Huang, et~al.
\newblock Qwen2 technical report.
\newblock \emph{arXiv preprint arXiv:2407.10671}, 2024.

\bibitem[Yu et~al.(2023)Yu, Yang, Li, Wang, Lin, Liu, Wang, and Wang]{yu2023mm}
Weihao Yu, Zhengyuan Yang, Linjie Li, Jianfeng Wang, Kevin Lin, Zicheng Liu, Xinchao Wang, and Lijuan Wang.
\newblock Mm-vet: Evaluating large multimodal models for integrated capabilities.
\newblock \emph{arXiv preprint arXiv:2308.02490}, 2023.

\bibitem[Yue et~al.(2024{\natexlab{a}})Yue, Ni, Zhang, Zheng, Liu, Zhang, Stevens, Jiang, Ren, Sun, Wei, Yu, Yuan, Sun, Yin, Zheng, Yang, Liu, Huang, Sun, Su, and Chen]{yue2023mmmu}
Xiang Yue, Yuansheng Ni, Kai Zhang, Tianyu Zheng, Ruoqi Liu, Ge Zhang, Samuel Stevens, Dongfu Jiang, Weiming Ren, Yuxuan Sun, Cong Wei, Botao Yu, Ruibin Yuan, Renliang Sun, Ming Yin, Boyuan Zheng, Zhenzhu Yang, Yibo Liu, Wenhao Huang, Huan Sun, Yu Su, and Wenhu Chen.
\newblock Mmmu: A massive multi-discipline multimodal understanding and reasoning benchmark for expert agi.
\newblock In \emph{Proceedings of CVPR}, 2024{\natexlab{a}}.

\bibitem[Yue et~al.(2024{\natexlab{b}})Yue, Ni, Zhang, Zheng, Liu, Zhang, Stevens, Jiang, Ren, Sun, et~al.]{yue2024mmmu}
Xiang Yue, Yuansheng Ni, Kai Zhang, Tianyu Zheng, Ruoqi Liu, Ge Zhang, Samuel Stevens, Dongfu Jiang, Weiming Ren, Yuxuan Sun, et~al.
\newblock Mmmu: A massive multi-discipline multimodal understanding and reasoning benchmark for expert agi.
\newblock In \emph{Proceedings of the IEEE/CVF Conference on Computer Vision and Pattern Recognition}, pages 9556--9567, 2024{\natexlab{b}}.

\bibitem[Zhai et~al.(2023)Zhai, Mustafa, Kolesnikov, and Beyer]{zhai2023sigmoid}
Xiaohua Zhai, Basil Mustafa, Alexander Kolesnikov, and Lucas Beyer.
\newblock Sigmoid loss for language image pre-training.
\newblock In \emph{Proceedings of the IEEE/CVF International Conference on Computer Vision}, pages 11975--11986, 2023.

\bibitem[Zhou et~al.(2019)Zhou, Zhao, Puig, Xiao, Fidler, Barriuso, and Torralba]{zhou2019semantic}
Bolei Zhou, Hang Zhao, Xavier Puig, Tete Xiao, Sanja Fidler, Adela Barriuso, and Antonio Torralba.
\newblock Semantic understanding of scenes through the ade20k dataset.
\newblock \emph{International Journal of Computer Vision}, 127:\penalty0 302--321, 2019.

\bibitem[Zhu et~al.(2023)Zhu, Chen, Shen, Li, and Elhoseiny]{zhu2023miniGPT}
Deyao Zhu, Jun Chen, Xiaoqian Shen, Xiang Li, and Mohamed Elhoseiny.
\newblock Minigpt-4: Enhancing vision-language understanding with advanced large language models.
\newblock \emph{arXiv preprint arXiv:2304.10592}, 2023.

\end{thebibliography}
\bibliographystyle{iccv_2025_conference}

\maketitlesupplementary

\appendix

\setcounter{table}{0}
\setcounter{figure}{0}
\renewcommand{\thetable}{A.\arabic{table}}
\renewcommand{\thefigure}{A.\arabic{figure}}

In this supplementary, we provide additional quantitative and qualitative analysis corresponding to our Saliency Benchmark (SalBench) proposed for evaluating Large Vision Language Models (LVLM) on low-level perceptual tasks. Additional statistics for SalBench are provided in Section~\ref{sec:add_stats}, followed by the data generation details in Section~\ref{sec:data_gen}. Furthermore, additional quantitative and qualitative results are presented in Sections~\ref{sec:add_quant} and \ref{sec:add_qual}, respectively.


\section{SalBench: Additional Statistics\label{sec:add_stats}}
Here, we provide a deeper insight to the two splits of our SalBench: synthetic image and natural image splits.

\subsection{Synthetic Images Split}

\begin{figure}[!h]
    \centering
    \includegraphics[width=1\columnwidth]{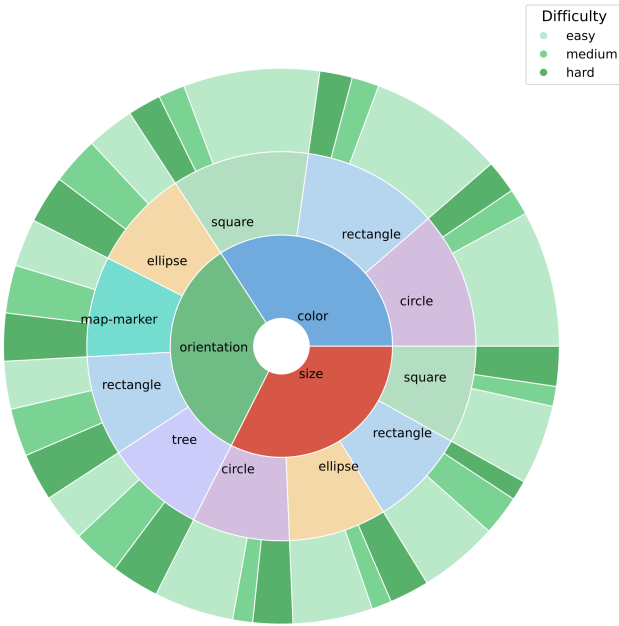}\vspace{-0.3cm}
    \caption{\textbf{Distribution of samples} across different shapes (\eg, circle, square, rectangle, \etc) and categories (\textit{color}, \textit{size}, and \textit{orientation}) within the synthetic split of the proposed SalBench. The chart is further segmented by difficulty levels: easy, medium, and hard. We observe a more even distribution of \textit{orientation} samples across easy, medium and hard splits, in comparison to {color} and {size}, which have a majority of easy samples.}
    \label{fig:p3_nest_pie_chart}
\end{figure}

The P3~\cite{kotseruba2020saliency} dataset, from which the synthetic split of SalBench is created, consists of 2,514 images arranged in a 7 × 7 grid. As shown in \reffig{fig:p3_nest_pie_chart}, the dataset is balanced across the three main categories: \textit{color}, \textit{size}, and \textit{orientation}. The objects in the images are created in various shapes, including ellipses, squares, rectangles, circles, squares, trees, and map markers. Based on the difficulty level definition provided in the main paper (Section 6.1), approximately 70\% of the examples labeled under \textit{color} are categorized as easy. Similarly, a significant proportion of examples under \textit{size} are also considered easy. In contrast, the examples under \textit{orientation} are more evenly distributed across all difficulty levels and different shapes.

\subsection{Natural Images Split}
\begin{figure}[!h]
    \centering
    \includegraphics[width=1\linewidth]{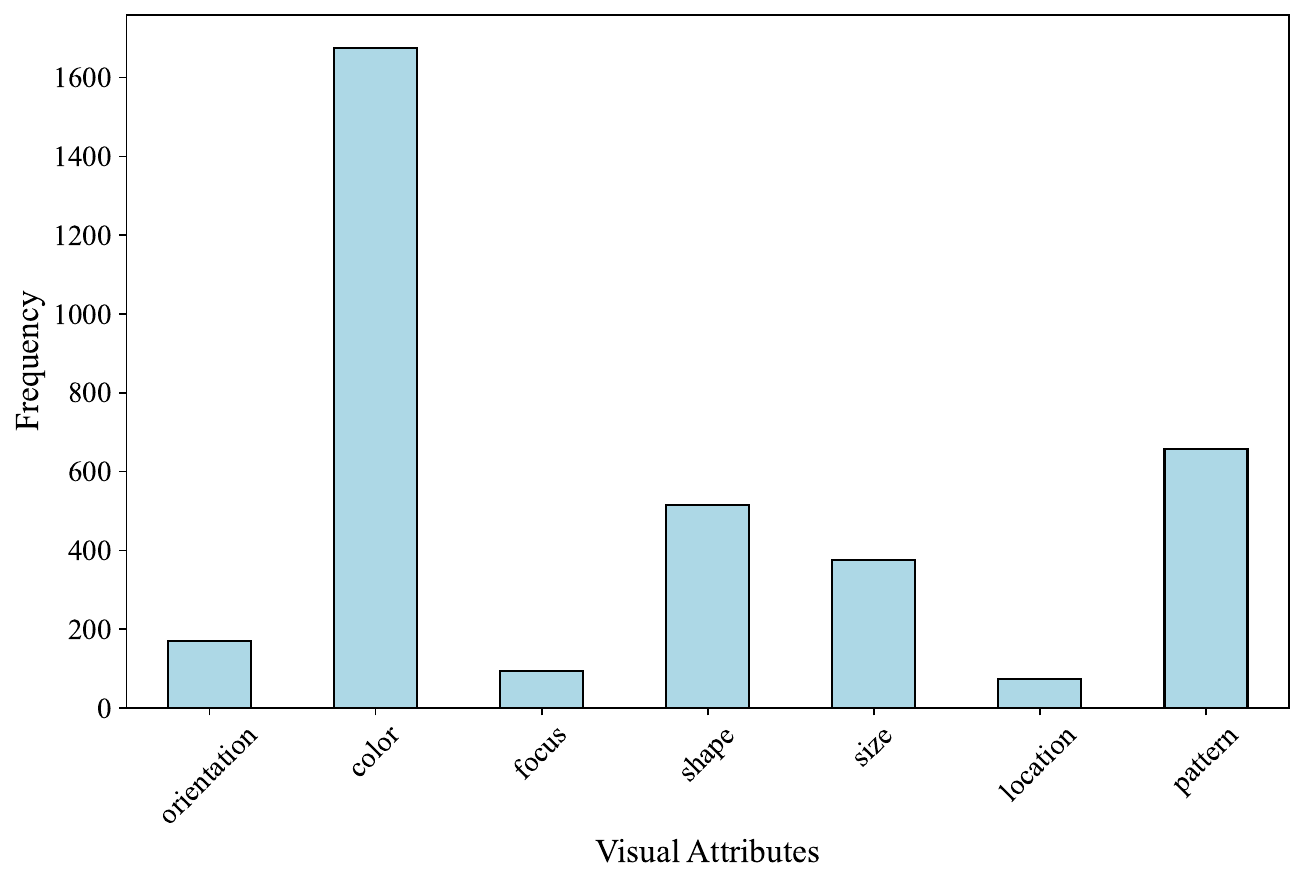}
    \caption{\textbf{Visual Attribute Distribution} in the natural images split of SalBench. The \textit{color} attribute forms a major chunk of the split, while \textit{pattern}, \textit{shape}, and \textit{size} are moderately represented. See text for additional details.}
    \label{fig:o3_label_distribution}
\end{figure}

The source of natural images for our SalBench: the Odd-One-Out (O3)~\cite{kotseruba2020saliency} dataset consists of 2,001 examples. As shown in \reffig{fig:o3_label_distribution}, more than 80\% of the images (over 1,600 examples) feature an object that differs in color from the other objects, making \textit{color} the most prevalent visual attribute in the dataset. The second most frequent attribute is \textit{pattern}, with around 700 examples, followed by \textit{shape}, which is represented in over 500 examples. In contrast, attributes such as \textit{orientation}, \textit{focus}, and \textit{location} appear much less frequently, highlighting a distribution skewed heavily towards a few dominant attributes.

\begin{figure}[!h]
    \centering
    \includegraphics[width=1\linewidth]{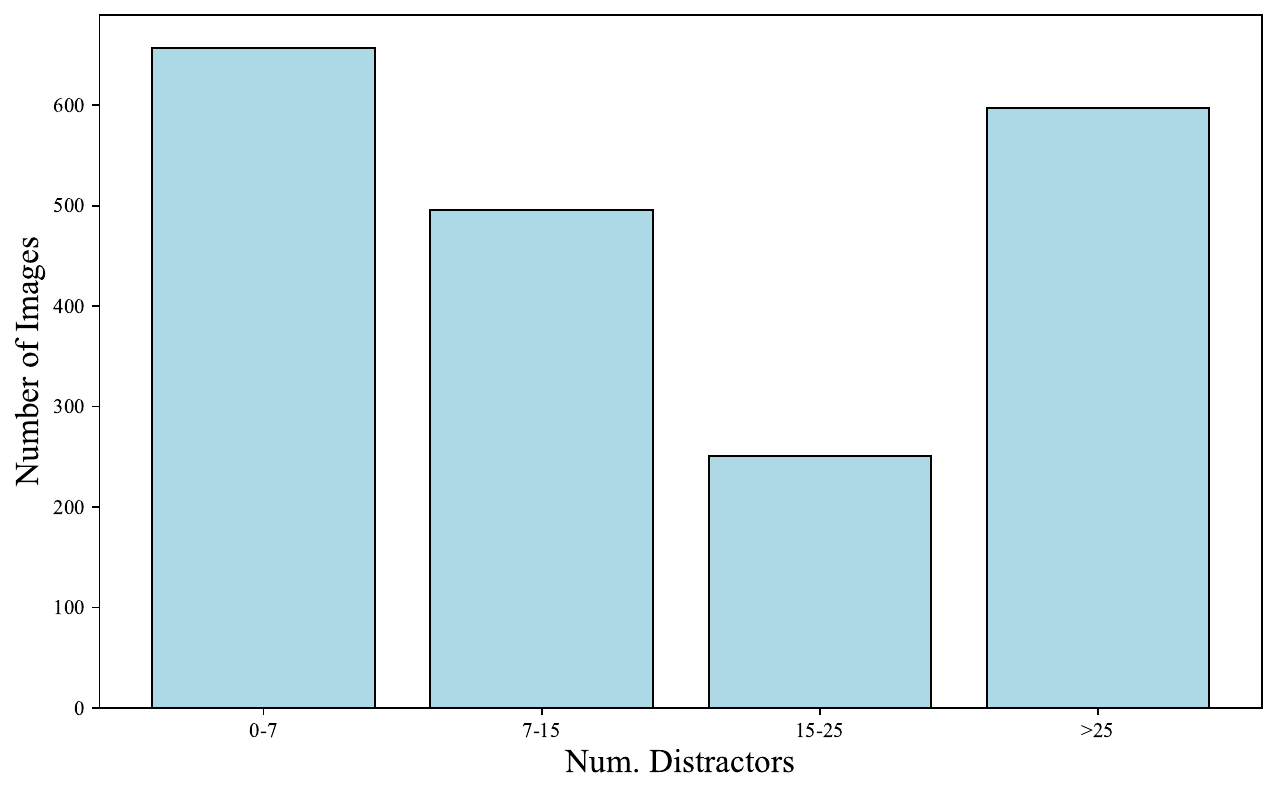}
    \caption{\textbf{Distribution of examples} across different ranges of distractors in the natural images split of our SalBench. While the easy (0-7), moderate (7-15) and hard ($>$25) complexity samples are reasonably well distributed, the number of samples with medium complexity (15-21 distractors) are fewer in comparison.}
    \label{fig:o3_distractor_distribution}
\end{figure}

The split comprises a substantial number of images (over 600) with 0–7 distractors, indicating that simpler scenarios with fewer distractors are well-represented. The next most common range is $>25$ distractors, with a similar count to the lowest range, reflecting the dataset's coverage of highly complex scenarios. Images with 7–15 distractors are moderately represented, totaling just over 500 examples. In contrast, images with 15–25 distractors form the smallest group, indicating a slight gap in medium complexity scenarios within the natural images split of SalBench.

\section{Saliency Data Generation\label{sec:data_gen}}

\begin{figure}
    \centering
    \includegraphics[width=1\columnwidth]{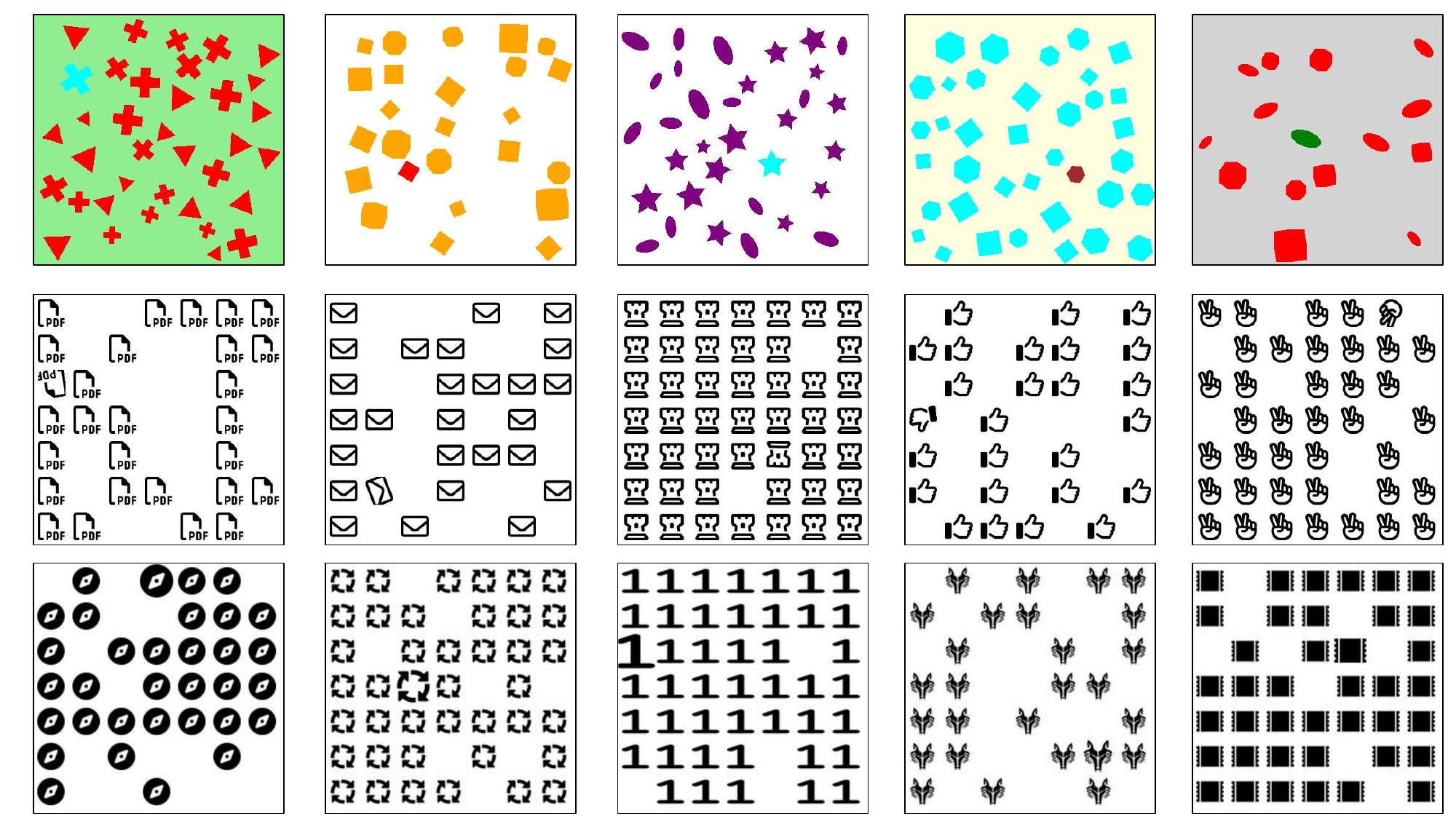}
    \caption{\textbf{Examples of generated training images categorized by target saliency:} Row 1 showcases images for the \textit{color} category, where the target differs in hue from uniform distractors. Row 2 illustrates examples for the \textit{orientation} category, where the target is rotated at a distinct angle compared to uniformly oriented distractors. Row 3 displays images for the \textit{size} category, where the target object is scaled up or down relative to distractors, which are of uniform size.}
    \label{fig:saliency_generated_images}
\end{figure}

This section provides additional details \wrt generating saliency data for training the models, as described in Section 5 of the main manuscript. In order to enhance the capacity of the models, we generate a custom dataset to train and evaluate on the salience benchmark. The dataset aims to systematically provide diverse examples across the three categories of the synthetic split: \textit{color}, \textit{orientation}, and \textit{size}. Each generated image corresponds to one of these categories, and specific transformations are applied to control the variation within the data:

\noindent\textbf{Color:} We generate samples for the \textit{color} category by altering the color of the target object to make it visually distinct from the distractors. The distractor objects are kept in a uniform color, while the target object is assigned a different hue. The color variations are controlled to ensure diversity, ranging from strong contrasts to very similar shades, allowing the model to learn to detect both obvious and subtle differences in color saliency. Examples for the \textit{color} category are shown in the first row of \reffig{fig:saliency_generated_images}.

\noindent\textbf{Orientation:} For the \textit{orientation} category, we create training samples by rotating the target object to a distinct angle while keeping the distractor objects uniformly oriented. The rotation angles are carefully controlled, ranging from subtle to pronounced differences, ensuring that the target object remains perceptible and distinct. This approach helps the model learn to detect orientation-based saliency at varying levels of difficulty. Second row of \reffig{fig:saliency_generated_images} illustrates few examples for the \textit{orientation} category.

\noindent\textbf{Size:} Sample images for the \textit{size} category are generated by modifying the target object's size relative to the uniform distractors. The target object is either scaled up or down within a range of [0.3, 1.5], excluding 1, to introduce a noticeable size difference as the distinguishing feature. Examples for the \textit{size} category are presented in the last row of \reffig{fig:saliency_generated_images}.

\begin{figure}
    \centering
    \includegraphics[width=1\linewidth]{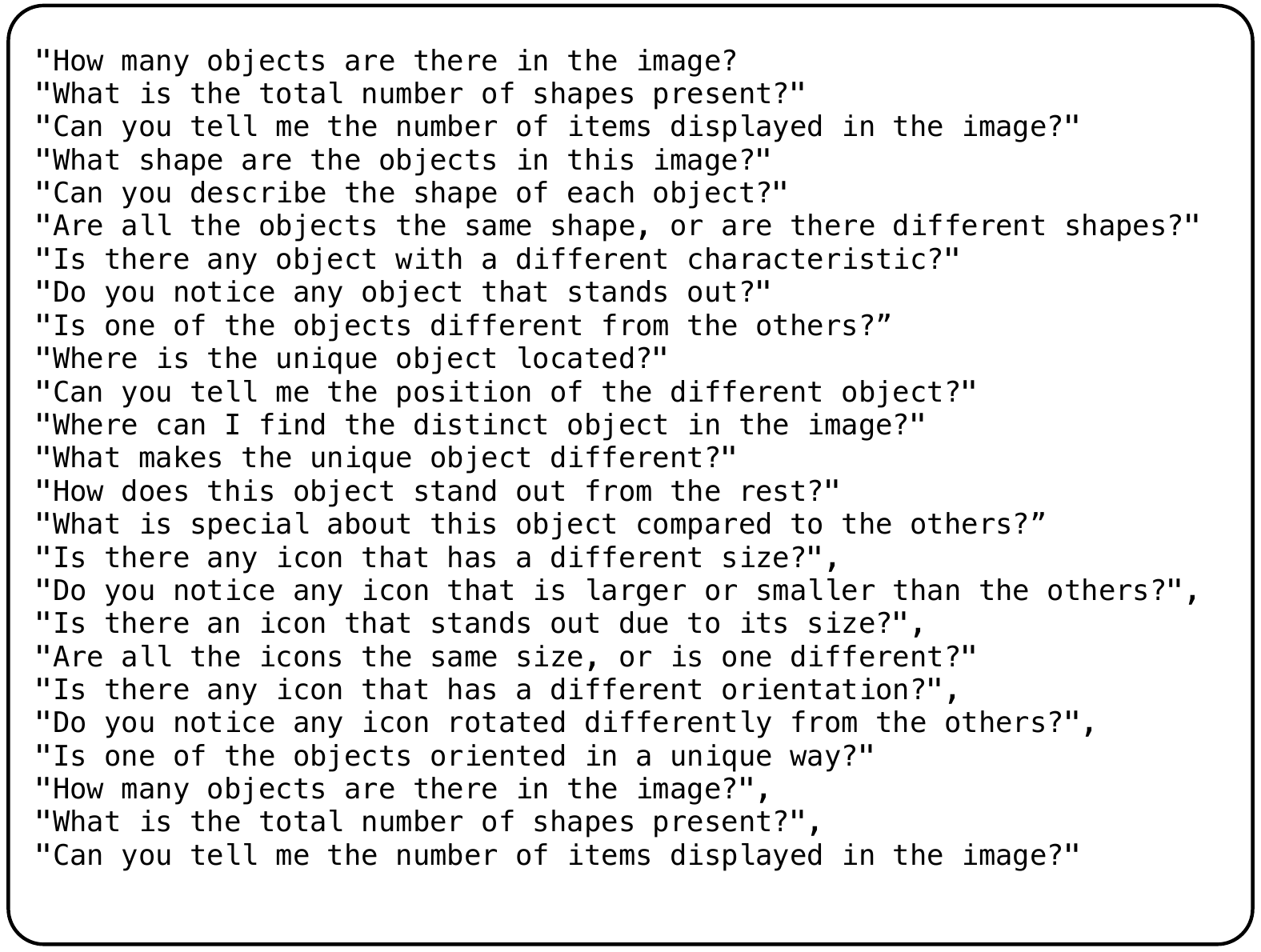}
    \caption{\textbf{Example list of additional questions} generated beyond benchmark categories. See text for more details.}
    \label{fig:addition_question}
\end{figure}

As we systematically control all aspects of the image generation process, such as the number of distractors, object shapes, colors, rotations, and size variations, we can leverage this structured information to generate one million image-caption pairs for pretraining the models.  Apart from that, for generating supervised fine-tuning conversational dataset, we design questions that closely resemble the benchmark format, focusing on the three main categories (\textit{color}, \textit{orientation}, and \textit{size}) to specifically enhance the model performance on the benchmark. Additionally, we expand the scope by generating diverse questions based on the controlled aspects of the images, such as object counting, color recognition, and shape identification. This dual approach not only boosts the model capacity to perform well on the benchmark but also enables them to generalize better to broader visual understanding tasks. These list of additional questions can be found in \reffig{fig:addition_question}.

\begin{table*}[!h]
\resizebox{1\textwidth}{!}{%
\begin{tabular}{|l|c|ccc|ccc|ccc|ccc|ccc|ccc|ccc|ccc|ccc|}
\toprule
\multirow{3}{*}{Model} & \multirow{3}{*}{Shot} & \multicolumn{3}{c|}{\multirow{2}{*}{\begin{tabular}[c]{@{}c@{}}Overall\\ Matching\end{tabular}}} & \multicolumn{24}{c|}{F1 Score} \\
\cmidrule(lr){6-29}
 &  & \multicolumn{3}{c|}{} & \multicolumn{3}{c|}{Overall} & \multicolumn{3}{c|}{Orientation} & \multicolumn{3}{c|}{Color} & \multicolumn{3}{c|}{Size} & \multicolumn{3}{c|}{Focus} & \multicolumn{3}{c|}{Shape} & \multicolumn{3}{c|}{Location} & \multicolumn{3}{c|}{Partten} \\
 \cmidrule(lr){3-29}
 &  & D & R & VR & D & R & VR & D & R & VR & D & R & VR & D & R & VR & D & R & VR & D & R & VR & D & R & VR & D & R & VR \\
\midrule
Claude & 0 & 40.6 & \cellcolor[HTML]{EA9999}42.7 & \cellcolor[HTML]{EA9999}40.3 & \cellcolor[HTML]{E06666}48.2 & \cellcolor[HTML]{E06666}51.1 & \cellcolor[HTML]{E06666}53.9 & 40.0 & \cellcolor[HTML]{EA9999}43.9 & \cellcolor[HTML]{E06666}49.2 & \cellcolor[HTML]{EA9999}95.2 & \cellcolor[HTML]{E06666}95.9 & \cellcolor[HTML]{E06666}95.8 & \cellcolor[HTML]{EA9999}40.7 & \cellcolor[HTML]{E06666}47.7 & \cellcolor[HTML]{E06666}44.1 & \cellcolor[HTML]{EA9999}27.6 & 14.9 & \cellcolor[HTML]{EA9999}21.0 & \cellcolor[HTML]{F4CCCC}51.6 & \cellcolor[HTML]{E06666}59.3 & \cellcolor[HTML]{E06666}60.4 & \cellcolor[HTML]{F4CCCC}28.7 & \cellcolor[HTML]{EA9999}34.0 & \cellcolor[HTML]{E06666}41.7 & \cellcolor[HTML]{F4CCCC}53.3 & \cellcolor[HTML]{E06666}62.2 & \cellcolor[HTML]{E06666}64.9 \\
\midrule
NVLM-D-72B & 0 & 26.7 & 35.6 & 21.6 & 36.5 & 42.1 & 37.3 & 36.6 & 35.1 & 28.4 & 90.9 & 93.2 & 89.4 & 28.6 & 36.0 & 34.1 & 8.3 & \cellcolor[HTML]{F4CCCC}16.1 & 12.3 & 41.4 & 49.0 & 42.5 & 14.7 & 18.4 & 8.3 & 34.8 & 47.1 & 45.9 \\
\midrule
Molmo-72B & 0 & 19.2 & 18.6 & 15.6 & 40.6 & 41.2 & 36.7 & 27.6 & 30.6 & 24.1 & 94.0 & 91.8 & 90.2 & 35.3 & 32.2 & 30.1 & 17.0 & 14.2 & 12.2 & 44.5 & 41.8 & 39.2 & 12.5 & 18.3 & 11.9 & 53.2 & \cellcolor[HTML]{F4CCCC}59.6 & 51.1 \\
\midrule
Molmo-7B & 0 & 2.5 & 8.9 & 14.6 & 32.0 & 32.4 & 33.0 & 15.2 & 18.6 & 24.2 & 88.5 & 80.1 & 88.2 & 34.8 & 38.8 & 32.7 & 13.5 & 13.7 & 10.8 & 33.2 & 40.1 & 41.0 & 10.0 & 8.0 & 7.7 & 28.8 & 27.0 & 29.9 \\
\midrule
Llama3.2-Vision-11B & 0 & 2.8 & 0.0 & 0.0 & 32.1 & 29.1 & 29.7 & 17.7 & 17.1 & 27.1 & 90.6 & 89.3 & 85.6 & 31.1 & 33.4 & 18.1 & 12.7 & 11.5 & 9.3 & 37.5 & 44.6 & 45.5 & 8.4 & 8.1 & 22.5 & 20.6 & 0.0 & 0.0 \\
\midrule
PaliGemma-3B-448 & 0 & 1.4 & 1.0 & 0.7 & 27.6 & 1.2 & 2.3 & 16.5 & 8.1 & 13.6 & 84.3 & 0.7 & 1.6 & 27.2 & 0.0 & 0.0 & 11.6 & 0.0 & 0.0 & 32.5 & 0.0 & 0.0 & 10.4 & 0.0 & 0.0 & 13.4 & 0.0 & 0.0 \\
\midrule
\multirow{3}{*}{Phi3-4B} & 0 & 7.0 & 4.5 & 6.4 & 32.1 & 32.8 & 32.8 & 2.1 & 2.1 & 1.9 & 91.1 & 87.5 & 88.2 & 25.2 & 29.3 & 26.3 & 13.5 & 11.3 & 14.3 & 40.2 & 42.1 & 41.1 & 7.5 & 7.8 & 7.4 & 45.2 & 43.9 & 49.6 \\
 & 3 & 0.0 & 1.7 & 3.6 & 34.1 & 32.0 & 32.1 & 15.5 & 14.9 & 12.0 & 89.6 & 88.7 & 88.1 & 30.6 & 29.2 & 23.5 & 9.4 & 10.8 & 11.1 & 40.3 & 38.9 & 39.8 & 7.0 & 7.3 & 8.3 & 46.5 & 34.8 & 42.2 \\
 & 5 & 0.0 & 1.2 & 1.3 & 31.1 & 32.1 & 32.2 & 16.6 & 14.3 & 12.7 & 78.7 & 88.9 & 89.1 & 28.9 & 31.2 & 28.7 & 8.8 & 10.8 & 7.1 & 38.3 & 32.1 & 40.7 & 6.6 & 7.8 & 7.7 & 41.3 & 39.1 & 39.8 \\
 \midrule
\multirow{3}{*}{Phi3.5-Vision-3.5B} & 0 & 12.6 & 2.3 & 7.3 & 23.2 & 27.5 & 27.5 & 1.1 & 22.1 & 12.7 & 91.1 & 86.2 & 88.6 & 29.9 & 22.7 & 22.6 & 4.8 & 11.8 & 9.8 & 9.4 & 37.2 & 39.1 & 1.4 & 7.9 & 7.2 & 24.4 & 4.4 & 27.2 \\
 & 3 & 0.1 & 3.4 & 9.2 & 23.3 & 28.8 & 28.8 & 16.0 & 15.6 & 13.5 & 58.8 & 89.6 & 90.4 & 26.5 & 24.7 & 25.5 & 9.8 & 9.7 & 11.5 & 31.9 & 38.9 & 39.2 & 6.9 & 7.2 & 7.4 & 12.9 & 15.8 & 28.7 \\
 & 5 & 0.5 & 0.4 & 10.3 & 25.2 & 30.8 & 30.8 & 15.2 & 15.6 & 8.7 & 52.5 & 90.2 & 88.5 & 28.5 & 31.5 & 21.2 & 8.9 & 8.8 & 8.3 & 34.1 & 41.1 & 40.9 & 7.3 & 7.8 & 7.0 & 29.6 & 21.3 & 40.5 \\
 \midrule
\multirow{3}{*}{LLava 1.6-7B}& 0 & 11.1 & 20.4 & 22.8 & 24.6 & 21.4 & 20.8 & 13.4 & 3.3 & 1.1 & 91.1 & 72.4 & 71.9 & 19.3 & 23.4 & 22.8 & 10.9 & 8.5 & 10.7 & 15.8 & 28.6 & 22.9 & 8.9 & 4.5 & 3.6 & 12.6 & 9.1 & 12.4 \\
 & 3 & 0.0 & 0.1 & 0.2 & 7.1 & 15.2 & 17.8 & 3.6 & 1.1 & 5.2 & 10.4 & 15.2 & 29.3 & 12.2 & 21.5 & 20.8 & 4.3 & 10.3 & 9.1 & 9.5 & 30.7 & 32.7 & 5.4 & 8.4 & 5.5 & 5.4 & 19.4 & 21.9 \\
 & 5 & 0.6 & 0.0 & 0.0 & 11.4 & 10.9 & 9.7 & 0.0 & 0.0 & 0.0 & 24.1 & 4.3 & 0.7 & 21.5 & 22.3 & 20.1 & 5.5 & 7.1 & 7.2 & 17.4 & 30.2 & 27.9 & 5.6 & 7.7 & 5.9 & 5.6 & 6.5 & 5.8 \\
 \midrule
\multirow{3}{*}{Idefics2-8B}& 0 & 37.1 & 5.5 & 4.2 & 19.5 & 29.6 & 33.8 & 7.6 & 15.6 & 11.9 & 91.9 & 72.5 & 85.3 & 19.6 & 30.0 & 32.8 & 0.4 & 11.6 & 16.0 & 9.6 & 46.2 & 44.7 & 5.4 & 7.5 & 7.5 & 4.3 & 23.5 & 38.3 \\
 & 3 & 8.4 & 24.3 & 8.7 & 21.1 & 28.4 & 31.1 & 13.0 & 8.3 & 11.5 & 62.3 & 88.7 & 84.5 & 17.1 & 11.4 & 21.7 & 13.5 & 12.2 & 10.3 & 25.0 & 40.4 & 40.8 & 5.8 & 7.2 & 8.2 & 11.3 & 30.6 & 40.4 \\
 & 5 & 16.1 & 24.2 & 10.5 & 34.7 & 28.3 & 30.9 & 22.5 & 2.3 & 2.1 & 88.0 & 90.5 & 88.4 & 30.0 & 13.6 & 23.7 & 11.8 & 10.0 & 9.9 & 39.2 & 38.1 & 43.0 & 8.6 & 6.9 & 8.6 & 42.9 & 36.6 & 40.8 \\
 \midrule
\multirow{3}{*}{Idefics3-8B} & 0 & 16.1 & 20.7 & 17.1 & 24.3 & 24.3 & 22.1 & 0.0 & 5.0 & 2.3 & 91.5 & 90.7 & 91.6 & 38.5 & 35.0 & 9.3 & 11.0 & 11.1 & 4.5 & 5.8 & 6.0 & 32.9 & 6.2 & 5.0 & 9.1 & 17.2 & 18.0 & 5.0 \\
 & 3 & 8.7 & 10.1 & 6.2 & 26.9 & 26.9 & 21.9 & 8.1 & 7.5 & 1.1 & 84.0 & 86.4 & 90.6 & 22.2 & 23.0 & 5.8 & 13.1 & 12.0 & 11.9 & 32.2 & 31.0 & 38.9 & 7.0 & 6.5 & 4.5 & 21.8 & 22.0 & 0.6 \\
 & 5 & 4.4 & 9.0 & 5.4 & 22.3 & 26.9 & 20.9 & 5.5 & 8.5 & 0.0 & 65.1 & 88.3 & 90.7 & 15.1 & 17.5 & 3.5 & 15.1 & 14.8 & 6.4 & 27.6 & 28.0 & 39.8 & 5.4 & 8.7 & 5.6 & 22.7 & 22.5 & 0.0 \\
 \midrule
\multirow{3}{*}{VILA-1.5-8B} & 0 & 3.8 & 0.0 & 0.0 & 23.5 & 13.0 & 15.8 & 0.0 & 6.2 & 0.0 & 85.2 & 19.2 & 27.1 & 31.8 & 21.1 & 27.3 & 1.6 & 3.1 & 8.1 & 35.4 & 34.8 & 36.6 & 8.8 & 4.9 & 9.1 & 1.8 & 2.1 & 2.7 \\
 & 3 & 1.2 & 0.8 & 0.0 & 25.1 & 28.8 & 28.8 & 16.6 & 11.6 & 6.0 & 68.3 & 72.4 & 79.5 & 22.1 & 31.0 & 28.3 & 9.7 & 10.7 & 9.1 & 24.9 & 35.5 & 36.5 & 8.9 & 7.2 & 7.2 & 25.5 & 22.3 & 36.8 \\
 & 5 & 0.4 & 5.0 & 6.0 & 23.2 & 30.8 & 30.8 & 18.2 & 19.0 & 18.0 & 59.5 & 74.6 & 76.4 & 24.7 & 35.0 & 32.0 & 11.6 & 14.1 & 12.0 & 28.6 & 40.0 & 38.0 & 8.3 & 7.0 & 8.0 & 11.8 & 25.0 & 25.0 \\
 \midrule
\multirow{3}{*}{Qwen2-VL-2B} & 0 & 34.1 & 4.6 & 5.0 & 19.2 & 22.1 & 20.9 & 25.7 & 19.0 & 17.9 & 90.2 & 90.8 & 91.2 & 18.2 & 8.3 & 3.5 & 0.0 & 0.0 & 0.0 & 0.0 & 26.0 & 31.0 & 0.0 & 8.3 & 0.0 & 0.3 & 2.1 & 2.4 \\
 & 3 & 4.8 & 18.9 & 3.5 & 25.2 & 21.4 & 20.2 & 7.7 & 17.5 & 15.0 & 87.2 & 90.3 & 90.5 & 27.9 & 2.9 & 2.4 & 0.0 & 0.0 & 0.0 & 38.8 & 34.5 & 33.7 & 5.9 & 3.4 & 0.0 & 8.5 & 0.9 & 0.0 \\
 & 5 & 2.7 & 26.3 & 25.9 & 25.3 & 21.7 & 20.9 & 15.8 & 19.0 & 18.7 & 90.3 & 90.5 & 90.3 & 28.1 & 11.8 & 6.8 & 0.0 & 0.0 & 0.0 & 34.4 & 27.8 & 24.6 & 3.0 & 2.2 & 0.0 & 5.4 & 0.3 & 0.0 \\
 \midrule
\multirow{3}{*}{Qwen2-VL-7B} & 0 & 9.1 & 10.2 & 7.0 & 32.5 & 32.5 & 35.2 & 31.0 & 30.1 & 17.5 & 92.1 & 92.0 & 91.5 & 32.3 & 33.5 & 34.5 & 2.4 & 2.7 & 3.8 & 32.1 & 36.4 & 41.9 & 7.5 & 7.9 & 10.5 & 32.3 & 33.2 & 46.7 \\
 & 3 & 2.8 & 4.0 & 2.1 & 35.6 & 36.0 & 34.1 & 22.4 & 25.3 & 14.7 & 90.4 & 92.5 & 91.1 & 33.1 & 34.5 & 30.4 & 14.7 & 15.0 & 10.7 & 42.8 & 41.0 & 41.3 & 8.4 & 11.2 & 9.0 & 37.8 & 38.6 & 41.6 \\
 & 5 & 2.0 & 2.1 & 3.2 & 37.2 & 37.2 & 29.3 & 24.6 & 22.0 & 10.0 & 91.2 & 91.5 & 91.1 & 32.3 & 32.0 & 31.6 & 13.8 & 11.2 & 4.9 & 32.3 & 43.0 & 40.9 & 8.3 & 9.5 & 9.7 & 47.8 & 43.5 & 16.8 \\
 \midrule
\multirow{3}{*}{Qwen2-VL-72B} & 0 & 14.3 & 16.7 & 14.3 & 41.7 & 44.6 & 41.7 & 23.7 & 30.0 & 23.7 & 93.7 & \cellcolor[HTML]{F4CCCC}94.8 & \cellcolor[HTML]{F4CCCC}93.7 & \cellcolor[HTML]{F4CCCC}39.0 & \cellcolor[HTML]{F4CCCC}42.3 & \cellcolor[HTML]{EA9999}39.0 & 12.8 & \cellcolor[HTML]{E06666}19.8 & 12.8 & 47.2 & 51.0 & 47.2 & 13.4 & 13.2 & 13.4 & \cellcolor[HTML]{E06666}61.9 & \cellcolor[HTML]{EA9999}61.0 & \cellcolor[HTML]{EA9999}61.9 \\
 & 3 & 28.2 & 34.2 & 28.2 & \cellcolor[HTML]{F4CCCC}43.9 & 43.6 & \cellcolor[HTML]{EA9999}43.2 & 24.8 & 28.3 & 24.8 & 93.1 & 94.1 & 93.1 & 38.0 & 39.4 & \cellcolor[HTML]{F4CCCC}37.9 & \cellcolor[HTML]{F4CCCC}18.9 & 16.0 & 18.9 & 48.1 & \cellcolor[HTML]{F4CCCC}53.1 & \cellcolor[HTML]{F4CCCC}48.1 & 23.1 & 17.6 & 23.1 & \cellcolor[HTML]{EA9999}56.7 & 57.1 & \cellcolor[HTML]{F4CCCC}56.7 \\
 & 5 & 39.5 & 31.0 & 27.0 & \cellcolor[HTML]{F4CCCC}43.9 & \cellcolor[HTML]{F4CCCC}44.9 & 42.3 & 27.0 & 29.7 & 21.6 & 93.7 & 94.7 & 93.1 & \cellcolor[HTML]{E06666}41.9 & \cellcolor[HTML]{EA9999}43.9 & 35.8 & 15.5 & 13.1 & 19.8 & \cellcolor[HTML]{E06666}58.2 & \cellcolor[HTML]{EA9999}54.2 & \cellcolor[HTML]{EA9999}49.3 & 20.2 & 20.0 & 21.2 & 50.8 & 58.8 & 55.4 \\
 \midrule
\multirow{3}{*}{InternVL-4B} & 0 & 14.9 & 4.6 & 4.5 & 26.6 & 29.8 & 30.7 & 0.0 & 10.5 & 15.4 & 91.4 & 90.3 & 91.4 & 14.3 & 25.3 & 22.4 & 6.3 & 11.7 & 9.3 & 41.8 & 41.0 & 41.0 & 8.0 & 10.7 & 12.2 & 24.6 & 19.4 & 23.4 \\
 & 3 & 4.1 & 2.2 & 2.3 & 27.7 & 27.4 & 29.5 & 16.3 & 15.8 & 16.3 & 78.0 & 85.2 & 89.3 & 25.7 & 26.5 & 25.0 & 8.8 & 8.8 & 10.0 & 36.7 & 33.9 & 36.1 & 2.6 & 6.5 & 7.6 & 26.0 & 14.9 & 22.0 \\
 & 5 & 3.2 & 1.6 & 2.4 & 33.4 & 28.1 & 30.4 & 16.9 & 15.4 & 17.5 & 90.1 & 87.2 & 90.4 & 26.8 & 27.6 & 27.9 & 10.0 & 7.4 & 7.8 & 40.1 & 37.9 & 39.7 & 9.3 & 8.0 & 9.2 & 40.9 & 13.1 & 20.5 \\
 \midrule
\multirow{3}{*}{InternVL-8B} & 0 & 7.4 & 32.8 & \cellcolor[HTML]{F4CCCC}37.4 & 20.0 & 23.0 & 24.8 & 1.2 & 6.7 & 2.2 & 92.3 & 90.2 & 91.3 & 3.6 & 12.4 & 18.2 & 12.4 & 6.8 & 7.2 & 8.7 & 18.0 & 22.0 & 16.2 & 11.4 & 7.2 & 5.5 & 15.8 & 25.6 \\
 & 3 & 9.7 & 23.8 & 5.8 & 30.5 & 24.2 & 31.7 & 14.5 & 11.9 & 13.9 & 80.5 & 89.0 & 90.9 & 27.6 & 9.1 & 25.1 & 9.9 & 13.3 & 10.4 & 33.8 & 16.2 & 35.4 & 7.2 & 0.0 & 5.2 & 39.8 & 30.0 & 40.9 \\
 & 5 & 7.7 & 23.0 & 6.7 & 27.8 & 25.0 & 31.4 & 15.8 & 6.4 & 11.6 & 79.6 & 90.7 & 91.1 & 26.4 & 11.6 & 27.8 & 10.8 & 6.8 & 7.0 & 28.5 & 22.7 & 37.8 & 7.7 & 2.2 & 4.1 & 25.8 & 34.6 & 40.5 \\
 \midrule
\multirow{3}{*}{GPT-4o} & 0 & \cellcolor[HTML]{E06666}45.2 & \cellcolor[HTML]{E06666}46.5 & \cellcolor[HTML]{E06666}42.9 & \cellcolor[HTML]{EA9999}47.6 & \cellcolor[HTML]{EA9999}47.3 & \cellcolor[HTML]{F4CCCC}42.6 & \cellcolor[HTML]{E06666}51.7 & \cellcolor[HTML]{E06666}52.8 & \cellcolor[HTML]{EA9999}48.7 & \cellcolor[HTML]{E06666}95.5 & \cellcolor[HTML]{EA9999}95.7 & \cellcolor[HTML]{EA9999}94.6 & 32.9 & 28.0 & 14.1 & \cellcolor[HTML]{E06666}30.2 & \cellcolor[HTML]{EA9999}19.3 & \cellcolor[HTML]{E06666}21.9 & \cellcolor[HTML]{EA9999}52.4 & 49.9 & 42.3 & \cellcolor[HTML]{E06666}35.6 & \cellcolor[HTML]{E06666}40.3 & \cellcolor[HTML]{EA9999}34.5 & 34.8 & 45.2 & 42.2 \\
 & 3 & \cellcolor[HTML]{F4CCCC}42.8 & 39.8 & 30.2 & 38.9 & 37.5 & 35.7 & \cellcolor[HTML]{EA9999}49.8 & 33.7 & 32.9 & 93.8 & 92.9 & 87.0 & 21.9 & 21.7 & 15.6 & 10.8 & 3.5 & 11.6 & 46.2 & 44.4 & 41.3 & 27.9 & 30.2 & 20.8 & 28.7 & 42.3 & 41.1 \\
 & 5 & \cellcolor[HTML]{EA9999}43.4 & \cellcolor[HTML]{F4CCCC}42.3 & 30.7 & 41.9 & 39.8 & 38.4 & \cellcolor[HTML]{F4CCCC}46.8 & \cellcolor[HTML]{F4CCCC}42.6 & \cellcolor[HTML]{F4CCCC}40.3 & \cellcolor[HTML]{F4CCCC}94.2 & 94.2 & 87.4 & 28.9 & 19.2 & 14.9 & 10.7 & 9.5 & \cellcolor[HTML]{F4CCCC}20.3 & 47.6 & 44.9 & 40.6 & \cellcolor[HTML]{EA9999}29.6 & \cellcolor[HTML]{F4CCCC}31.2 & \cellcolor[HTML]{F4CCCC}26.1 & 35.2 & 37.2 & 39.1 \\
   \bottomrule
\end{tabular}%
\vspace{-0.2cm}
}

\caption{\textbf{Performance comparison between different LVLM on the natural image split of SalBench.} The comparison is given in terms of overall matching accuracy (columns 3 to 5) followed by F1 scores for overall (columns 6 to 8) and individual attributes (columns 9 to 29). The evaluation is also reported for both zero-shot and few-shot settings. Note: \textit{D}: Detection task; \textit{R}: Referring task; \textit{VR}: Visual Referring. The first highest, second highest, and third higest scores among models are highlighted in \colorbox[HTML]{E06666}{\phantom{xx}}, \colorbox[HTML]{EA9999}{\phantom{xx}}, and \colorbox[HTML]{F4CCCC}{\phantom{xx}}, respectively. See Section~\ref{sec:add_quant} for additional details.\vspace{-0.2cm}}
\label{tab:overall_o3_performance}
\end{table*}


\begin{table*}
\centering
\resizebox{0.8\textwidth}{!}{%
\begin{tabular}{|l|c|ccc|ccc|ccc|ccc|ccc|}
\toprule
\multirow{3}{*}{Model} & \multirow{3}{*}{Shot} & \multicolumn{3}{c|}{\multirow{2}{*}{\begin{tabular}[c]{@{}c@{}}Overall\\ Matching\end{tabular}}} & \multicolumn{12}{c|}{F1 Score} \\
\cmidrule{6-17}
 &  & \multicolumn{3}{c|}{} & \multicolumn{3}{c|}{Overall} & \multicolumn{3}{c|}{Orientation} & \multicolumn{3}{c|}{Color} & \multicolumn{3}{c|}{Size} \\
 \cmidrule{3-17}
 &  &\textit{ D} & \textit{R} & \textit{VR} & \textit{D} & \textit{R} & \textit{VR} & \textit{D} & \textit{R} & \textit{VR} & \textit{D} & \textit{R} & \textit{VR} & \textit{D} & \textit{R} & \textit{VR} \\
  \midrule
Claude-sonet & 0 & 86.4 & 89.0 & \cellcolor[HTML]{EA9999}87.8 & 86.7 & 90.3 & \cellcolor[HTML]{EA9999}87.7 & 83.4 & 87.6 & \cellcolor[HTML]{E06666}85.3 & 94.6 & 95.4 & 95.5 & 82.0 & 87.9 & \cellcolor[HTML]{EA9999}82.2 \\

  \midrule
NVLM-D-72B & 0 & 83.4 & 57.9 & 59.8 & 83.2 & 73.7 & 51.7 & 77.4 & 75.1 & 61.8 & \cellcolor[HTML]{E06666}98.0 & 80.2 & 80.4 & 74.1 & 65.7 & 12.7 \\

  \midrule
Molmo-7B & 0 & 71.3 & 45.4 & 30.1 & 67.2 & 38.0 & 28.4 & 40.8 & 62.3 & 34.5 & 95.3 & 23.3 & 15.7 & 69.3 & 28.5 & 22.3 \\

  \midrule
Molmo-72B & 0 & 84.1 & 67.0 & 75.5 & 83.4 & 65.6 & 73.6 & 80.7 & 73.4 & 77.5 & 96.5 & 69.4 & 84.5 & 72.9 & 54.0 & 58.5 \\

  \midrule
LLama3.2-Vision-11B & 0 & 51.4 & 17.6 & 55.5 & 48.7 & 52.4 & 52.4 & 52.6 & 57.9 & 59.7 & 62.7 & 58.6 & 69.7 & 30.9 & 40.7 & 27.8 \\

  \midrule
PaliGemma-3B-448 & 0 & 39.7 & 7.1 & 2.4 & 41.4 & 9.5 & 4.8 & 0.9 & 4.9 & 0.0 & 67.0 & 21.5 & 2.8 & 55.1 & 2.0 & 11.7 \\

 \midrule
\multirow{3}{*}{Phi3-4B}  & 0 & 51.3 & 59.0 & 52.1 & 41.2 & 55.3 & 47.2 & 12.4 & 66.3 & 45.9 & 45.3 & 50.5 & 62.8 & 65.9 & 49.1 & 32.9 \\
 & 3 & 43.4 & 39.0 & 47.1 & 33.5 & 27.1 & 38.6 & 24.0 & 17.3 & 5.8 & 26.5 & 54.9 & 55.0 & 50.0 & 9.1 & 55.0 \\
 & 5 & 34.2 & 35.1 & 50.8 & 17.0 & 18.9 & 46.7 & 0.0 & 4.7 & 34.5 & 51.0 & 51.6 & 66.6 & 0.0 & 0.4 & 39.1 \\

  \midrule
\multirow{3}{*}{Phi3.5-Vísion-3.5B}  & 0 & 44.0 & 59.9 & 64.9 & 35.0 & 53.7 & 63.6 & 2.1 & 53.7 & 53.7 & 49.2 & 50.9 & 71.3 & 53.7 & 56.6 & 65.9 \\
 & 3 & 26.7 & 49.8 & 34.7 & 19.5 & 41.0 & 20.8 & 0.0 & 0.5 & 3.0 & 18.2 & 66.7 & 9.9 & 40.3 & 55.8 & 49.5 \\
& 5 & 35.2 & 24.1 & 33.8 & 29.3 & 11.1 & 19.0 & 1.5 & 0.2 & 0.0 & 38.9 & 26.0 & 7.6 & 47.5 & 7.1 & 49.4 \\

  \midrule
\multirow{3}{*}{LLava 1.6-7B} & 0 & 31.2 & 18.2 & 17.7 & 16.3 & 10.1 & 16.6 & 0.0 & 0.0 & 0.0 & 0.1 & 12.3 & 49.9 & 48.9 & 18.1 & 0.0 \\
 & 3 & 32.4 & 17.7 & 34.2 & 16.4 & 8.8 & 17.0 & 0.0 & 1.4 & 0.0 & 0.0 & 10.1 & 50.9 & 49.0 & 15.1 & 0.0 \\
 & 5 & 32.4 & 19.9 & 34.2 & 16.4 & 9.1 & 17.0 & 0.0 & 0.2 & 0.0 & 0.0 & 18.1 & 50.9 & 49.0 & 9.1 & 0.0 \\

  \midrule
\multirow{3}{*}{Idefic2-8B}  & 0 & 64.5 & 45.2 & 56.0 & 64.3 & 36.6 & 49.5 & 62.9 & 51.1 & 63.8 & 78.1 & 9.7 & 64.1 & 51.9 & 49.2 & 20.5 \\
 & 3 & 66.9 & 42.6 & 48.7 & 66.3 & 34.2 & 39.5 & 66.6 & 9.7 & 66.3 & 79.4 & 39.8 & 9.5 & 53.0 & 53.1 & 9.7 \\
& 5 & 66.7 & 49.6 & 43.1 & 67.2 & 42.6 & 34.5 & 65.3 & 8.6 & 54.5 & 79.2 & 62.9 & 11.9 & 57.2 & 56.3 & 37.0 \\

  \midrule
\multirow{3}{*}{Idefic3-8B} & 0 & 40.2 & 58.3 & 35.5 & 28.4 & 52.8 & 19.2 & 24.1 & 54.9 & 2.3 & 54.3 & 51.0 & 49.7 & 6.9 & 52.5 & 5.5 \\
 & 3 & 50.9 & 35.9 & 50.7 & 40.3 & 20.7 & 40.6 & 0.5 & 0.5 & 3.4 & 62.9 & 52.6 & 63.6 & 57.6 & 8.9 & 54.8 \\
 & 5 & 36.3 & 34.5 & 62.9 & 21.4 & 18.1 & 58.3 & 0.0 & 0.2 & 64.3 & 51.8 & 51.3 & 85.7 & 12.3 & 2.7 & 25.0 \\

  \midrule
 \multirow{3}{*}{VILA-1.5-8B}& 0 & 34.2 & 30.4 & 47.5 & 40.0 & 15.8 & 17.0 & 17.6 & 0.0 & 0.5 & 56.3 & 28.8 & 50.5 & 46.1 & 18.7 & 0.0 \\
 & 3 & 34.2 & 36.9 & 34.2 & 17.0 & 28.8 & 17.0 & 0.0 & 0.0 & 0.5 & 51.0 & 47.6 & 50.5 & 0.0 & 38.5 & 0.0 \\
 & 5 & 34.2 & 39.5 & 34.2 & 17.0 & 30.8 & 17.0 & 0.0 & 0.0 & 0.5 & 51.0 & 51.3 & 50.5 & 0.0 & 41.3 & 0.0 \\

  \midrule
\multirow{3}{*}{Qwen2-VL-2B} & 0 & 30.3 & 34.5 & 34.5 & 26.3 & 20.6 & 20.2 & 14.5 & 5.0 & 10.7 & 5.9 & 7.0 & 1.6 & 58.3 & 49.8 & 49.6 \\
 & 3 & 35.7 & 35.3 & 32.4 & 23.3 & 21.8 & 16.3 & 0.0 & 0.0 & 0.0 & 17.5 & 15.2 & 0.0 & 53.8 & 50.1 & 49.0 \\
 & 5 & 35.3 & 32.6 & 33.1 & 23.8 & 16.5 & 17.7 & 0.0 & 0.0 & 4.1 & 15.2 & 0.7 & 0.0 & 54.6 & 49.0 & 49.3 \\

  \midrule
\multirow{3}{*}{Qwen2-VL-7B}  & 0 & 60.2 & 40.0 & 59.9 & 55.7 & 34.2 & 57.4 & 23.7 & 17.7 & 53.6 & 82.0 & 45.0 & 66.9 & 61.6 & 40.3 & 51.5 \\
 & 3 & 63.7 & 34.2 & 69.8 & 53.8 & 17.0 & 64.2 & 2.5 & 0.0 & 33.5 & 94.8 & 50.9 & 84.9 & 64.1 & 0.0 & 74.0 \\
& 5 & 64.5 & 34.2 & 73.4 & 54.9 & 17.7 & 72.0 & 4.5 & 0.0 & 56.3 & 95.6 & 50.9 & 84.1 & 64.6 & 2.0 & 75.5 \\

  \midrule
\multirow{3}{*}{Qwen2-VL-72B} & 0 & 89.1 & \cellcolor[HTML]{E06666}93.6 & 76.0 & 88.8 & \cellcolor[HTML]{E06666}93.6 & 74.7 & 85.2 & \cellcolor[HTML]{E06666}91.3 & 72.5 & 97.2 & \cellcolor[HTML]{E06666}98.3 & 86.0 & 83.9 & \cellcolor[HTML]{E06666}91.1 & 65.7 \\
 & 3 & \cellcolor[HTML]{E06666}89.3 & \cellcolor[HTML]{EA9999}93.1 & 86.1 & \cellcolor[HTML]{EA9999}89.3 & \cellcolor[HTML]{EA9999}93.1 & 85.9 & \cellcolor[HTML]{EA9999}86.7 & \cellcolor[HTML]{EA9999}90.4 & 82.9 & 95.8 & \cellcolor[HTML]{EA9999}97.9 & \cellcolor[HTML]{EA9999}96.2 & \cellcolor[HTML]{E06666}85.5 & \cellcolor[HTML]{E06666}91.1 & 78.8 \\
 & 5 & \cellcolor[HTML]{EA9999}89.2 & \cellcolor[HTML]{F4CCCC}92.7 & \cellcolor[HTML]{E06666}88.0 & \cellcolor[HTML]{E06666}89.9 & \cellcolor[HTML]{F4CCCC}92.6 & \cellcolor[HTML]{E06666}87.9 & \cellcolor[HTML]{E06666}88.3 & \cellcolor[HTML]{F4CCCC}90.0 & \cellcolor[HTML]{EA9999}84.8 & 96.1 & 97.4 & \cellcolor[HTML]{E06666}96.5 & \cellcolor[HTML]{EA9999}85.4 & \cellcolor[HTML]{F4CCCC}90.5 & \cellcolor[HTML]{E06666}82.3 \\

  \midrule
\multirow{3}{*}{InternVL-4B}
 & 0 & 47.2 & 69.5 & 58.9 & 41.5 & 63.4 & 52.2 & 25.4 & 31.2 & 67.2 & 64.5 & 88.2 & 67.1 & 34.7 & 70.6 & 22.4 \\
 & 3 & 34.2 & 37.3 & 49.9 & 17.0 & 25.3 & 41.7 & 0.0 & 0.0 & 2.3 & 50.9 & 24.9 & 66.5 & 0.0 & 50.9 & 56.5 \\
 & 5 & 34.2 & 48.0 & 58.1 & 17.0 & 39.1 & 52.5 & 0.0 & 0.0 & 61.7 & 50.9 & 61.4 & 76.5 & 0.0 & 55.9 & 19.5 \\

  \midrule
\multirow{3}{*}{InternVL-8B} & 0 & 65.6 & 74.2 & 37.0 & 58.7 & 71.9 & 23.0 & 66.9 & 50.4 & 9.9 & 95.8 & 93.7 & 52.0 & 13.4 & 71.5 & 7.1 \\
 & 3 & 60.6 & 61.7 & 66.9 & 52.3 & 51.7 & 64.4 & 7.4 & 1.6 & 44.5 & 87.0 & 90.9 & 85.7 & 62.6 & 62.4 & 63.0 \\
& 5 & 51.0 & 62.5 & 61.6 & 43.9 & 53.7 & 50.5 & 15.6 & 8.6 & 66.5 & 60.4 & 89.2 & 83.6 & 55.6 & 63.3 & 1.4 \\

  \midrule
\multirow{3}{*}{GPT-4o}  & 0 & \cellcolor[HTML]{EA9999}89.2 & 88.7 & 74.7 & \cellcolor[HTML]{F4CCCC}89.2 & 88.4 & 73.5 & \cellcolor[HTML]{F4CCCC}86.3 & 85.2 & 73.9 & 97.2 & 96.7 & 94.6 & \cellcolor[HTML]{F4CCCC}84.0 & 83.5 & 52.0 \\
 & 3 & 87.7 & 88.0 & 86.3 & 88.4 & 87.7 & 86.7 & 85.8 & 84.7 & 82.8 & \cellcolor[HTML]{F4CCCC}97.3 & 95.6 & \cellcolor[HTML]{F4CCCC}95.8 & 82.8 & 82.7 & 81.4 \\
& 5 & 86.0 & 89.0 & \cellcolor[HTML]{F4CCCC}87.1 & 86.0 & 89.1 & \cellcolor[HTML]{F4CCCC}87.4 & 82.8 & 85.3 & \cellcolor[HTML]{F4CCCC}84.4 & \cellcolor[HTML]{EA9999}97.6 & \cellcolor[HTML]{EA9999}97.9 & 95.7 & 77.5 & 84.1 & \cellcolor[HTML]{F4CCCC}82.0 \\
\bottomrule
\end{tabular}%
}

\caption{\textbf{Performance comparison on the synthetic image split of SalBench.} Note: \textit{D}: Detection task; \textit{R}: Referring task; \textit{VR}: Viusal Referring. The first highest, second highest, and third higest scores among models are highligt in \colorbox[HTML]{E06666}{\phantom{xx}}, \colorbox[HTML]{EA9999}{\phantom{xx}}, and \colorbox[HTML]{F4CCCC}{\phantom{xx}}, respectively. See \reftab{tab:overall_o3_performance} caption and Section~\ref{sec:add_quant} for more details.}
\label{tab:overall_p3_performance}
\end{table*}

\begin{table*}
\centering
\resizebox{0.8\textwidth}{!}{%
\begin{tabular}{lccccccccccc}
\toprule
\cellcolor[HTML]{FFFFFF} &  &  & \multicolumn{3}{c}{Color} & \multicolumn{3}{c}{Orientation} & \multicolumn{3}{c}{Size} \\
\multirow{-2}{*}{Model} & \multirow{-2}{*}{Shots} & \multirow{-2}{*}{Difficulty} & D & R & VR & D & R & VR & D & R & VR \\
   \midrule
 & 0 & easy & \cellcolor[HTML]{93C47D}100.0 & 99.7 & 99.8 & 91.0 & 96.2 & \cellcolor[HTML]{93C47D}99.0 & 92.1 & 95.6 & 85.4 \\
 & 0 & medium & 97.3 & 94.6 & 97.3 & 92.4 & \cellcolor[HTML]{6D9EEB}96.9 & \cellcolor[HTML]{6D9EEB}98.3 & 63.3 & 71.4 & 73.3 \\
\multirow{-3}{*}{Claude} & 0 & hard & 54.5 & 43.8 & 47.1 & 81.3 & 86.5 & 93.8 & 50.0 & 56.0 & 36.0 \\
   \midrule
 & 0 & easy & 69.9 & 43.7 & 57.7 & 59.4 & 8.7 & 93.4 & 21.9 & \cellcolor[HTML]{CCCCCC}2.3 & 19.6 \\
 & 0 & medium & 75.2 & 45.0 & 59.7 & 68.8 & 8.0 & 93.4 & 21.0 & \cellcolor[HTML]{CCCCCC}0.0 & 16.2 \\
\multirow{-3}{*}{Llama3.2-Vision-11B} & 0 & hard & 66.1 & 28.9 & 40.5 & 59.4 & 9.0 & 93.1 & 14.0 & \cellcolor[HTML]{CCCCCC}0.0 & 7.3 \\
  \midrule
 & 0 & easy & 96.7 & 6.3 & 45.2 & 24.3 & 100.0 & 92.0 & 98.5 & 29.8 & 42.1 \\
 & 0 & medium & 96.0 & 5.4 & 42.3 & 27.1 & 99.7 & 91.3 & \cellcolor[HTML]{6D9EEB}100.0 & 19.0 & 39.5 \\
\multirow{-3}{*}{Molmo-72B} & 0 & hard & 66.9 & \cellcolor[HTML]{CCCCCC}1.7 & 19.8 & 16.3 & 99.7 & 84.4 & 99.3 & 19.3 & 30.0 \\
  \midrule
 & 0 & easy & \cellcolor[HTML]{93C47D}100.0 & 99.5 & 90.2 & 69.4 & 46.5 & 89.6 & 83.3 & 15.6 & 9.6 \\
 & 0 & medium & \cellcolor[HTML]{6D9EEB}100.0 & 98.0 & 79.9 & 79.9 & 55.6 & 87.8 & 61.0 & 15.7 & \cellcolor[HTML]{CCCCCC}1.9 \\
\multirow{-3}{*}{NVLM} & 0 & hard & 81.8 & 74.4 & 43.0 & 94.1 & 81.9 & 88.2 & 44.7 & 8.7 & \cellcolor[HTML]{CCCCCC}4.7 \\
   \midrule
   
 & 0 & easy & 99.8 & 99.0 & 96.1 & \cellcolor[HTML]{93C47D}96.2 & 89.2 & 97.6 & 93.3 & 96.9 & 42.5 \\
 & 0 & medium & \cellcolor[HTML]{6D9EEB}100.0 & \cellcolor[HTML]{6D9EEB}100.0 & 93.3 & 96.9 & 88.5 & 94.8 & 59.0 & 59.0 & 39.0 \\
\multirow{-3}{*}{GPT-4o} & 0 & hard & 66.1 & 66.1 & 57.9 & \cellcolor[HTML]{EA9999}98.6 & 96.5 & \cellcolor[HTML]{EA9999}96.2 & 36.7 & 35.3 & 11.3 \\
  \midrule
  
 & 3 & easy & 99.2 & 99.8 & 98.4 & 95.1 & 89.9 & 89.9 & 91.0 & \cellcolor[HTML]{93C47D}98.8 & 91.5 \\
 & 3 & medium & 96.6 & \cellcolor[HTML]{6D9EEB}100.0 & 94.0 & 93.1 & 92.4 & 84.4 & 58.6 & 71.0 & 77.1 \\
\multirow{-3}{*}{GPT-4o} & 3 & hard & \cellcolor[HTML]{EA9999}74.4 & 73.6 & 57.0 & 94.4 & 92.0 & 80.6 & 35.3 & 43.3 & 56.7 \\

   \midrule
 & 5 & easy & 99.5 & 99.8 & 98.7 & 96.9 & 91.3 & 89.9 & 80.2 & 96.7 & 92.7 \\
 
 & 5 & medium & 98.7 & \cellcolor[HTML]{6D9EEB}100.0 & 96.0 & \cellcolor[HTML]{6D9EEB}96.5 & 93.4 & 87.5 & 52.9 & 71.0 & \cellcolor[HTML]{6D9EEB}84.8 \\

\multirow{-3}{*}{GPT-4o} & 5 & hard & 70.2 & 71.1 & 52.1 & 99.0 & 91.0 & 82.6 & 29.3 & 32.7 & 46.7 \\

   \midrule
 & 0 & easy & 97.6 & 99.7 & 99.8 & 91.3 & 41.3 & \cellcolor[HTML]{CCCCCC}0.0 & 11.3 & 88.5 & \cellcolor[HTML]{CCCCCC}0.0 \\
 & 0 & medium & 99.3 & 99.3 & 99.3 & 94.1 & 45.1 & \cellcolor[HTML]{CCCCCC}0.3 & \cellcolor[HTML]{CCCCCC}3.3 & 97.1 & \cellcolor[HTML]{CCCCCC}0.0 \\
\multirow{-3}{*}{InyternVL-2-4B} & 0 & hard & 67.8 & 61.2 & \cellcolor[HTML]{EA9999}96.7 & 95.1 & 20.1 & \cellcolor[HTML]{CCCCCC}0.0 & \cellcolor[HTML]{CCCCCC}0.0 & 99.3 & \cellcolor[HTML]{CCCCCC}0.0 \\

  \midrule
 & 3 & easy & \cellcolor[HTML]{CCCCCC}0.0 & 93.8 & 93.5 & \cellcolor[HTML]{CCCCCC}0.0 & \cellcolor[HTML]{CCCCCC}0.0 & 64.6 & \cellcolor[HTML]{CCCCCC}0.0 & 97.1 & 41.0 \\
 & 3 & medium & \cellcolor[HTML]{CCCCCC}0.0 & 91.3 & 92.6 & \cellcolor[HTML]{CCCCCC}0.0 & \cellcolor[HTML]{CCCCCC}2.1 & 66.7 & \cellcolor[HTML]{CCCCCC}0.0 & \cellcolor[HTML]{6D9EEB}95.7 & 52.9 \\
\multirow{-3}{*}{InyternVL-2-4B} & 3 & hard & \cellcolor[HTML]{CCCCCC}0.0 & 59.5 & 86.8 & \cellcolor[HTML]{CCCCCC}0.0 & \cellcolor[HTML]{CCCCCC}0.3 & 58.0 & \cellcolor[HTML]{CCCCCC}0.0 & 92.0 & 44.0 \\

  \midrule
 & 5 & easy & \cellcolor[HTML]{CCCCCC}0.0 & 92.2 & 73.5 & \cellcolor[HTML]{CCCCCC}0.0 & \cellcolor[HTML]{CCCCCC}2.8 & 100.0 & \cellcolor[HTML]{CCCCCC}0.0 & 95.6 & \cellcolor[HTML]{CCCCCC}0.0 \\
 & 5 & medium & \cellcolor[HTML]{CCCCCC}0.0 & 92.6 & 71.8 & \cellcolor[HTML]{CCCCCC}0.0 & 7.6 & 100.0 & \cellcolor[HTML]{CCCCCC}0.0 & 92.9 & \cellcolor[HTML]{CCCCCC}0.0 \\
\multirow{-3}{*}{InyternVL-2-4B} & 5 & hard & \cellcolor[HTML]{CCCCCC}0.0 & 63.6 & 60.3 & \cellcolor[HTML]{CCCCCC}0.0 & \cellcolor[HTML]{CCCCCC}3.1 & 100.0 & \cellcolor[HTML]{CCCCCC}0.0 & \cellcolor[HTML]{EA9999}96.0 & \cellcolor[HTML]{CCCCCC}0.0 \\

  \midrule
 & 0 & easy & \cellcolor[HTML]{93C47D}100.0 & \cellcolor[HTML]{93C47D}100.0 & 96.1 & 95.8 & \cellcolor[HTML]{93C47D}96.9 & 59.0 & 94.2 & \cellcolor[HTML]{93C47D}98.8 & 19.0 \\
 & 0 & medium & \cellcolor[HTML]{6D9EEB}100.0 & \cellcolor[HTML]{6D9EEB}100.0 & 83.9 & 95.8 & 96.2 & 53.5 & 50.5 & 74.3 & 6.7 \\
\multirow{-3}{*}{Qwen2-VL72B} & 0 & hard & 66.1 & 80.2 & 61.2 & \cellcolor[HTML]{EA9999}98.6 & \cellcolor[HTML]{EA9999}98.3 & 46.9 & 46.0 & 62.0 & \cellcolor[HTML]{CCCCCC}2.0 \\
  \midrule

 & 3 & easy & \cellcolor[HTML]{93C47D}100.0 & \cellcolor[HTML]{93C47D}100.0 & 97.9 & 86.1 & 92.4 & 70.8 & 98.8 & 98.5 & 41.7 \\
 & 3 & medium & \cellcolor[HTML]{6D9EEB}100.0 & \cellcolor[HTML]{6D9EEB}100.0 & 92.6 & 88.5 & 93.1 & 66.7 & 66.7 & 78.6 & 25.7 \\

 \multirow{-3}{*}{Qwen2-VL72B} & 3 & hard & 66.9 & 82.6 & 76.0 & 94.4 & 95.5 & 65.3 & 52.0 & 66.7 & 20.0 \\

    \midrule
 & 5 & easy & \cellcolor[HTML]{93C47D}100.0 & \cellcolor[HTML]{93C47D}100.0 & 96.7 & 92.7 & 94.8 & 81.9 & 97.5 & \cellcolor[HTML]{93C47D}98.8 & 40.8 \\
 
 & 5 & medium & \cellcolor[HTML]{6D9EEB}100.0 & \cellcolor[HTML]{6D9EEB}100.0 & 93.3 & 93.4 & 93.8 & 78.8 & 58.6 & 71.9 & 27.6 \\
\multirow{-3}{*}{Qwen2-VL72B} & 5 & hard & 62.0 & 81.0 & 70.2 & 95.1 & 93.4 & 74.3 & 46.0 & 68.0 & 26.0 \\

   \midrule
 & 0 & easy & 92.7 & \cellcolor[HTML]{93C47D}100.0 & \cellcolor[HTML]{93C47D}100.0 & \cellcolor[HTML]{CCCCCC}2.4 & 12.5 & 44.4 & 97.9 & 42.7 & 42.3 \\
 & 0 & medium & 85.9 & \cellcolor[HTML]{6D9EEB}100.0 & \cellcolor[HTML]{6D9EEB}100.0 & \cellcolor[HTML]{CCCCCC}3.1 & 16.7 & 45.5 & 80.5 & \cellcolor[HTML]{CCCCCC}3.8 & 19.5 \\
\multirow{-3}{*}{Qwen2-VL-7B} & 0 & hard & 59.5 & \cellcolor[HTML]{EA9999}100.0 & 97.5 & \cellcolor[HTML]{CCCCCC}2.8 & 11.5 & 34.0 & 94.7 & 26.0 & 45.3 \\

   \midrule
 & 3 & easy & \cellcolor[HTML]{CCCCCC}0.0 & \cellcolor[HTML]{93C47D}100.0 & 99.8 & \cellcolor[HTML]{CCCCCC}0.0 & \cellcolor[HTML]{CCCCCC}0.0 & 20.8 & \cellcolor[HTML]{93C47D}100.0 & \cellcolor[HTML]{CCCCCC}0.0 & \cellcolor[HTML]{93C47D}92.5 \\
 & 3 & medium & \cellcolor[HTML]{CCCCCC}0.0 & \cellcolor[HTML]{6D9EEB}100.0 & 99.3 & \cellcolor[HTML]{CCCCCC}0.0 & \cellcolor[HTML]{CCCCCC}0.0 & 24.7 & \cellcolor[HTML]{6D9EEB}100.0 & \cellcolor[HTML]{CCCCCC}0.0 & 87.1 \\
\multirow{-3}{*}{Qwen2-VL-7B} & 3 & hard & \cellcolor[HTML]{CCCCCC}0.0 & \cellcolor[HTML]{EA9999}100.0 & 89.3 & \cellcolor[HTML]{CCCCCC}0.0 & \cellcolor[HTML]{CCCCCC}0.0 & 18.1 & \cellcolor[HTML]{EA9999}100.0 & \cellcolor[HTML]{CCCCCC}0.0 & 84.0 \\

   \midrule
 & 5 & easy & \cellcolor[HTML]{CCCCCC}0.0 & \cellcolor[HTML]{93C47D}100.0 & \cellcolor[HTML]{93C47D}100.0 & \cellcolor[HTML]{CCCCCC}0.0 & \cellcolor[HTML]{CCCCCC}0.0 & 52.8 & \cellcolor[HTML]{93C47D}100.0 & \cellcolor[HTML]{CCCCCC}0.0 & 74.4 \\
 & 5 & medium & \cellcolor[HTML]{CCCCCC}0.0 & \cellcolor[HTML]{6D9EEB}100.0 & \cellcolor[HTML]{6D9EEB}100.0 & \cellcolor[HTML]{CCCCCC}0.0 & \cellcolor[HTML]{CCCCCC}0.0 & 51.7 & \cellcolor[HTML]{6D9EEB}100.0 & \cellcolor[HTML]{CCCCCC}0.0 & 71.0 \\
\multirow{-3}{*}{Qwen2-VL-7B} & 5 & hard & \cellcolor[HTML]{CCCCCC}0.0 & \cellcolor[HTML]{EA9999}100.0 & 95.9 & \cellcolor[HTML]{CCCCCC}0.0 & \cellcolor[HTML]{CCCCCC}0.0 & 42.0 & \cellcolor[HTML]{EA9999}100.0 & \cellcolor[HTML]{CCCCCC}0.7 & \cellcolor[HTML]{EA9999}61.3 \\
\bottomrule
\end{tabular}%
}

\caption{\textbf{Performance comparison between different LVLM across the categories and difficulty levels on the synthetic images split of SalBench.} The highest accuracy scores among models across the levels Easy, Medium, and Hard are highlighted in \colorbox[HTML]{52be80}{\phantom{xx}}, \colorbox[HTML]{6495ED}{\phantom{xx}}, and \colorbox[HTML]{f1948a}{\phantom{xx}}, respectively. The lowest values are highlighted in \colorbox[HTML]{ccd1d1}{\phantom{xx}}. See Section~\ref{sec:difficult_levels} for more details.}
\label{tab:o3_performance_level}
\end{table*}

\section{Additional Quantitative Results\label{sec:add_quant}}
This section aims to supplement the quantitative analysis presented in the main manuscript. \reftab{tab:overall_o3_performance} and \reftab{tab:overall_p3_performance}, extended from Table 1 in the mains text, present the comprehensive comparison on both Matching accuracy and F1 score, especially on all categories of synthetic and natural image splits, respectively.

\noindent\textbf{Matching Performance:} Across the  natural and synthetic images splits, the performance of the investigated models reveals notable disparities. On the natural images split (reported in \reftab{tab:overall_o3_performance}), the matching scores are predominantly low, with most models achieving less than 10\%. However, a few models, such as GPT-4o, Claude\cite{claude}, and the unexpectedly strong Idefics2-8B \cite{laurenccon2024matters}, stood out by attaining scores around 40\%. In stark contrast, comparison on the synthetic images split (presented in \reftab{tab:overall_p3_performance}) demonstrates significantly higher scores, with the Qwen2-VL-72B \cite{wang2024qwen2} model achieving a remarkable 93.6\% in the Referring task.
Furthermore, task-specific trends emerge. Compared to Detection and Visual Referring tasks, models consistently performed better on Referring tasks across both benchmarks. These results underscore the challenges posed by natural images and highlight the relative ease and clarity offered by the synthetic images split.

On the natural images split (\reftab{tab:overall_o3_performance}), matching scores decline across all models in the few-shot setting, with the notable exception of Qwen2-VL-72B , which demonstrates a significant improvement in the Detection task, rising from 14.3\% to 39\%. Conversely, on the synthetic split (\reftab{tab:overall_p3_performance}), the few-shot setting generally enhances matching scores. For instance, Idefics3-8B  shows remarkable progress in the Visual Referring task, increasing from 35.5\% to 62.9\%. These trends highlight the contrasting impact of few-shot settings between natural and synthetic splits, reflecting the relative complexity of the datasets and tasks.

\noindent\textbf{Color - Where Models Shine Brightest:} Among the evaluated visual attributes, \textit{color} stands out as a category, where LVLM demonstrate exceptional performance. On the synthetic split, with the exception of models like LLaVA and PaliGemma, most achieve higher F1 scores in color-based tasks compared to \textit{orientation} and \textit{size}. Notably, GPT-4o  and Qwen2-VL-72B  deliver near-perfect results in this category. Similarly, even on the more complex natural images split, models maintain relatively strong performance in identifying color-based distinctions. For instance, Claude surpasses its performance on other visual attributes such as \textit{size} and \textit{orientation}. This consistent strength across splits highlights the potential of current LVLMs to excel in color-related tasks.

\noindent\textbf{Focus - Challenges in Natural Image Benchmarks:} Identifying the difference between an object and others in a natural image falls under the \textit{focus} attribute. This task is particularly challenging not only for models but also for humans viewing such images with their own eyes. For this attribute, the target object will be either sharply in focus compared to blurred distractors or intentionally out of focus while the surrounding objects remain clear \reffig{fig:o3-example-1} and \reffig{fig:o3-example-2}. Even top-performing models, such as ChatGPT-4o , which achieves an F1 score of only 30\% on the Detection task, encounter significant challenges with this attribute, performing noticeably worse compared to attributes like \textit{color}. The difficulty stems from the intricate nature of natural images, where depth of field, motion blur, and bokeh effects create complex visual environments. These photographic elements require models to interpret fine-grained variations in sharpness and visual clarity, skills they are not explicitly trained to handle.

\noindent\textbf{Open-source \emph{vs.} Close-Source:} From \reftab{tab:overall_o3_performance}, we observe that the two closed-source  models (GPT-4o and Claude) achieve the highest matching accuracy and F1 scores, in both overall and across individual categories. For instance, GPT-4o attains an impressive matching score of 45.2\% on the natural images split, outperforming open-source counterparts like Qwen2-VL-72B, which achieves only 14.3\%. Similarly, Claude excels with an overall F1 score of 40.6\%, significantly ahead of open-source models such as Molmo-72B, which achieves just 19.2\%.

However, on the synthetic split, the dominance shifts, with the open-source model Qwen2-VL-72B leading the pack. This model achieves a remarkable matching score of 93.6\% on the Referring task, far surpassing even the best closed-source models. The high performance of Qwen2-VL-72B underscores the effectiveness of open-source models when working with structured and controlled datasets.
This contrast highlights a divide in capabilities: closed-source models excel in real-world image complexity, while open-source models demonstrate strength in synthetic benchmarks, where tasks often involve clear and controlled visual distinctions.

\subsection{Performance Across Difficulty Levels\label{sec:difficult_levels}}
From \reftab{tab:o3_performance_level}, it is clear that model performance on the synthetic split declines as task difficulty increases. For easy examples, most models perform exceptionally well, particularly in the \textit{color} category, where models like Claude, NVLM, and Qwen’s variants achieve perfect accuracy (100\%) across all tasks (Detection, Referring, and Visual Referring). However, as the difficulty level increases to medium, accuracy declines noticeably, even for top-performing models. For instance, GPT-4o’s accuracy on \textit{size} decreases by nearly 30\% in both Detection and Referring tasks and by 10\% in Visual Referring.
The hard category poses the greatest challenge, with even leading models scoring below 70\% on average in most attributes. However, accuracy for \textit{orientation} remains relatively stable compared to other categories. For example, Claude demonstrates a smaller decline, maintaining approximately 81\% accuracy in the Visual Referring task despite the increased difficulty. This stability suggests that orientation-based tasks, which rely on spatial differences, may be less sensitive to added complexity than attributes like \textit{size}, where subtle distinctions become increasingly difficult to detect.

\noindent\textbf{Attribute-Specific Trends:} Models perform exceptionally well in color-based tasks, with accuracy often exceeding 90\% in all difficulty levels. In contrast, \textit{size} and \textit{orientation} tasks are more challenging, especially in the hard category, where performance frequently falls below 50\%. These results highlight the varying difficulty of visual attributes, suggesting that models are better optimized for color-based distinctions while struggling with more nuanced spatial relationships.

\noindent\textbf{Models with Limited Category Coverage:}
A notable observation from \reftab{tab:o3_performance_level} is that some models exhibit selective performance, responding effectively to only two categories while failing entirely in others. For instance, Qwen2-VL-7B achieves perfect accuracy (100\%) in the \textit{color} and \textit{size} categories but fails to provide meaningful predictions for tasks in the \textit{orientation} category, with accuracy consistently at 0\%.

\section{Additional Qualitative Comparisons\label{sec:add_qual}}
\begin{figure}[!t]
    \centering
    \includegraphics[width=1\columnwidth]{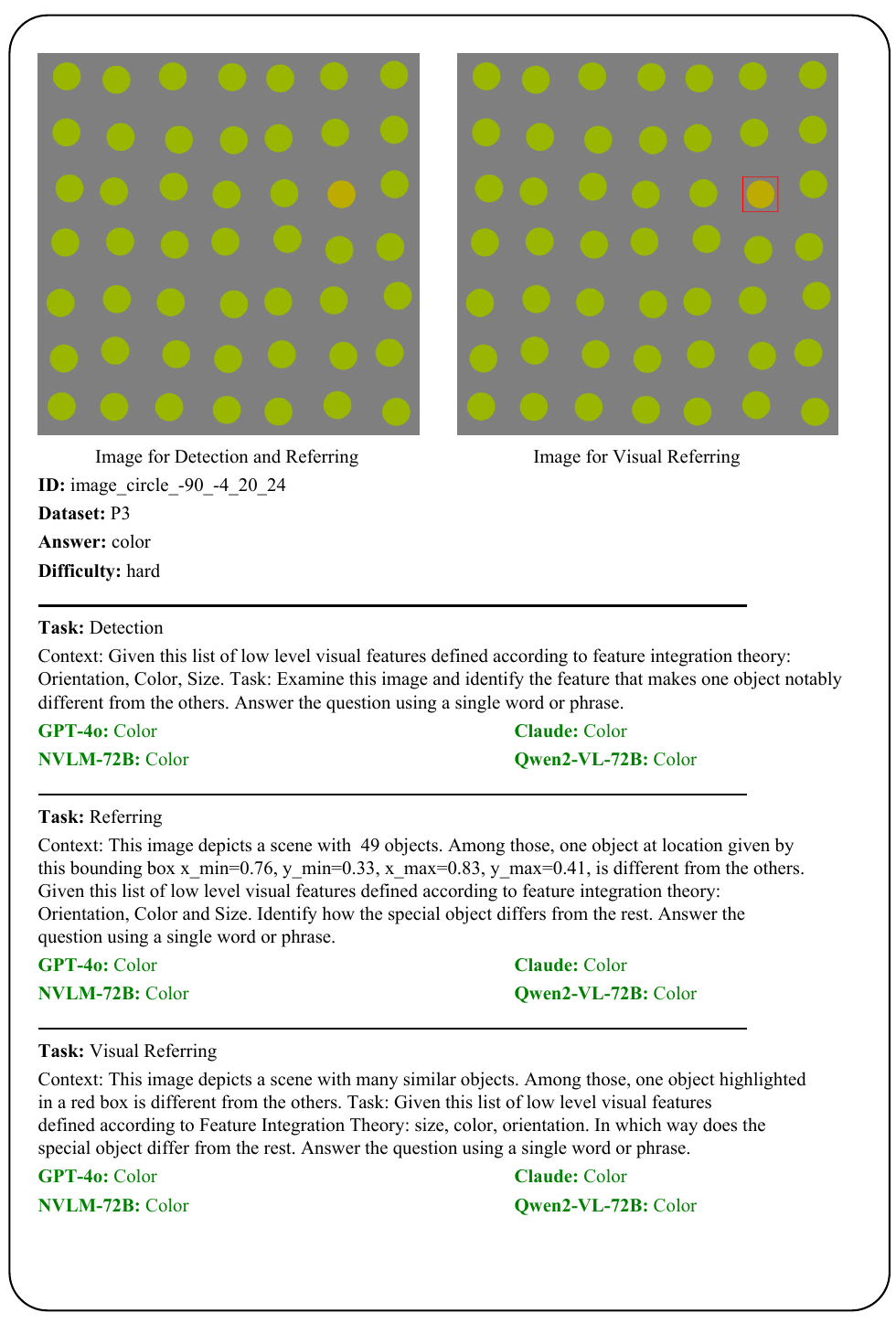}
    \caption{A hard example for the \textit{color} attribute in synthetic split, where all models correctly identify the distinct object, despite its color being only slightly different from the others.}
    \label{fig:p3-color-1}
\end{figure}

\begin{figure}[!t]
    \centering
    \includegraphics[width=1\columnwidth]{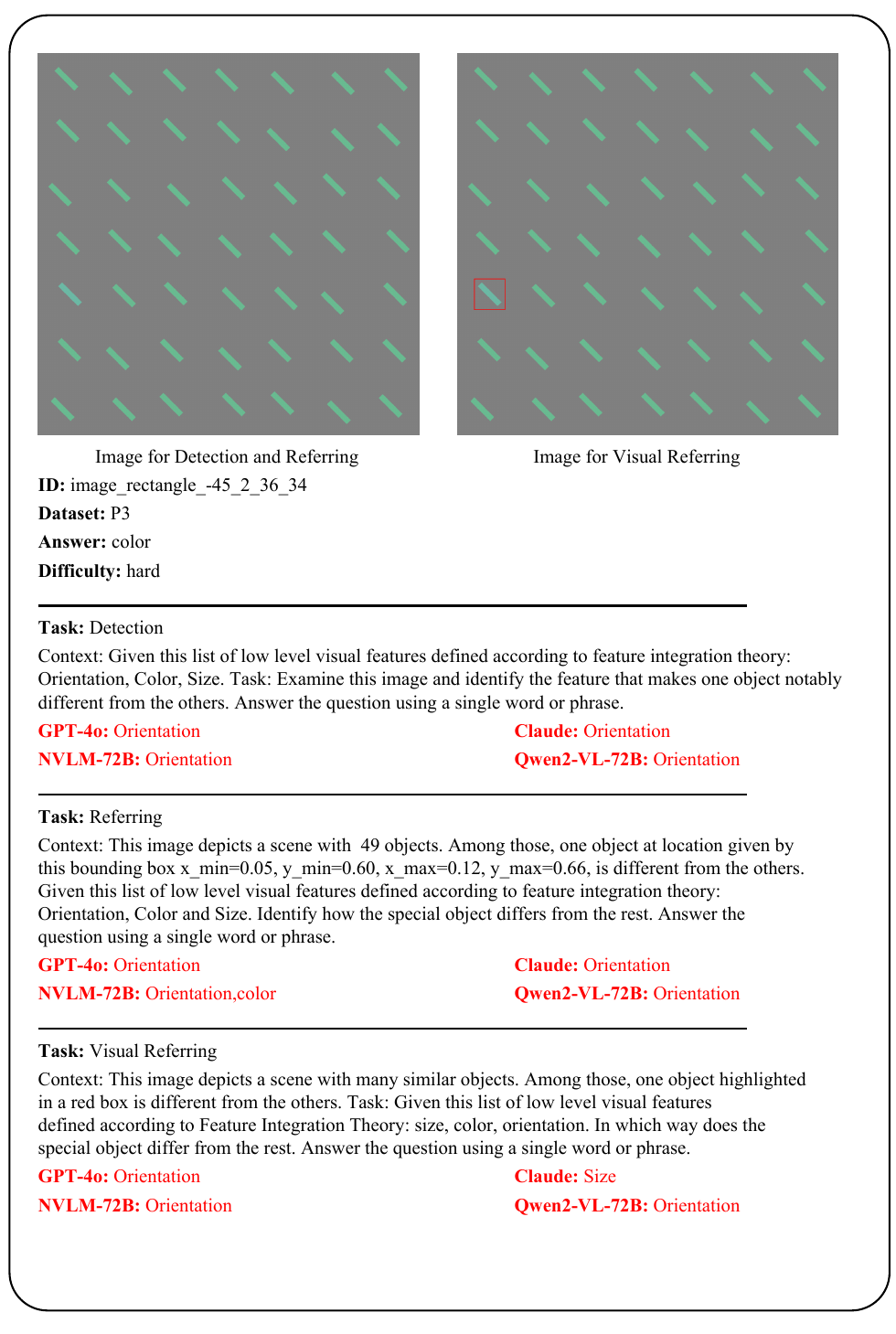}
    \caption{A hard example for the \textit{color} attribute in synthetic split, where the target's color is very similar to the distractors. All models fail to correctly identify the distinct feature, highlighting their difficulty in distinguishing subtle color variations.}
    \label{fig:p3-color-2}
\end{figure}

\begin{figure}[!t]
    \centering
    \includegraphics[width=1\columnwidth]{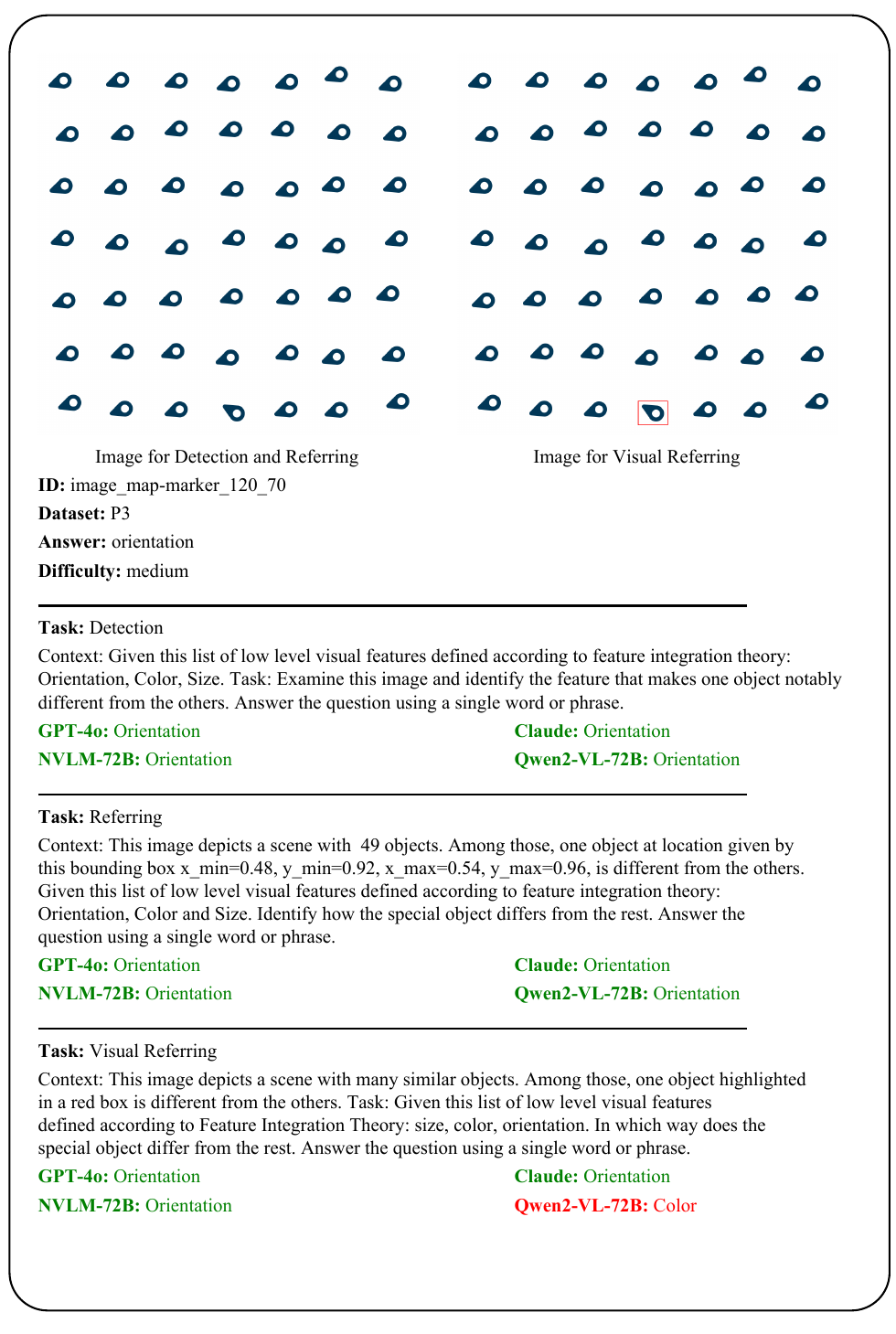}
    \caption{A medium example for the \textit{orientation} attribute in synthetic split, where the target's direction is rotated about 30 degree compared to the distractors. Most models correctly identify the distinct feature.}
    \label{fig:p3-orien-1}
\end{figure}

\begin{figure}[!t]
    \centering
    \includegraphics[width=1\columnwidth]{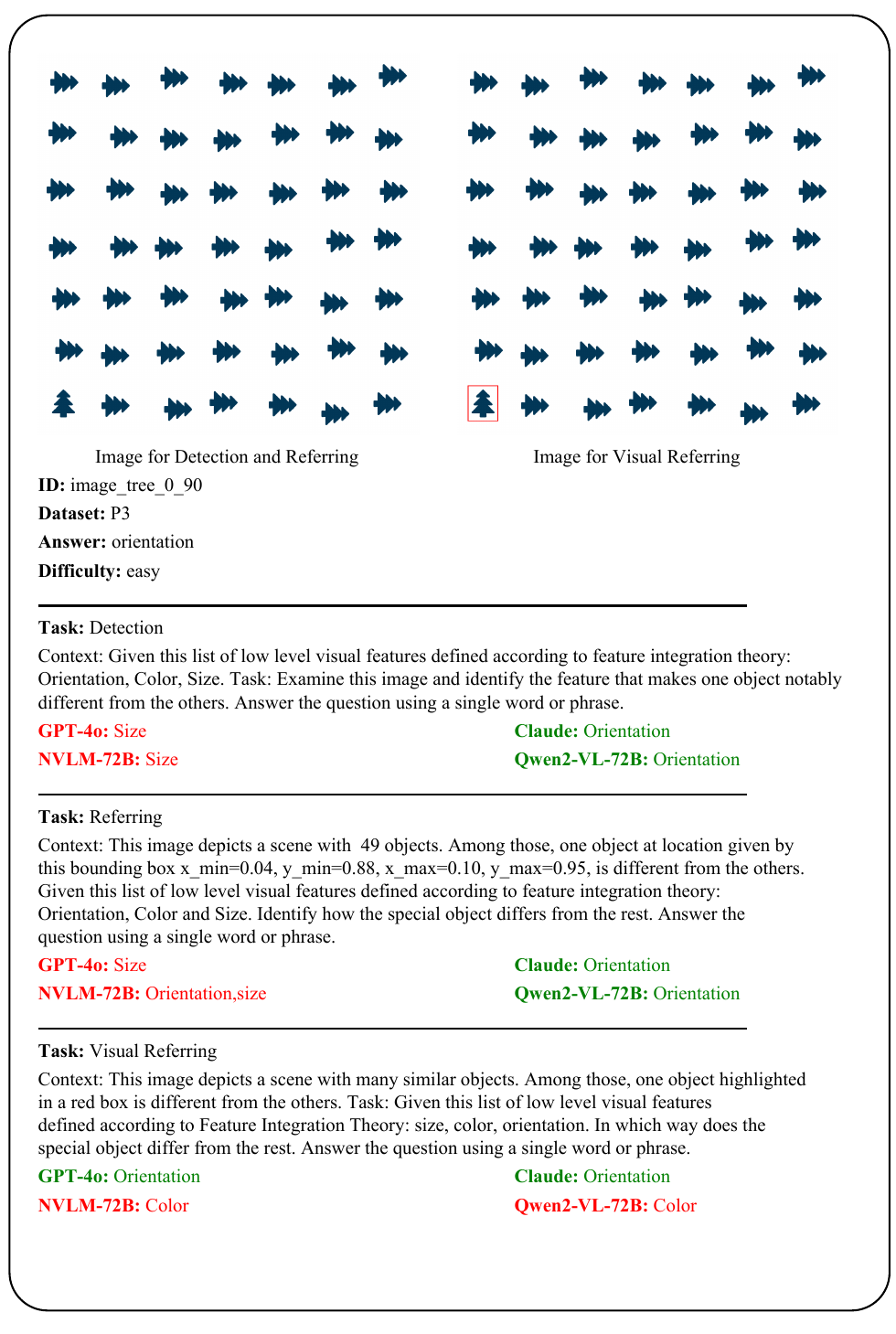}
    \caption{A easy example for the \textit{orientation} attribute in synthetic split, where the target's direction is rotated about 90 degree compared to the distractors. Some models like GPT-4o and  NVLM fail to correctly identify the distinct feature.}
    \label{fig:p3-orien-2}
\end{figure}

\begin{figure}[!t]
    \centering
    \includegraphics[width=1\columnwidth]{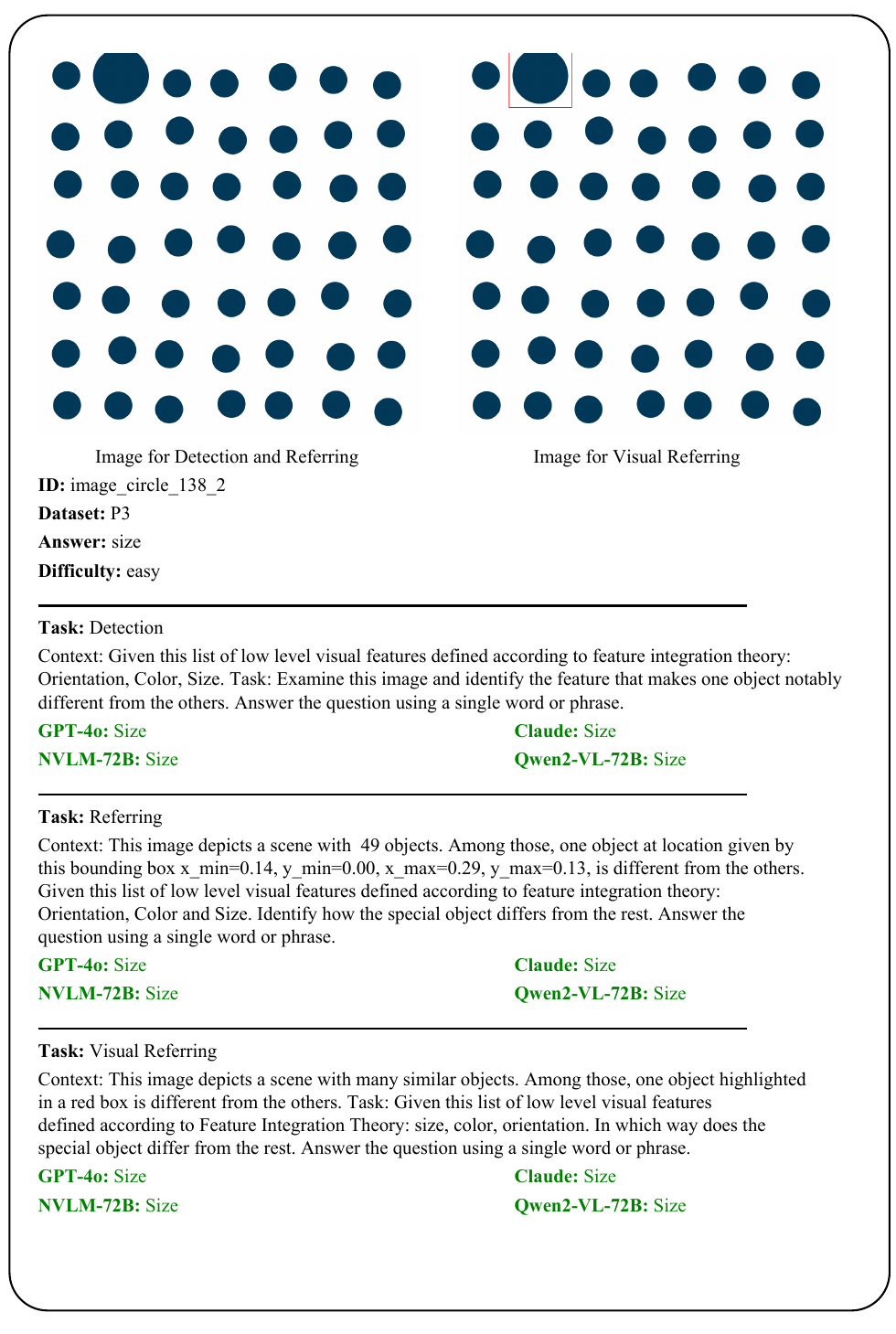}
    \caption{An easy example for the \textit{size} attribute in the synthetic split, where the circle target is four times larger than the others, making it straightforward for models to identify the odd visual attribute.}
    \label{fig:p3-size-1}
\end{figure}

\begin{figure}[!t]
    \centering
    \includegraphics[width=1\columnwidth]{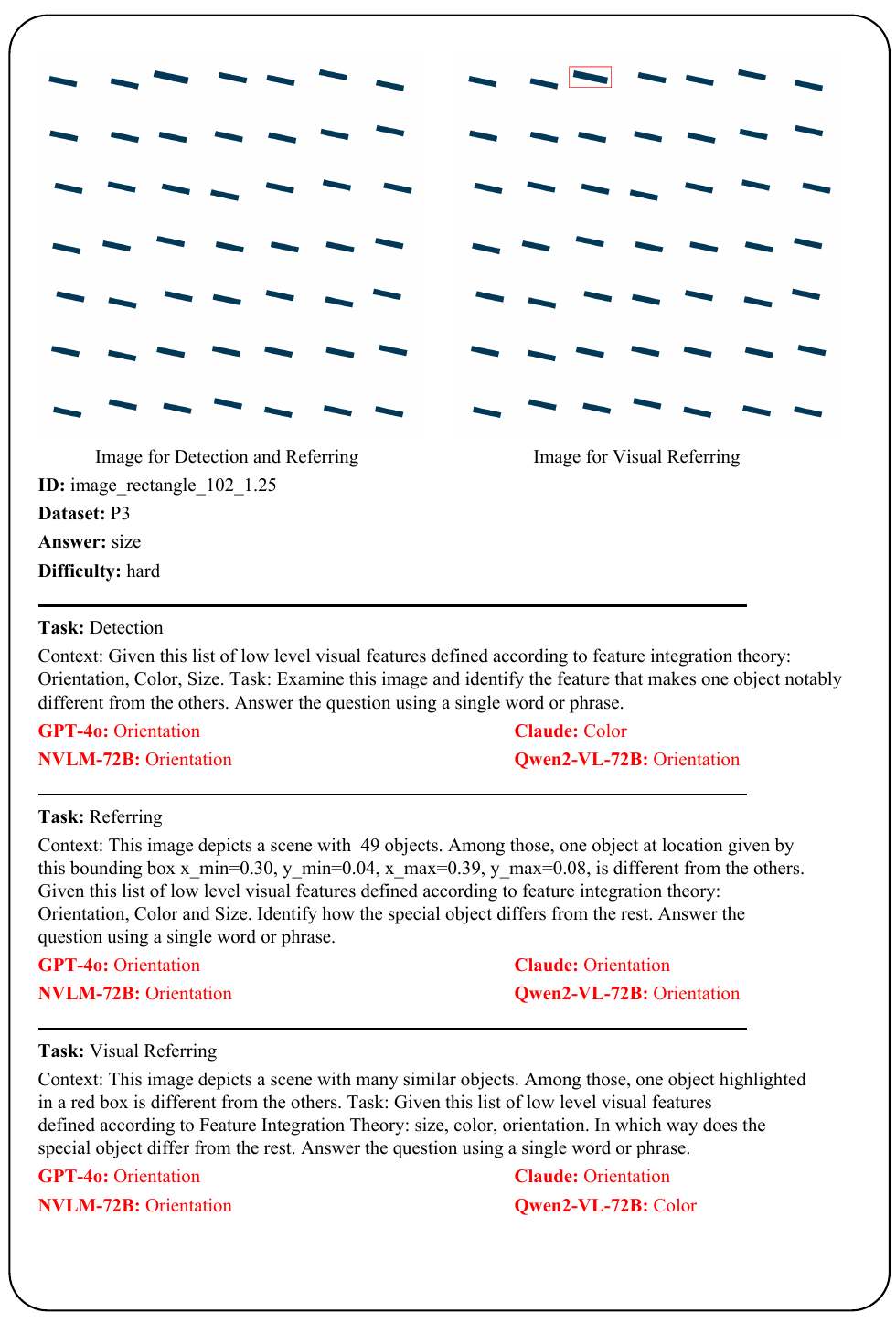}
    \caption{A hard example for the \textit{size} attribute in the synthetic split, where the distinct object's size difference is subtle. Most models incorrectly classify the distinguishing feature as \textit{orientation} or \textit{color}, reflecting the difficulty in identifying small size variations.}
    \label{fig:p3-size-2}
\end{figure}

\begin{figure}[!t]
    \centering
    \includegraphics[width=1\columnwidth]{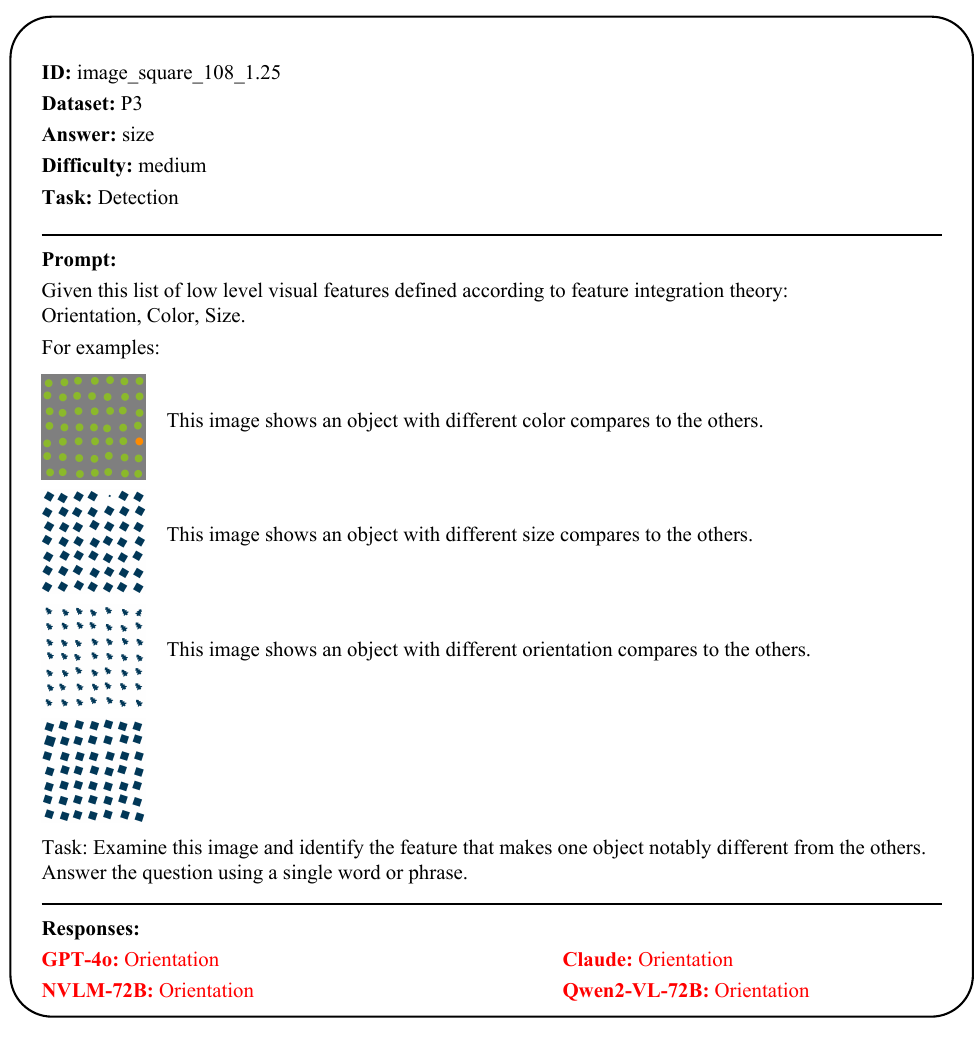}
    \caption{An example from the synthetic split under a 3-shot setting for the Detection task, with medium difficulty. The task involves identifying the \textit{size} attribute as the distinguishing feature. However, all models misclassify the feature as \textit{orientation}.}
    \label{fig:p3-fewshot-1}
\end{figure}

\begin{figure}[!t]
    \centering
    \includegraphics[width=1\columnwidth]{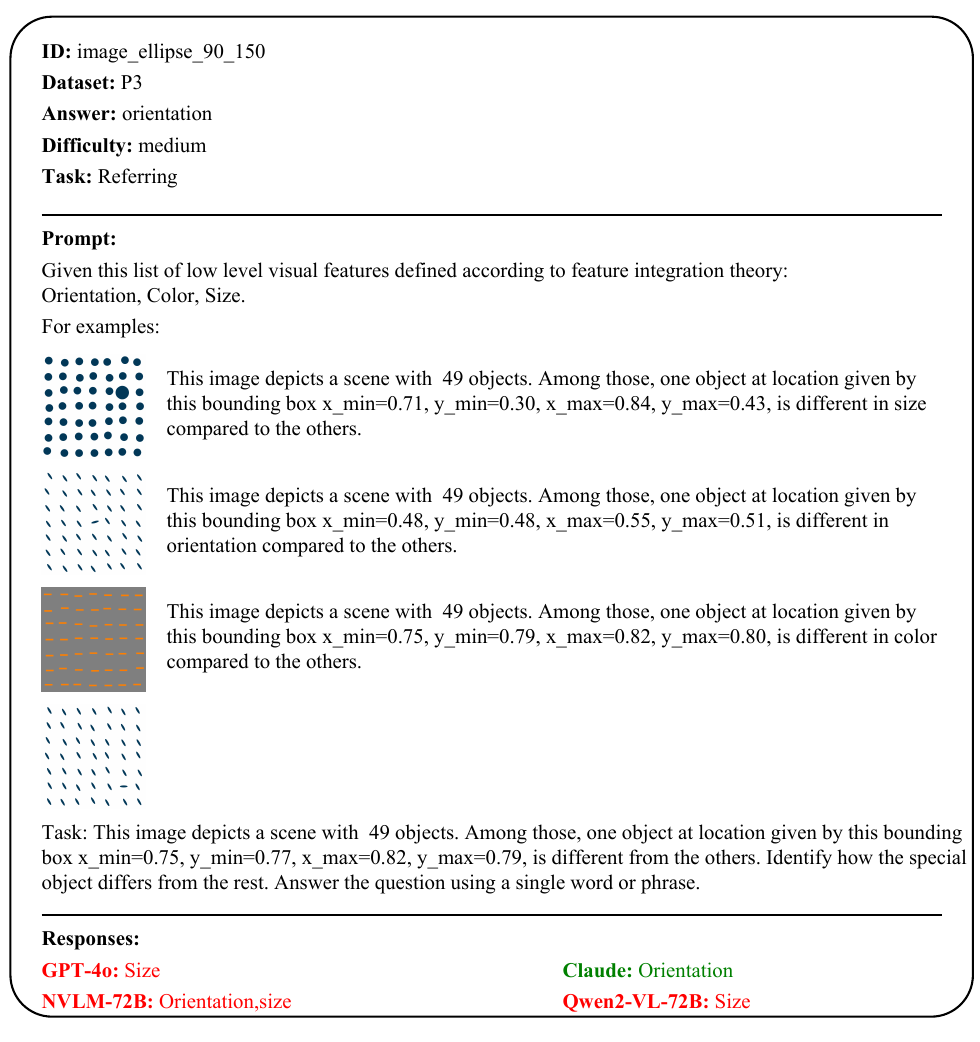}
    \caption{An example from the synthetic split under few-shot setting for Referring task with medium difficulty. Only Claude correctly identifies \textit{orientation}, while other models, such as GPT-4o and Qwen2-VL-72B, misclassify the feature as \textit{size}.}
    \label{fig:p3-fewshot-2}
\end{figure}

\begin{figure}[!t]
    \centering
    \includegraphics[width=1\columnwidth]{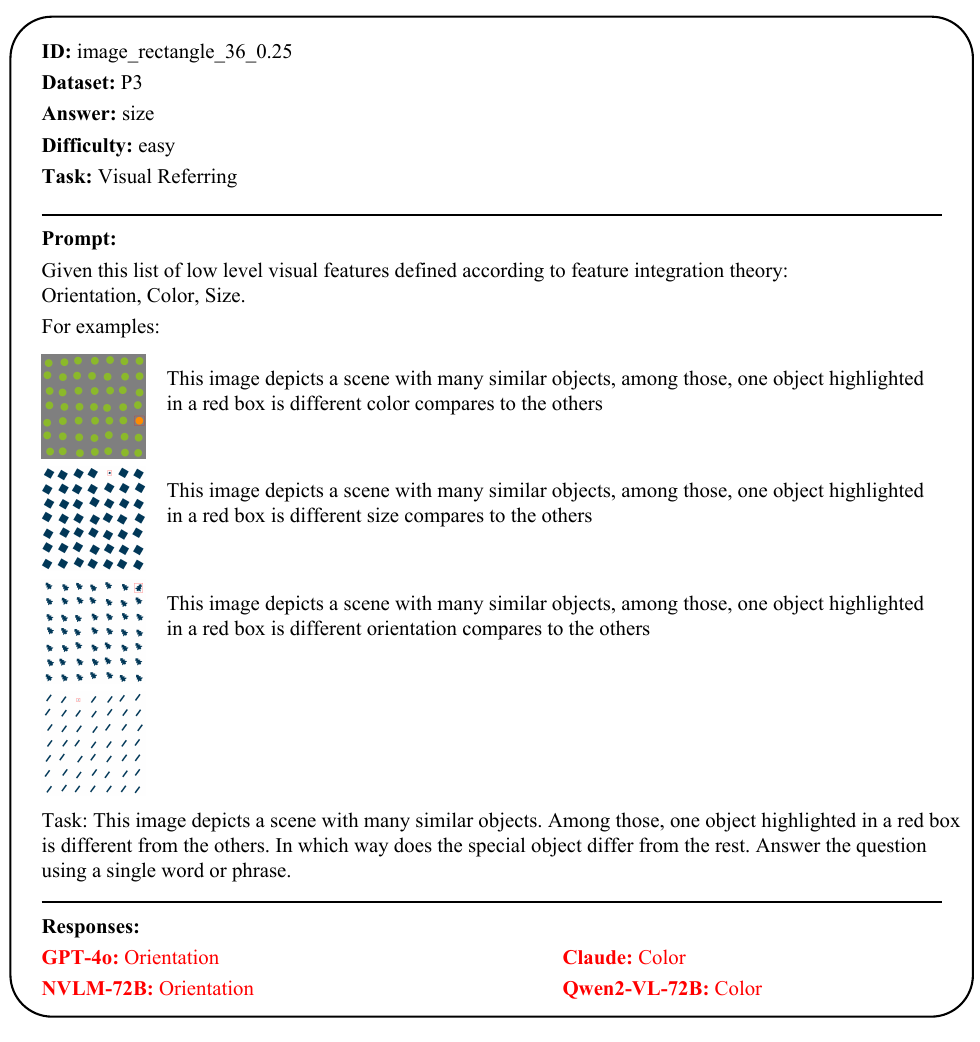}
    \caption{An example from the synthetic split in a few-shot Visual Referring task with easy difficulty. Claude and Qwen2-VL-72B incorrectly classify the feature as \textit{color}, while GPT-4o and NVLM-72B misclassify it as \textit{orientation}, despite the correct answer being \textit{size}.}
    \label{fig:p3-fewshot-3}
\end{figure}

\begin{figure}[!t]
    \centering
    \includegraphics[width=1\columnwidth]{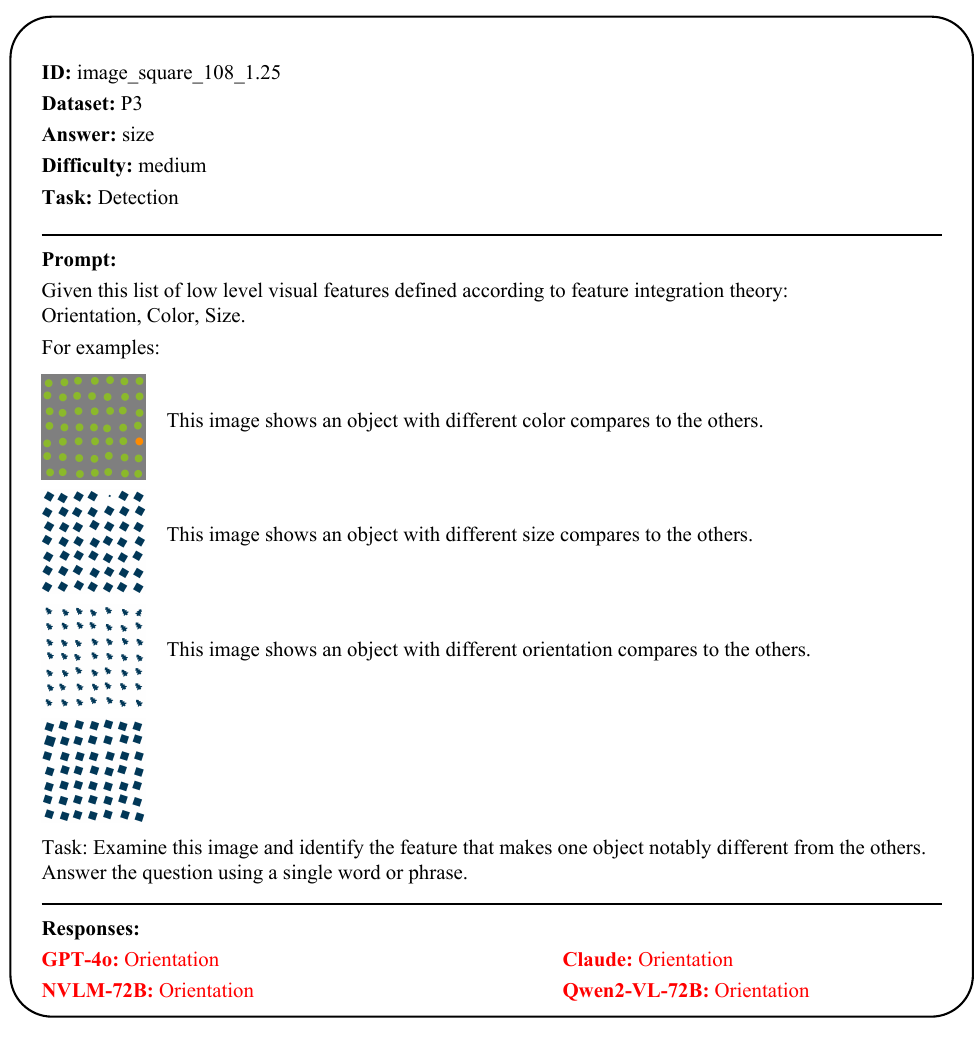}
    \caption{An example from the synthetic split for Detection task with medium difficulty. Despite the correct answer being \textit{size}, all models, including GPT-4o, Claude, NVLM-72B, and Qwen2-VL-72B, misclassify the feature as \textit{orientation}.}
    \label{fig:p3-fewshot-4}
\end{figure}

\begin{figure}[!t]
    \centering
    \includegraphics[width=1\columnwidth]{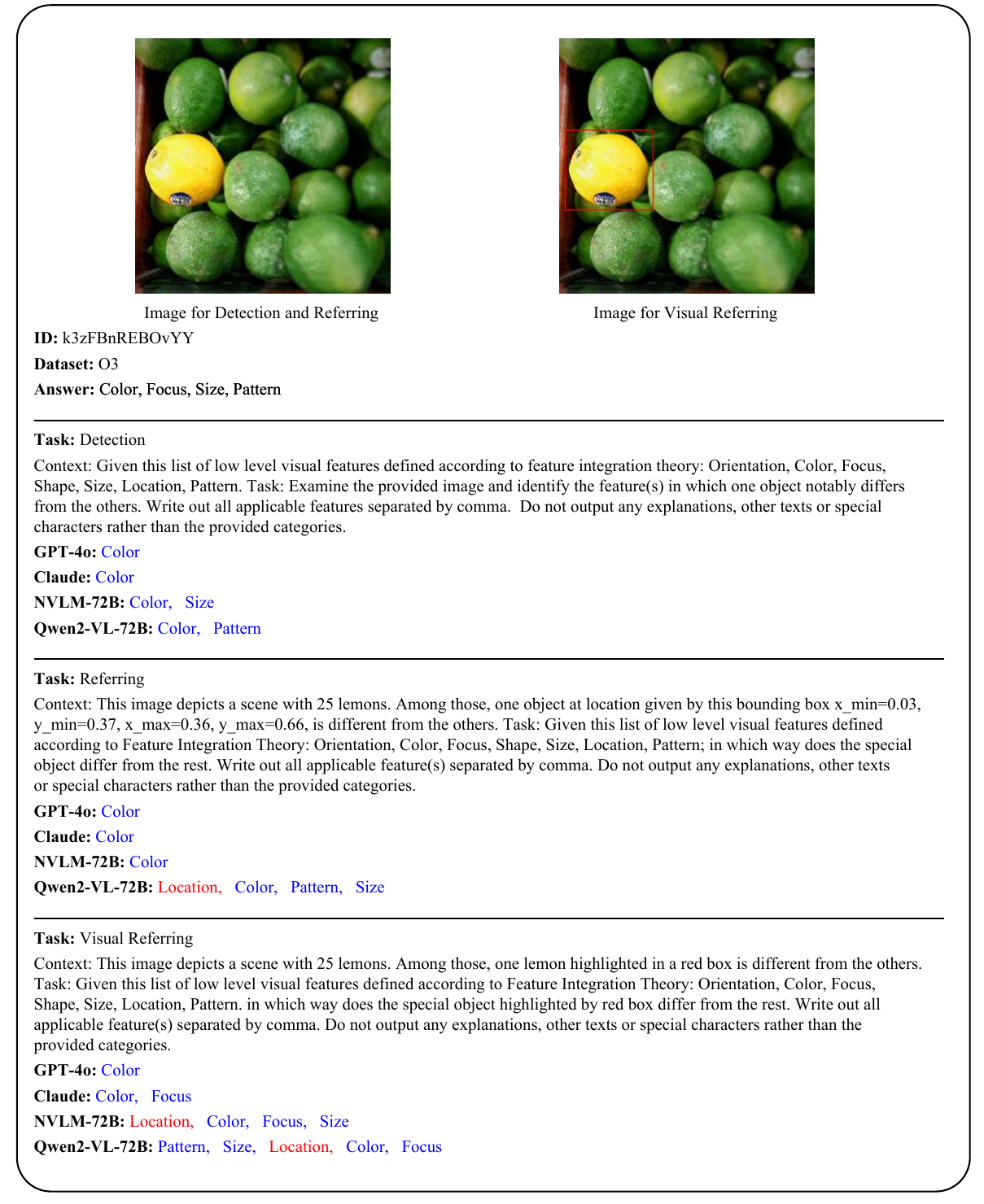}
    \caption{An example from the natural images split with an image comprising 25 lemons. We can clearly observe differences in \textit{color}, \textit{size}, and \textit{pattern} for the distinct lemon. However, for \textit{focus}, the variation is more subtle. While Qwen 2-VL-72B identifies most correctly, other models like GPT-4o and NVLM-72B predominantly predict  \textit{color}.
    Predictions in \textcolor{green}{green}, \textcolor{blue}{blue} \textcolor{red}{red} indicate exact, partially correct and incorrect responses, respectively.}
    \label{fig:o3-example-1}
\end{figure}

\begin{figure}[!t]
    \centering
    \includegraphics[width=1\columnwidth]{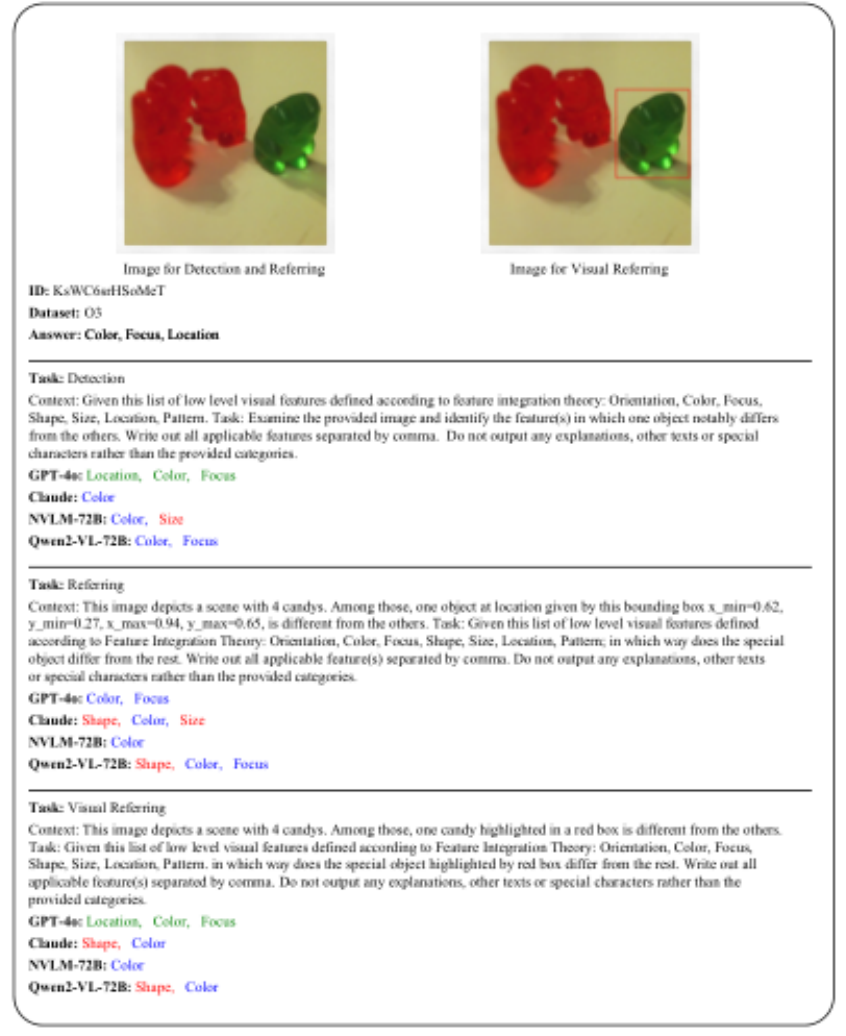}
    \caption{An example from the natural images split with a scene containing four candies. The distinct candy differs in \textit{color}, \textit{focus}, and \textit{location}. While some models, such as GPT-4o and Qwen2-VL-72B, correctly identify \textit{color} and \textit{focus}, others, like NVLM-72B, focus primarily on \textit{color}. Predictions in \textcolor{green}{green}, \textcolor{blue}{blue} \textcolor{red}{red} indicate exact match, partially correct and incorrect responses, respectively.}
    \label{fig:o3-example-2}
\end{figure}

\begin{figure}[!t]
    \centering
    \includegraphics[width=1\columnwidth]{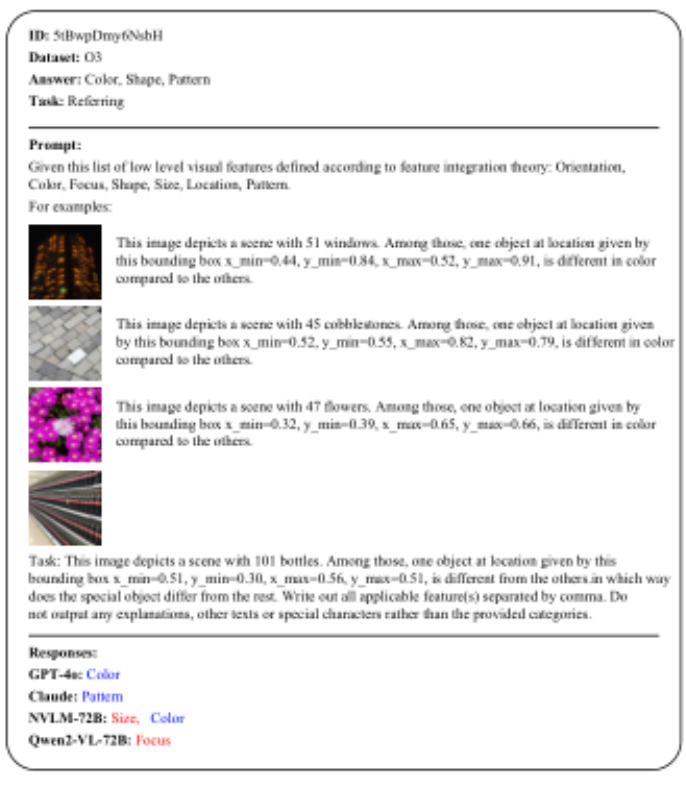}
     \caption{An example from the natural split under few-shot Referring task. The distinct object, a bottle, differs in \textit{color}, \textit{shape}, and \textit{pattern}. This task is particularly challenging due to the presence of numerous distractors with similar attributes, making it difficult for models to correctly identify the target. Responses vary, with GPT-4o focusing on \textit{color}, Claude identifying \textit{pattern}, NVLM-72B predicting \textit{size} and \textit{color}, and Qwen2-VL-72B focusing on \textit{focus}. Predictions in \textcolor{green}{green}, \textcolor{blue}{blue} \textcolor{red}{red} indicate exact match, partially correct and incorrect responses, respectively.}
    \label{fig:o3-example-3}
\end{figure}

\begin{figure}
    \centering
    \includegraphics[width=1\columnwidth]{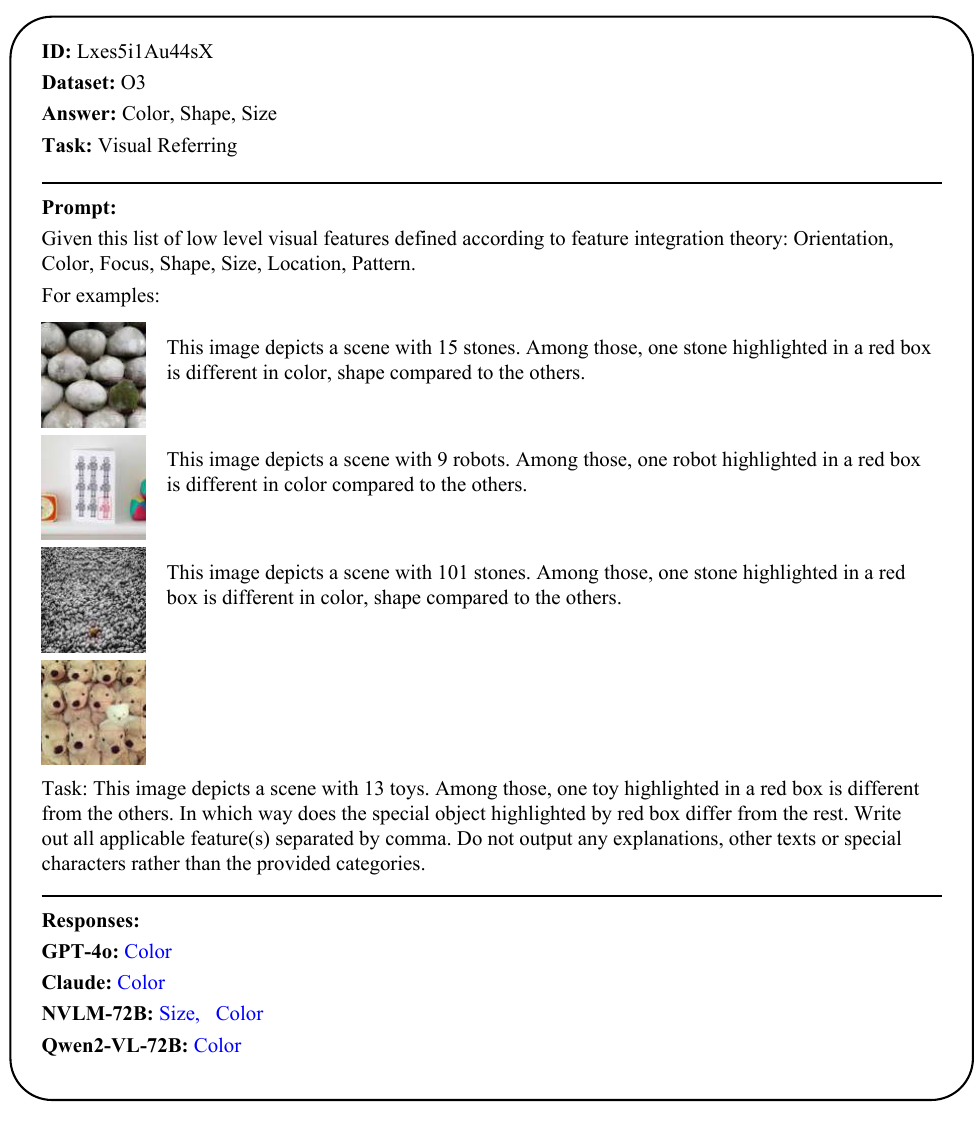}
     \caption{An example from the natural split in a few-shot Visual Referring task. The distinct object, a toy, differs in \textit{Color}, \textit{Shape}, and \textit{Size}. Despite the task requiring recognition of multiple attributes, all models fail to identify the \textit{Shape} attribute. GPT-4o, Claude, and Qwen2-VL-72B focus only on \textit{color}, while NVLM-72B identifies \textit{size} and \textit{color}. Predictions in \textcolor{green}{green}, \textcolor{blue}{blue} \textcolor{red}{red} indicate exact match, partially correct and incorrect responses, respectively.}
    \label{fig:o3-example-4}
\end{figure}

Here, we present the qualitative comparisons between the vision-language models predictions on few example images from our SalBench.
\reffig{fig:p3-color-1} to \reffig{fig:p3-fewshot-4} illustrate the comparison on the synthetic split under different settings and difficulty levels, while \reffig{fig:o3-example-1} to \reffig{fig:o3-example-4} show the example comparisons on natural images split. 

In summary, all these additional results and analysis highlight the importance of our proposed SalBench in evaluating the vision-language models on low-level perceptual tasks.

\end{document}